\documentclass[runningheads]{llncs}
\usepackage{graphicx}

\usepackage{tikz}
\usepackage{comment}
\usepackage{amsmath,amssymb} 
\usepackage{color}

\usepackage{booktabs}
\usepackage{caption}
\usepackage{subcaption}

\usepackage{xcolor}
\graphicspath{{img/}}
\usepackage{transparent}
\usepackage{amsmath}
\usepackage{mathrsfs}

\usepackage{hyperref}
\usepackage{cleveref}
\crefname{section}{Sec.}{Secs.}
\Crefname{section}{Section}{Sections}
\Crefname{table}{Table}{Tables}
\crefname{table}{Tab.}{Tabs.}
\crefname{figure}{Figure}{Figures}
\crefname{figure}{Fig.}{Figs.}
\crefname{equation}{Equation}{Equations}
\crefname{equation}{Eq.}{Eqs.}
\crefname{appendix}{Appendix}{Appendix~}
\usepackage[accsupp]{axessibility}  

\begin{document}
\pagestyle{headings}
\mainmatter
\def\ECCVSubNumber{100}  

\title{GraphFit: Learning Multi-scale Graph-Convolutional Representation 
for \\ Point Cloud Normal Estimation} 

\titlerunning{GraphFit}
%
\author{Keqiang Li$^\star$\inst{1,3} \and
Mingyang Zhao\thanks{The authors contribute equally in this work.}\inst{1,2} \and
Huaiyu Wu \inst{1}  \and
Dong-Ming Yan \inst{1,3}  \and \\
Zhen Shen \inst{1}  \and
Fei-Yue Wang \inst{1} \and
Gang Xiong \inst{1}}
\authorrunning{Li et al.}
%
\institute{SKLMCCS/NLPR, Institute of Automation, Chinese Academy of Sciences 
 \and
Beijing Academy of Artificial Intelligence \and
School of Artificial Intelligence, University of Chinese Academy of Sciences \\
\email{\{likeqiang2020,  huaiyu.wu, zhen.shen, feiyue.wang, gang.xiong \}@ia.ac.cn}
\email{{myzhao}@baai.ac.cn, {yandongming}@gmail.com}
}


\maketitle
\begin{abstract}
We propose a precise and efficient normal estimation method that can deal with noise and nonuniform density for unstructured 3D point clouds. Unlike existing approaches that
directly take patches and ignore the local neighborhood relationships, which make them susceptible to challenging regions such as sharp edges, we propose to learn \emph{graph convolutional feature representation} for normal estimation, which emphasizes more local neighborhood geometry and effectively encodes intrinsic relationships. Additionally, we design a novel adaptive module based on the \emph{attention mechanism} to integrate point features with their neighboring features, hence further enhancing the robustness of the proposed normal estimator against point density variations. To make it more distinguishable, we  introduce a \emph{multi-scale architecture} in the graph block to learn richer  geometric features. Our method outperforms competitors with the \emph{state-of-the-art} accuracy on various benchmark datasets, and is quite robust against noise, outliers, as well as the density variations. The code is available at 
{\textcolor{blue}{\url{https://github.com/UestcJay/GraphFit}}}.
\keywords{Normal estimation, unstructured 3D point clouds, graph convolution, multi-scale}
\end{abstract}

\section{Introduction}\label{sec:intro}
The normal estimation of point clouds is a fundamental problem in 3D computer vision and computer graphics, which has a wide variety of applications in practice. Commonly, the scanned point clouds only contain spatial locations along with sampling density, noise, outliers or textures, while lacking local surface geometry, like point normals. High-quality normals can facilitate a large number of downstream tasks, such as point cloud denoising~\cite{lu2020low,lu2020deep}, surface reconstruction~\cite{kazhdan2006poisson,fleishman2005robust} and model segmentation~\cite{che2018multi}. 

Normal estimation is essential and has been studied for a long time, yet not well-solved. Traditional methods~\cite{hoppe1992surface,levin1998approximation,cazals2005estimating,boulch2012fast} usually adopt \emph{principle component analysis} (PCA) and \emph{singular value decomposition} (SVD) for normal estimation. They attain satisfactory results for simple and clean data but suffer from noise, outliers, and complex shapes. Moreover, their performance heavily depends on the parameter tuning. Recently, several learning-based methods~\cite{guerrero2018pcpnet,ben2019nesti,zhou2020geometry,wang2020neighbourhood,hashimoto2019normal}  have been proposed to directly regress normals and have exhibited promising performance. Nevertheless, as pointed out in~\cite{zhu2021adafit}, direct regression brings in finite generalization and stability, especially for laser scanned real-world point clouds. 

\begin{figure}[t]
  \centering
    \begin{subfigure}{0.17\linewidth}
    \includegraphics[scale=0.4]{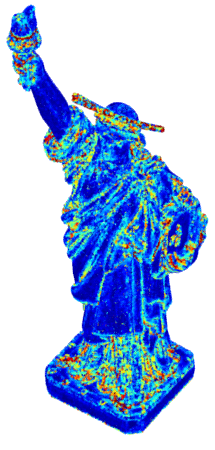}
    \caption{PCPNet}
  \end{subfigure}
      \begin{subfigure}{0.17\linewidth}
    \includegraphics[scale=0.4]{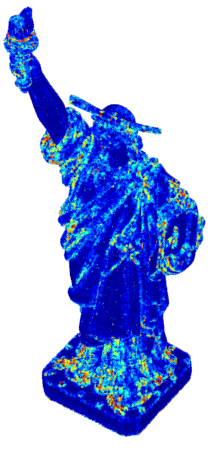}
    \caption{NestiNet}
  \end{subfigure}
  \begin{subfigure}{0.17\linewidth}
    \includegraphics[scale=0.4]{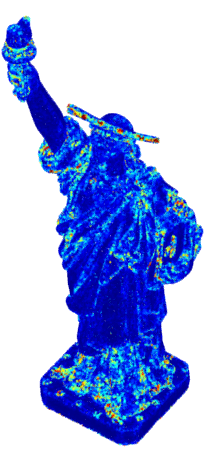}
    \caption{DeepFit}
  \end{subfigure}
  \begin{subfigure}{0.17\linewidth}
    \includegraphics[scale=0.4]{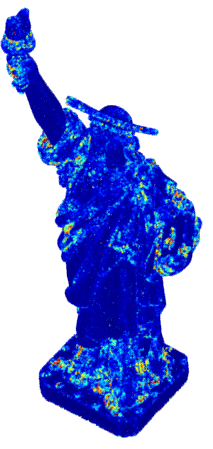}
    \caption{AdaFit}
  \end{subfigure}
  \begin{subfigure}{0.17\linewidth}
    \includegraphics[scale=0.4]{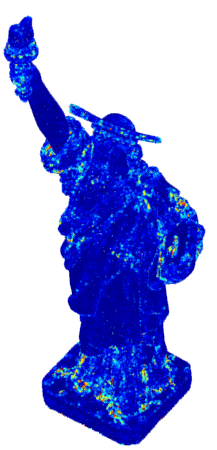}
    \caption{Ours}
  \end{subfigure}
    \begin{subfigure}{0.09\linewidth}
        \includegraphics[scale=0.23]{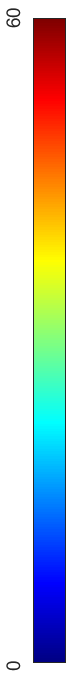}
  \end{subfigure}
 \caption{Comparison of the normal estimation error (colored by the heat map) of learning-based methods PCPNet~\cite{guerrero2018pcpnet}, NestiNet~\cite{ben2019nesti}, DeepFit~\cite{ben2020deepfit}, AdaFit~\cite{zhu2021adafit} and the proposed method, where our method shows higher estimation accuracy, especially on complex regions.}
   \label{fig:overview}
\end{figure}

Different from the previous brute-force regression, a more accurate paradigm is to combine traditional methods involving geometry information with learning-based models for normal estimation, which typically specifies a fixed neighborhood around each point and then fits a local surface such as a plane to infer the normal. However, when point clouds are contaminated by noise or outliers, the estimated normals tend to be erroneous. To mitigate this problem, the \emph{weighted least-squares} (WLS) fitting is invoked ~\cite{lenssen2020deep,ben2020deepfit,zhang2022geometry,zhu2021adafit}, in which point-wise weights are predicted. Despite that the improvements of accuracy and robustness, precise normal estimation in complex regions are still difficult, as illustrated in \cref{fig:overview}. In principle, point normals are local geometric properties, and are significantly affected by the  geometric relations among the local neighborhood. However, previous methods 
directly adopt patches for normal estimation, and usually ignore the \emph{intrinsic relationship} between points in the same patch, which empirically enables richer features and higher accuracy.

Motivated by this observation, in this paper, we present a novel method using graph-convolutional layers for robust and accurate normal estimation. Benefiting from the graph structure, we not only extract features but also encode local geometric relationship between points in the same patch. To enhance the feature
integration capability and generate richer geometric information, we further utilize the adaptive module based on a attention mechanism to integrate point features with its neighboring features. Additionally, we design a novel multi-scale representation to learn more accurate and expressive local information, as well as reducing the influence of noise and outliers. In contrast to the cascaded scale aggregation layer used in previous approaches, which directly takes the global features of varying patches, our proposed multi-scale representation fuses local information from two different scales in the feature space, and casts more attention on the local features. We compare the proposed method with representative state-of-the-art approaches on popular benchmark datasets. Results demonstrate that our method outperforms competitors with higher accuracy, and it is quite robust against noise, outliers and density variations. We further verify the advantages of our method by applying it to surface reconstruction and point cloud denoising in the supplemental material. To summarize, our main contributions are threefold as follows: 
\begin{itemize}
    \item We propose a new method for accurate and robust normal estimation via the graph-convolutional feature learning, which effectively integrates local features and their relationships among the same point cloud patch.
    \item We design an adaptive module using the attention mechanism to fuse the point features with its neighboring features, which brings high-quality feature integration.
    \item We introduce a multi-scale representation module to extract more expressive features, bringing higher robustness against noise and varying point density. 
\end{itemize}

\section{Related Work}\label{sec:related}

\subsection{Classical Methods}
The most popular and simplest way for normal estimation is based on the principal component analysis~\cite{hoppe1992surface}, in which the normal of a query point is calculated as the eigenvector corresponding to the smallest eigenvalue of a covariance matrix. Although this method is simple, it usually suffers from the choice of the neighboring size, noise, and outliers. Later, many variants such as \textit{moving least squares} (MLS)~\cite{levin1998approximation}, \emph{truncated Taylor expansion} (n-jet)~\cite{cazals2005estimating}, and fitting local spherical surfaces~\cite{guennebaud2007algebraic} have been proposed. These approaches usually select a large-scale neighborhood to improve the robustness against noise and outliers, meanwhile trying to keep a correct normal estimation for sharp features.
Particularly, Mitra et al.~\cite{mitra2003estimating} delicately analyze the influence of the neighboring size, the curvature, sampling density, and noise to find an optimal neighbor radius. To attain more features, some methods utilize 
\emph{Voronoi diagram}~\cite{alliez2007voronoi,amenta1999surface,dey2006provable,merigot2010voronoi} to estimate the structure of the underlying surface, while others adopt {\em Hough Transform} (HT)~\cite{boulch2012fast} to achieve analogous effects. Nevertheless, HT-based methods typically 
require high computational complexity and fine tuning of hyper-parameters. {Recent works have also been designed to robustly estimate normals for point clouds contaminated by outliers~\cite{nurunnabi2015outlier,khaloo2017robust} or noise~\cite{giraudot2013noise} with an  adaptive neighboring size~\cite{comino2018sensor,nurunnabi2014robust,castillo2013point}}.
\subsection{Learning-based Approaches}
Recently, with the marvelous success of deep learning in a wide variety of domains~\cite{qi2017pointnet,qi2017pointnet++,wang2019dynamic,rakotosaona2020pointcleannet,zhang2020pointfilter,hermosilla2019total,yu2018ec,pistilli2020learning,fan2021scf}, some learning-based attempts have been made to estimate normals of point clouds, which can be generally categorized into two types: \emph{regression-based}  and \emph{geometry-guided} methods.
\\
\textbf{Regression-based methods.} Due to the estimation of surface normals from patches and the use of fully connected layers, regression-based methods are commonly simple. For instance, Lu et al.~\cite{lu2020deep} transform point cloud patches into 2D height maps by computing the distances between points and a plane, while Ben-Shabat et al.~\cite{ben20183dmfv} modify the \emph{Fisher Vector} to describe points with their deviations from a \emph{Gaussian Mixture Model} (GMM). HoughCNN~\cite{boulch2016deep} projects points into a Hough space via the Hough transform and then uses a 2D CNN to regress the normal vector. Another line of studies focus on the direct regression for unstructured point clouds. Inspired by the high efficiency of PointNet~\cite{qi2017pointnet}, \cite{guerrero2018pcpnet} proposes a deep multi-scale PointNet for normal estimation (PCPNet), but requires a set of scale values. Hashimoto et al.~\cite{hashimoto2019normal} also use PointNet to extract the local features such as neighboring points while extract spatial representation by the voxel network 3DCNN. Nesti-Net~\cite{ben2019nesti} introduces a new normal estimation method for irregular 3D point clouds based on the mixture of experts and scale prediction, which yields high accuracy but suffers from high computational complexity.

\noindent{\textbf{Geometry-guided methods.}} Despite that regression-based methods
are direct and simple, they usually produce weak generalization and unstable prediction results. To circumvent these limitations, most recent works combine deep learning with classical geometric methods. IterNet~\cite{lenssen2020deep} and DeepFit~\cite{ben2020deepfit} employ a deep neural network to learn point-wise weights for weighted-least-squares fitting to estimate the normals, leading to good normal estimation quality and have been extended to estimate other geometrical properties such as principal curvatures. Zhang et al. \cite{zhang2022geometry} propose a geometry-guided network for robust surface normal estimation, which improves the learning performance and the interpretability of the weights. A more recent work, AdaFit~\cite{zhu2021adafit} incorporates a \emph{Cascaded Scale Aggregation} (CSA) layer to aggregate features from multiple neighborhood sizes and adds additional offsets to enable the output normals more robust and accurate. Different from previous approaches, in this paper, we propose to use graph-convolutional layers for feature learning and normal estimation, and incorporate a multi-scale architecture that emphasizes richer local geometric features to generate more accurate normal estimation results.

\section{Methodology}\label{sec:meth}
\noindent{\textbf{Problem definition.}} Given a 3D point cloud $\mathcal{P}=\left\{\mathbf{p}_{i}\in \mathbb{R}^3\right\}_{i=1}^{N}$ and a query point $\mathbf{p}_{i} \in \mathcal{P}$, our target is to solve for the normal of each point in $\mathcal{P}$. We first extract a local patch $N_k\left(\mathbf{p}_{i}\right)=\left\{\mathbf{p}_{i}\in \mathbb{R}^3 \mid i=1,2, \ldots, N_{p}\right\}  \subset \mathcal{P}$, $N_{p}$ is the size of $N_k\left(\mathbf{p}_{i}\right)$, then we employ the truncated Taylor expansion (n-jet) for surface fitting~\cite{cazals2005estimating} to represent any regular embedded smooth surface, as the graph of a bi-variate \textit{height function} with respect to any $z$ direction does not belong to the tangent space. An n-order Taylor expansion of the \textit{height function} is defined as 
\begin{equation}
z = f\left(x,y\right)
=J_{{\beta}, n}(x, y)
=\sum_{k=0}^{n} \sum_{j=0}^{k} {\beta}_{k-j, j} x^{k-j} y^{j},
\label{eq:n-jet}
\end{equation}
where $\boldsymbol{\beta} = \left\{{\beta}_{k-j, j} \mid j=0,1,\cdots,k; k=0,1,\cdots,n\right\}$ are the coefficients of the jet that consists of $N_{n}=(n+1)(n+2) / 2$ terms. 
The goal is to fit a surface to $N_k\left(\mathbf{p}_{i}\right)$. We define the \textit{Vandermonde matrix} $\mathbf{M} =\left(1, x_{i}, y_{i}, \ldots, x_{i} y_{i}^{n-1}, y_{i}^{n}\right)_{i=1, \ldots, N_{p}} \in \mathbb{R}^{N_{p} \times N_{n}}$ and the height function vector $\mathbf{z} =\left(z_{1}, z_{2},\ldots, z_{N_{p}}\right) \in \mathbb{R}^{N_{p}}$.
The sampled points in \cref{eq:n-jet} can be expressed as 
\begin{equation}
\mathbf{M} \boldsymbol{\beta}=\mathbf{z}.
\end{equation}
To enhance the robustness of the fitting method against noise and outliers, like~\cite{ben2020deepfit}, we adopt the weighted least-squares fitting of polynomial surfaces here, which can also relieve the problem of over-fitting or under-fitting~\cite{zhu2021adafit}. Then our objective is to predict the point-wise weight $w_{i}$ and offset $\left(\Delta x_{i},\Delta y_{i}, \Delta z_{i}\right)$ to adjust the point distribution. The optimization problem becomes:

\begin{figure*}[t]
  \centering
   \includegraphics[scale=0.36]{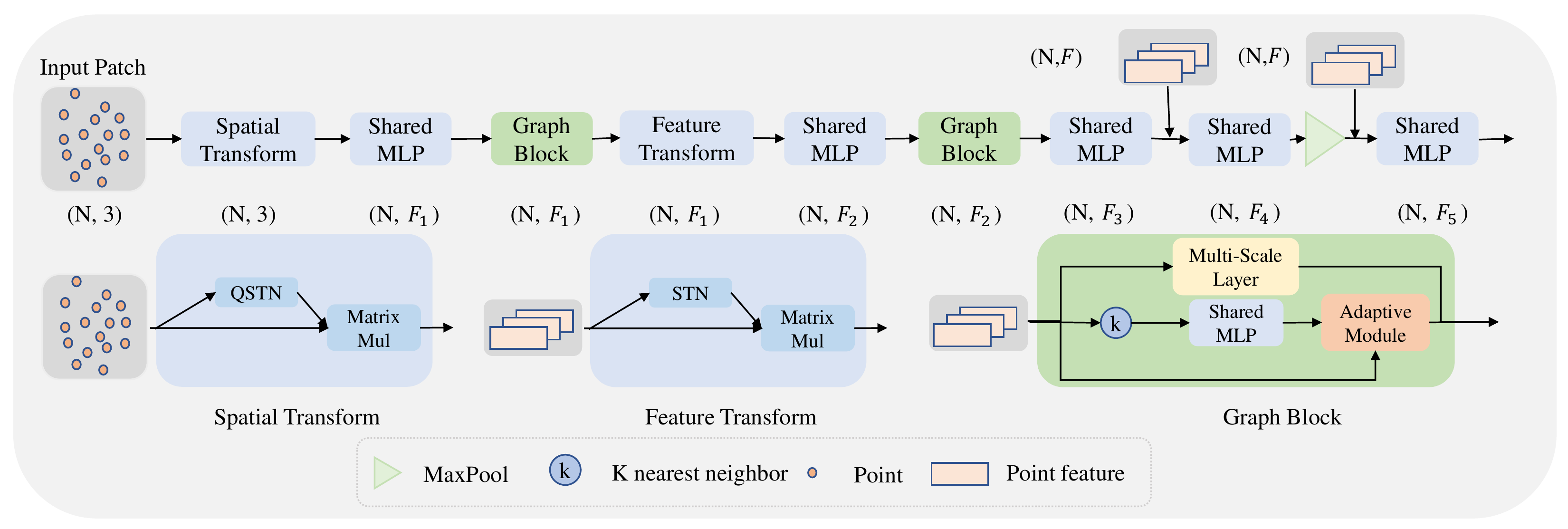}
   \caption{\textbf{Overview of our proposed network architecture}. Graph block is utilized to encode the relationship between neighbor points, which contains an adaptive model to effectively integrate point features with the local neighbor features. A multi-scale layer is employed to extract richer local features. Given an input patch, our network predicts a point-wise weight and offset to fit a surface for normal estimation.}
   \label{fig:network}
\end{figure*}
\begin{equation}
\hat{\boldsymbol{\beta}}=\underset{\beta}{\operatorname{argmin}} \sum_{i}^{N_{p}} w_{i}\left\|J_{\beta, n}\left(x_{i}+\Delta x_{i}, y_{i}+\Delta y_{i}\right)-\left(z_{i}+\Delta z_{i}\right)\right\|^{2}.
\label{eq:beta_adafit}
\end{equation}
The solution to \cref{eq:beta_adafit} can be expressed in a closed-form:
\begin{equation}
\hat{\boldsymbol{\beta}}=\left(\mathbf{M}^{\top} \mathbf{W} \mathbf{M}\right)^{-1}\left(\mathbf{M}^{\top} \mathbf{W} \mathbf{z}\right),
\label{eq:beta_solu}
\end{equation}
where $\mathbf{W}$ is a diagonal weight matrix $\mathbf{W}=\operatorname{diag}\left(w_{1}, w_{2}, \ldots, w_{N_{p}}\right) \in \mathbb{R}^{N_{p} \times N_{p}}$, in which the diagonal element $w_{i}$ is the weight of the point $\mathbf{p}_{i}$. Once getting the jet coefficients $\boldsymbol{\beta}$, the estimated normal $\hat{\mathbf{n}}_i$ of the point $\mathbf{p}_{i}$ is
\begin{equation}
\hat{\mathbf{n}}_i=\frac{\left(-\beta_{1},-\beta_{2}, 1\right)}{\left\|\left(-\beta_{1},-\beta_{2}, 1\right)\right\|_{2}},
\label{eq:normal}
\end{equation}
where we define $\beta_{1} = {\beta}_{1, 0}, \beta_{2} = {\beta}_{0, 1}$. Then the normals of neighboring points of the query point can be calculated by transforming n-jet to the implicit surface form, e.g., $F\left(x, y, z\right) = 0$: 
\begin{equation}
\hat{\mathbf{n}}_{j}=\left.\frac{\nabla F}{\|\nabla F\|}\right|_{p_{i, j}}=\left.\frac{\left(-\beta \frac{\partial \mathbf{M}^{T}}{\partial x}, \beta \frac{\partial \mathbf{M}^{T}}{\partial y}, 1\right)}{\|\nabla F\|}\right|_{p_{i, j}}.
\end{equation}

\noindent\textbf{{Motivation.}}\label{sec:leran-graph}
The point normal on the surface is a locally geometric property, which is significantly affected by their mutual relations. To investigate this characteristic, we utilize the local neighbor information of the points in the same patch to achieve more accurate surface fitting. An overview of the proposed network architecture is presented in \cref{fig:network}. The first block is composed of two point convolutions that gradually transform the 3D spatial points into a higher feature space. Then a cascade
of two \emph{graph blocks} is used, with several \emph{skip connections} to enhance the feature expression ability. These representations are then concatenated and fed into \textit{Multi-layer Perceptrons} (MLP) to predict weights and offsets for all points. We present detailed descriptions of the designed network architecture in the following.

 \begin{figure}[t]
  \centering
   \includegraphics[scale=0.36]{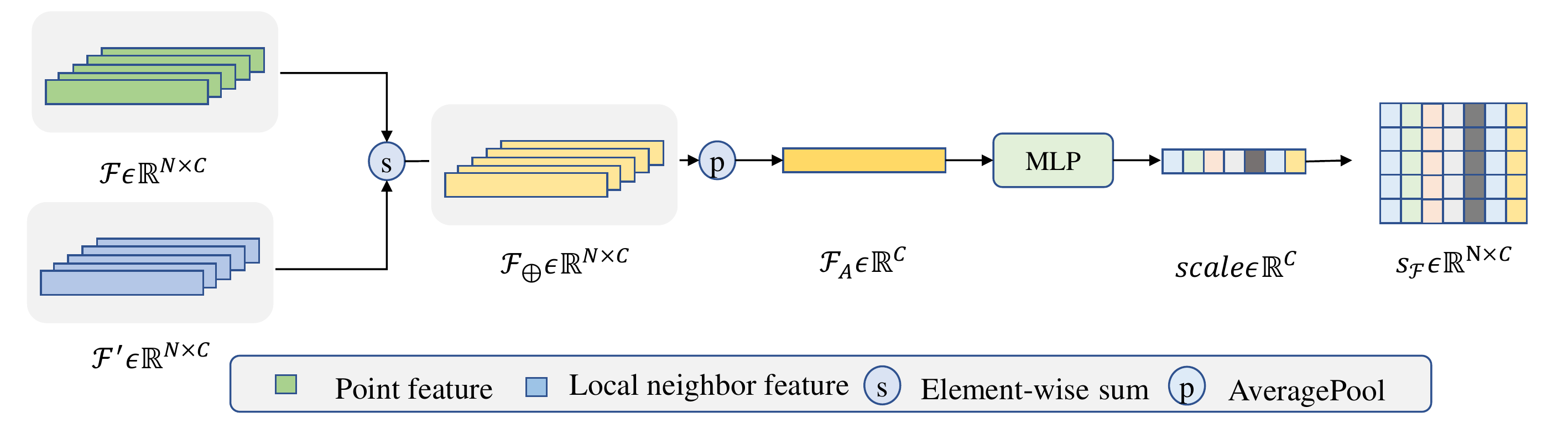}
   \caption{\textbf{Our proposed adaptive module}. It integrates neighbour features with point features based on the attention mechanism.}
   \label{fig:attention}
\end{figure}
\noindent{\textbf{{Graph block.}}} The core of the proposed network in our method is the graph block. \emph{Graph convolution}, as a generalization of the ordinary convolution to point data,  represents point clouds with graph structures. Suppose the features $\mathcal{F}=\left\{f_{i} \mid i=1,2, \ldots, N\right\} \in \mathbb{R}^{N \times C}$ correspond to the input point cloud  $\mathcal{P} \in \mathbb{R}^{N \times 3}$. Then we construct a graph by selecting the \textit{k-nearest neighbors} (k-NN) of each point with respect to the Euclidean distance in the feature space. We attain the local neighborhood information using a feature mapping function $\phi_{c}(\cdot)$ over the point features $\left(f_{i}, f_{j}\right)$:
\begin{equation}
g_{i j c}=\phi_{c}\left(\Delta f_{i j}\right), j \in N(i),   
\end{equation}
where $c = 1, 2,\cdots, C$. $\phi_{c}(\cdot)$ can be implemented as a \emph{shared MLP}, and $N(i)$ denotes the neighborhood of the feature $f_i$.
In order to combine the global shape structure and
the local neighborhood information~\cite{wang2019dynamic}, we define $\Delta f_{i j}=\left[f_{j}-f_{i}, f_{i}\right]$ as the input of $\phi_{c}(\cdot)$, where $\left[\cdot , \cdot \right]$ represents the \emph{concatenation operation}. Then we stack  $g_{ijc}$ of each channel to yield the local information feature $g_{i j}= \left[g_{i j 1}, g_{i j 2}, \ldots, g_{i j C}\right] \in \mathbb{R}^{C}$. Finally, we output the local neighborhood information for each input feature $f_{i}$:
\begin{equation}
f_{i}^{\prime}=\max _{j \in {N}(i)} g_{i j}.
\end{equation}
Moreover, we design an adaptive module to integrate the local patch information and the point feature. Inspired by the high efficiency of SENet~\cite{hu2018squeeze} with simple and lightweight channel attention mechanism, we design an adaptive module based on the attention mechanism between the point feature $\mathcal{F}=\left\{f_{i} \mid i=1,2, \ldots, N\right\} \in \mathbb{R}^{N \times C}$ and the local neighborhood information $\mathcal{F}^{\prime}=\left\{f_{i}^{\prime}\mid i=1,2, \ldots, N\right\} \in \mathbb{R}^{N \times C}$. The adaptive module is presented in \cref{fig:attention},  in which the scale is
 \begin{equation}
\mathbf{s}_{\mathcal{F}}=\operatorname{Sigmoid}\left(\phi\left(\operatorname{AvgPool} \left(\mathcal{F}^{\prime}+\mathcal{F}\right)\right)\right),
\end{equation}
where $+$ is \emph{element-wise sum operation}, and $\phi \left(\cdot \right)$ represents the feature mapping function implemented by the MLP. Finally, we attain the output adaptive feature $\Bar{\mathcal{F}}$ as
 \begin{equation}
\Bar{\mathcal{F}}=\mathbf{s}_{\mathcal{F}} \odot \mathcal{F} + \mathbf{s}_{\mathcal{F}^{\prime}} \odot \mathcal{F}^{\prime},
\end{equation}
where $\mathbf{s}_{\mathcal{F}^{\prime}}= \left(\mathbf{1}-\mathbf{s}_{\mathcal{F}}\right)$, and $\odot$ is the \emph{Hadamard product}. 

 \begin{figure}[t]
  \centering
   \includegraphics[scale=0.48]{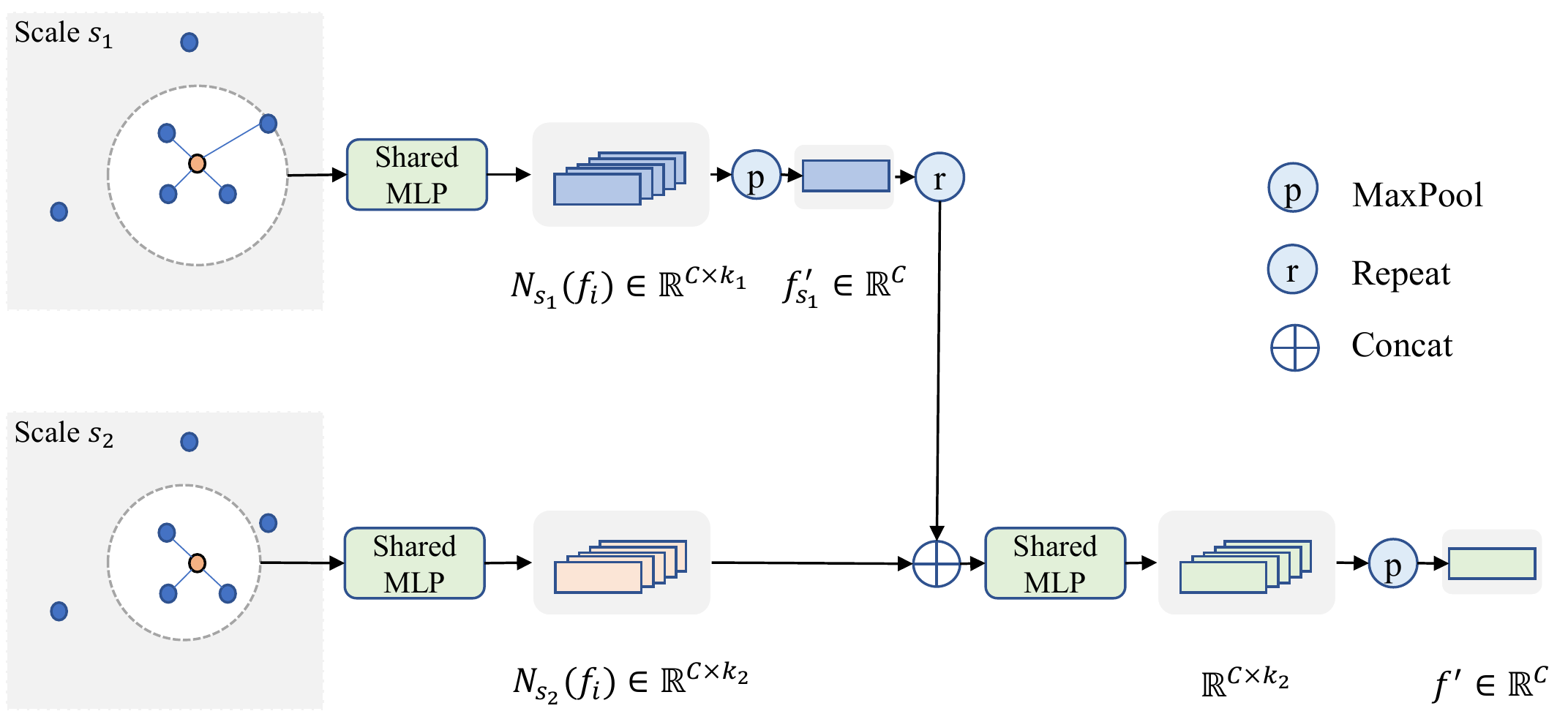}
   \caption{\textbf{Overview of the proposed multi-scale layer}. It is employed to extract richer local features.
   }
   \label{fig:multi-scale}
\end{figure}
\noindent{\textbf{{Multi-scale representation.}}} To extract more distinguishable local features, we also introduce a novel layer to enhance the multi-scale representation ability of the designed network. As shown in \cref{fig:multi-scale},
for a central feature $f_{i} \in \mathbb{R}^{C}$, we first construct a graph by selecting the $k_{1}$-nearest neighbors in the feature space, which indicates the scale $s_{1}$. After the operation mentioned above, we get the output of the local neighborhood feature $f_{i_{s_{1}}}^{\prime} \in \mathbb{R}^{C}$ as illustrated in the top branch of \cref{fig:multi-scale}. Then we select the scale $s_{2}$ including $k_{2}$-nearest neighbors in the feature space, where we usually set $s_{2} < s_{1}$. As observed in the bottom branch of \cref{fig:multi-scale}, the output local neighbor feature $N_{s_{2}}\left(f_{i}\right) \in \mathbb{R}^{C\times k_{2}}$ effectively integrates the feature $f_{i_{s_{1}}}^{\prime}$ of the scale $s_{1}$, and the channel-wise max-pooling function is defined as \begin{equation}
f_{i}^{\prime}=\operatorname{MaxPool} \left(\phi\left(\left[N_{s_{2}}\left(f_{i}\right), f_{i_{s_{1}}}^{\prime}\right]\right)\right),
\end{equation}
where $\phi$ is the MLP, and $\left[\cdot, \cdot \right]$ represents the concatenation operation. The use of multi-scale layer to represent features is of great importance, as it extracts much richer local features, enabling a more stable and robust prediction of weights and offsets.

\noindent{\textbf{{Loss functions.}}} Similar to~\cite{ben2020deepfit}, we use the \emph{angle loss} and the \emph{consistency loss} to train our proposed network. The angle loss measures the deviations of the ground truth normals and the estimated ones, while the consistency loss targets to constrain points that locate on the fitting surface. Furthermore, we regularize the transformation matrix for much easier optimization. Our objective function is defined as 
\begin{equation}
\mathcal{L}_{\text{tol}}=\left|\mathbf{n}_{gt} \times  \hat{\mathbf{n}}\right|+\mathcal{L}_{\text {con }}+\lambda_{3} \mathcal{L}_{\text {reg}},
\end{equation}
where $\lambda_{1}$, $\lambda_{2}$ and $\lambda_{3}$ are the weights to trade off different losses, and 
\begin{equation}
\mathcal{L}_{\text {con }}=\frac{1}{N_{p_{i}}}\left[-\lambda_{1} \sum_{j=1}^{N_{p_{i}}} \log \left(w_{j}\right)+\lambda_{2} \sum_{j=1}^{N_{p_{i}}} w_{j}\left|\mathbf{n}_{gt, j} \times \hat{\mathbf{n}}_{j}\right|\right], \quad \mathcal{L}_{\text{reg}}=\left|I-A A^{T}\right|.
\end{equation}


\section{Experimental Evaluation}\label{sec:exp}
\noindent{\textbf{{Implementation details.}}} We train our proposed network on the benchmark PCPNet dataset~\cite{guerrero2018pcpnet} including eight models: four CAD objects and four high-quality scanning figurines, whereas the test set has 19 different models. More details of the used models are provided in the supplemental material. To be fair, we adopt the same training and evaluation setup as PCPNet. Our network is trained on 32,768 (1,024 samples by 32 shapes) random subsets of the 3.2 M training samples at each epoch. The training process is implemented with PyTorch on Nvidia Tesla V100 GPU, using the Adam optimizer~\cite{kingma2014adam} with the batch size and the learning rate equal to 256 and $1e\mbox{-}3$, respectively. We run the training for 600 epochs totally. The polynomial order $n$ for the surface fitting is 3.

\noindent{\textbf{Evaluation criteria.}}
\begin{table*}[t]
\caption{RMSE comparison for unoriented normal estimation to traditional  (PCA~\cite{hoppe1992surface} and Jet~\cite{cazals2005estimating}) and learning-based methods on the PCPNet dataset. }
\begin{center}
\resizebox{\textwidth}{!}{ 
\begin{tabular}{@{}lcccccccc@{}}
\toprule
Aug.                     & Ours           & AdaFit & DeepFit & IterNet & Nesti-Net & PCPNet & Jet  & PCA   \\ \midrule
w/o Noise     & \textbf{4.45}  & 5.19   & 6.51    & 6.72    & 6.99      & 9.62   & 12.25 & 12.29 \\
$\sigma$ = 0.125\%          & \textbf{8.74}  & 9.05   & 9.21    & 9.95    & 10.11     & 11.37  & 12.84 & 12.87 \\
$\sigma$ = 0.6\%            & \textbf{16.05} & 16.44  & 16.72   & 17.18   & 17.63     & 18.87  & 18.33 & 18.38 \\
$\sigma$ = 1.2\%           & \textbf{21.64} & 21.94  & 23.12   & 21.96   & 22.28     & 23.28  & 27.68 & 27.5  \\
Gradient & \textbf{5.22}  & 5.90   & 7.31    & 7.73    & 9.00      & 11.70   & 13.13 & 12.81 \\
Striped    & \textbf{5.48}  & 6.01   & 7.92    & 7.51    & 8.47      & 11.16  & 13.39 & 13.66 \\
Average                  & \textbf{10.26} & 10.76  & 11.80    & 11.84   & 12.41     & 14.34  & 16.29 & 16.25 \\ \bottomrule
\end{tabular}
}
\label{tab:quan1}
\end{center}

\end{table*}
To assess the performance of the proposed method, we compare it with two types of representative state-of-the-art approaches: $1)$ the geometric methods PCA~\cite{hoppe1992surface} and n-jets~\cite{cazals2005estimating}; $2)$ deep-learning-based methods including PCPNet~\cite{guerrero2018pcpnet}, Nesti-Net~\cite{ben2019nesti}, IterNet~\cite{lenssen2020deep}, DeepFit~\cite{ben2020deepfit}, and AdaFit~\cite{zhu2021adafit}.
For Nesti-Net~\cite{ben2019nesti}, the mixture of experts model is used to obtain normals. Suppose the estimated normal set of the point cloud $\mathcal{P}$ is $\mathcal{N}(\mathcal{P})=\{\hat{\mathbf{n}}_i \in\mathbb{R}^3\}_{i=1}^{N_{\mathcal{P}}}$, then we use the \emph{root-mean-squared error} (RMSE) of angles between the predicted and the ground truth normals $\hat{\mathbf{n}}_i$ and $\mathbf{n}_i$ to evaluate the performance:
\begin{equation}
\operatorname{RMSE}(\mathcal{N}(\mathcal{P}))=\sqrt{\frac{1}{N_{\mathcal{P}}} \sum_{i=1}^{N_{\mathcal{P}}} \arccos^2\left(\hat{\mathbf{n}}_i,\mathbf{n}_i\right)},
\end{equation}
where $(\cdot , \cdot)$ is the inner product of two vectors.
Additionally, we use the metric of the percentage of good points $\operatorname{PGP}(\alpha)$ with an angle tolerance equal to $\alpha$ to report more detailed evaluation:
\begin{equation}
\operatorname{PGP}(\alpha)=\frac{1}{N_{\mathcal{P}}} \sum_{{i}=1}^{N_{\mathcal{P}}} \mathcal{I}\left( \arccos\left(\hat{\mathbf{n}}_i,\mathbf{n}_i\right) < \alpha \right),
\end{equation} where $\mathcal{I}$ is an \textit{indicator function}.
\begin{figure}[t]
  \centering
  \begin{subfigure}{0.3\linewidth}
    \includegraphics[scale=0.22]{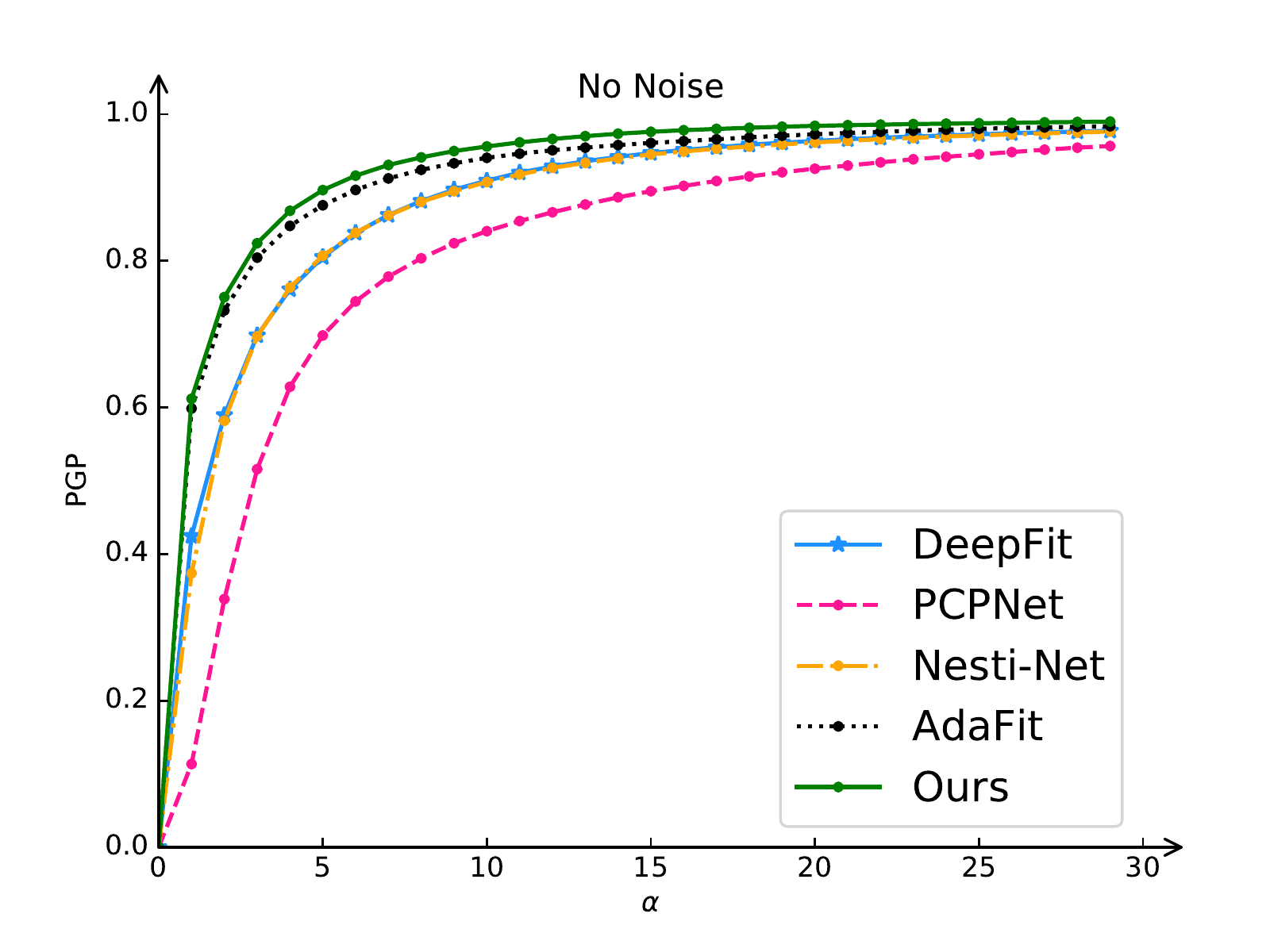}
  \end{subfigure}
  \begin{subfigure}{0.3\linewidth}
    \includegraphics[scale=0.22]{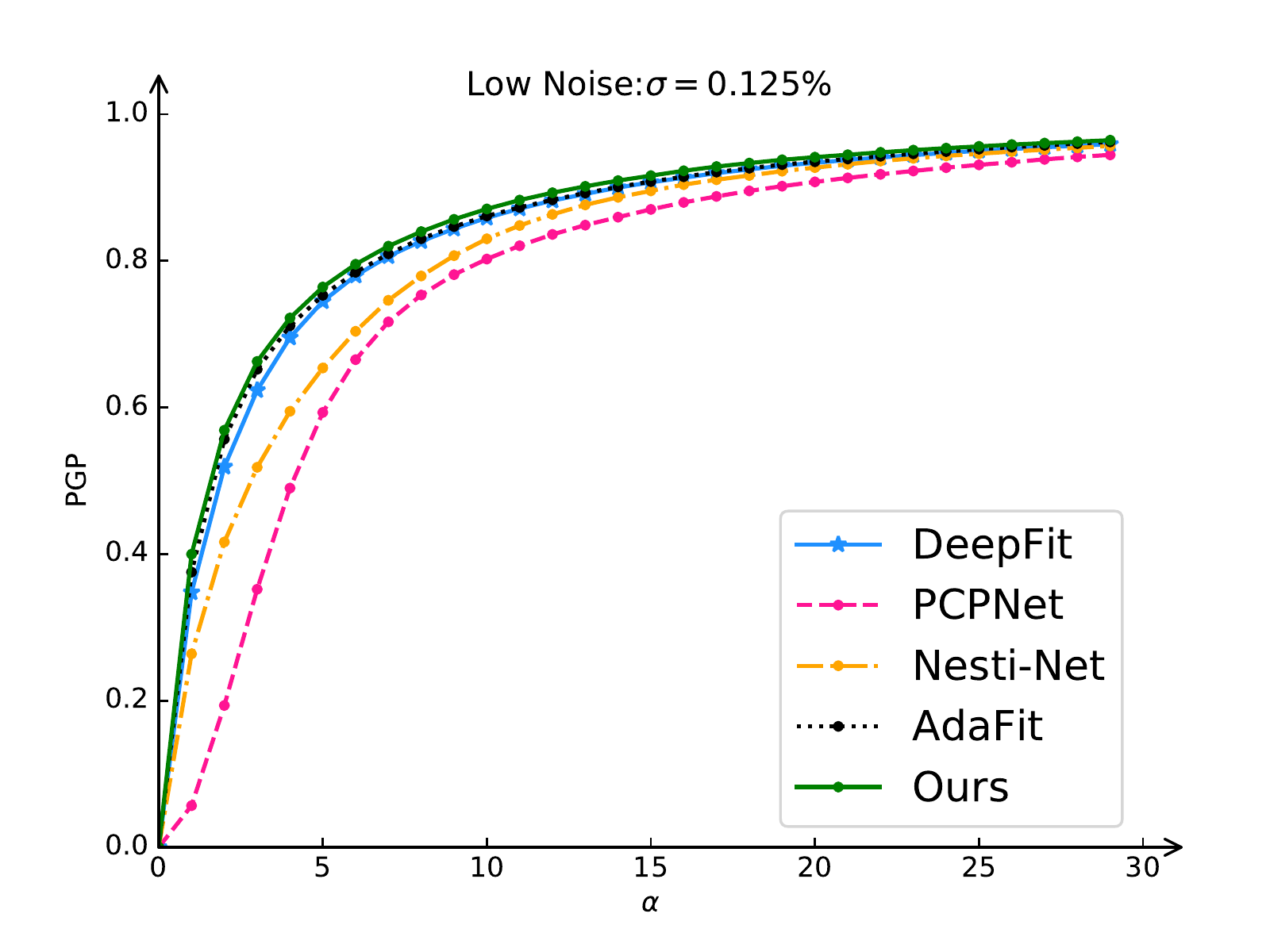}
  \end{subfigure}
  \begin{subfigure}{0.3\linewidth}
    \includegraphics[scale=0.22]{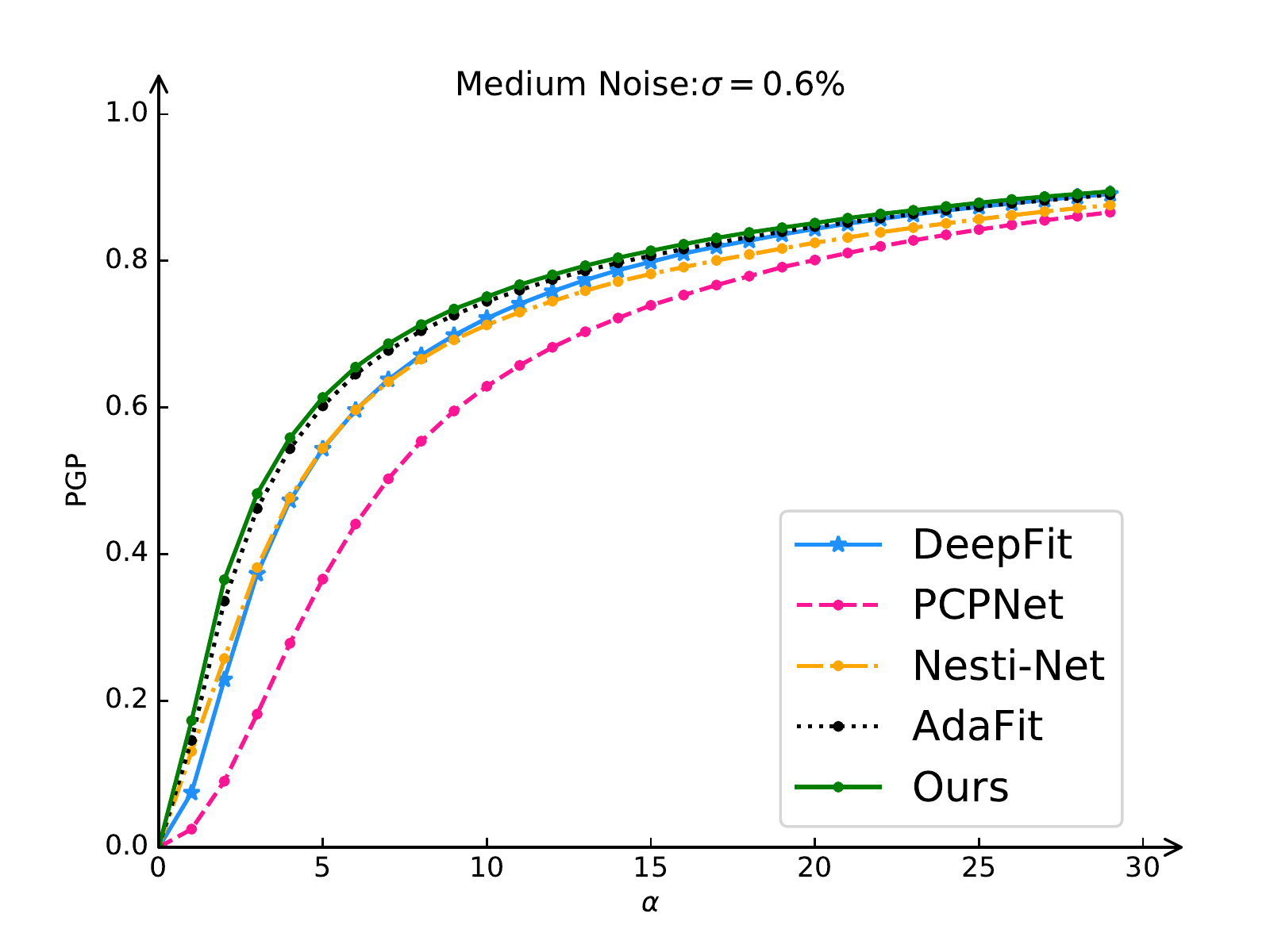}
  \end{subfigure}
  \begin{subfigure}{0.3\linewidth}
    \includegraphics[scale=0.22]{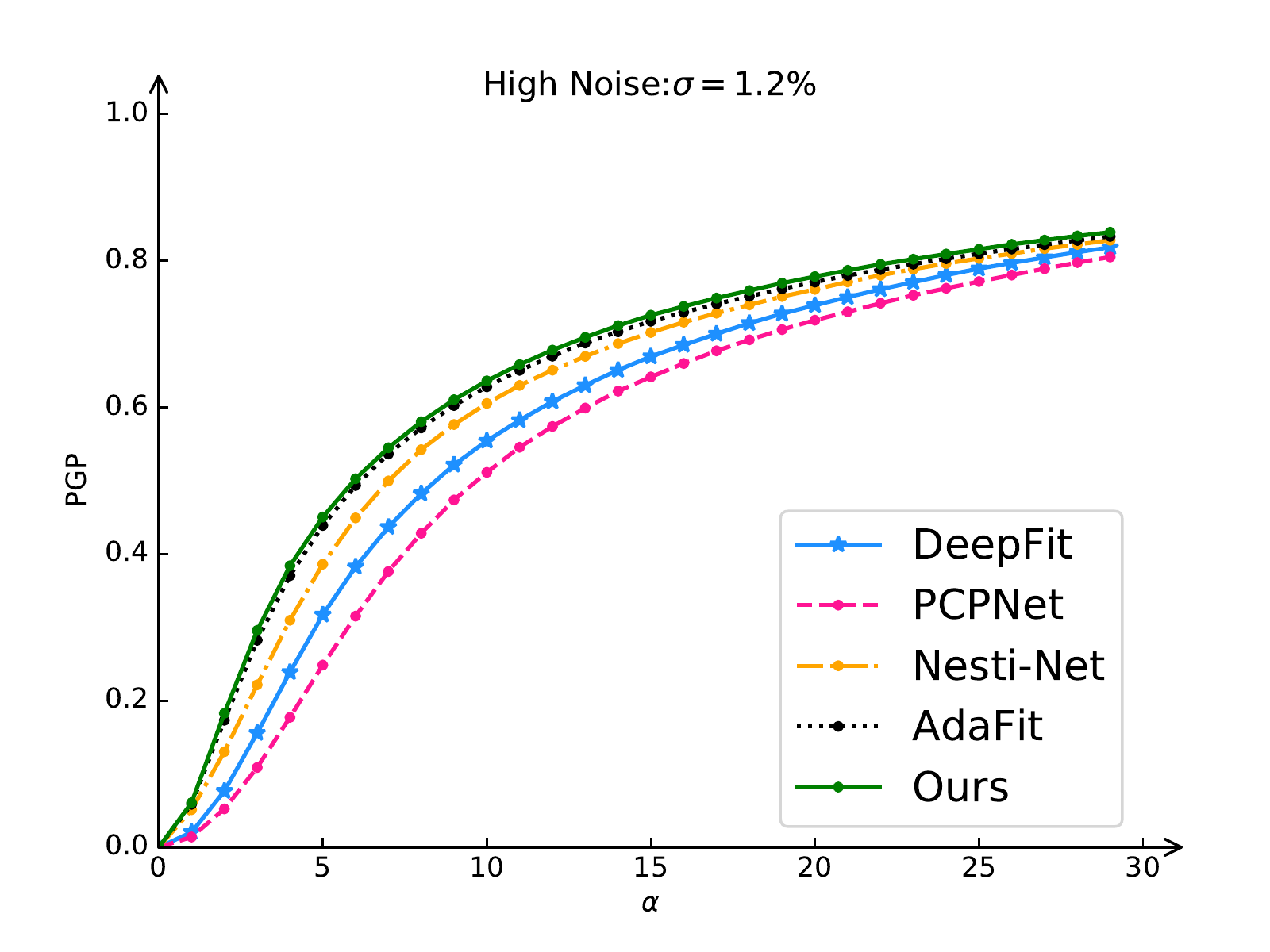}
  \end{subfigure}
  \begin{subfigure}{0.3\linewidth}
    \includegraphics[scale=0.22]{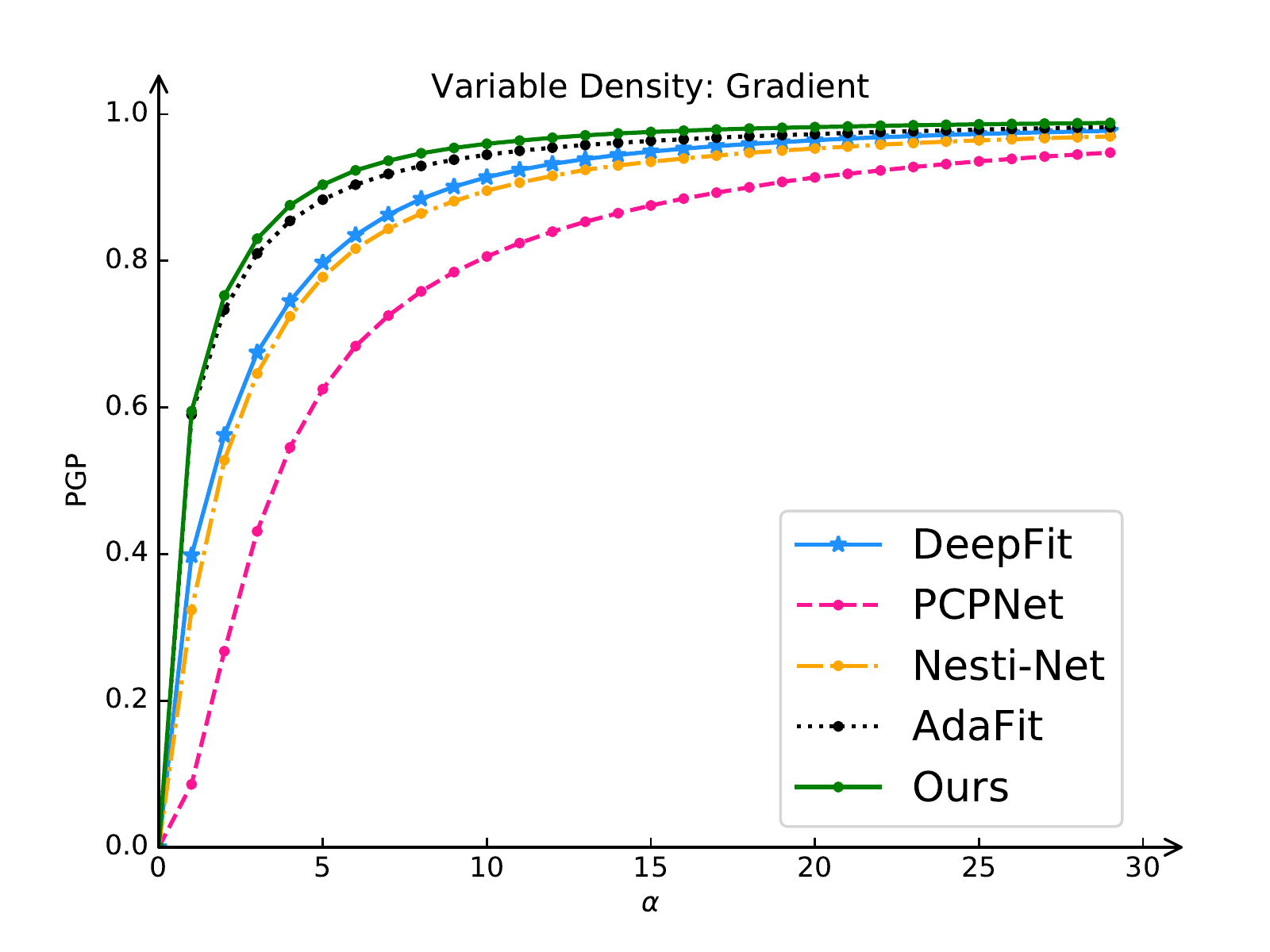}
  \end{subfigure}
  \begin{subfigure}{0.3\linewidth}
    \includegraphics[scale=0.22]{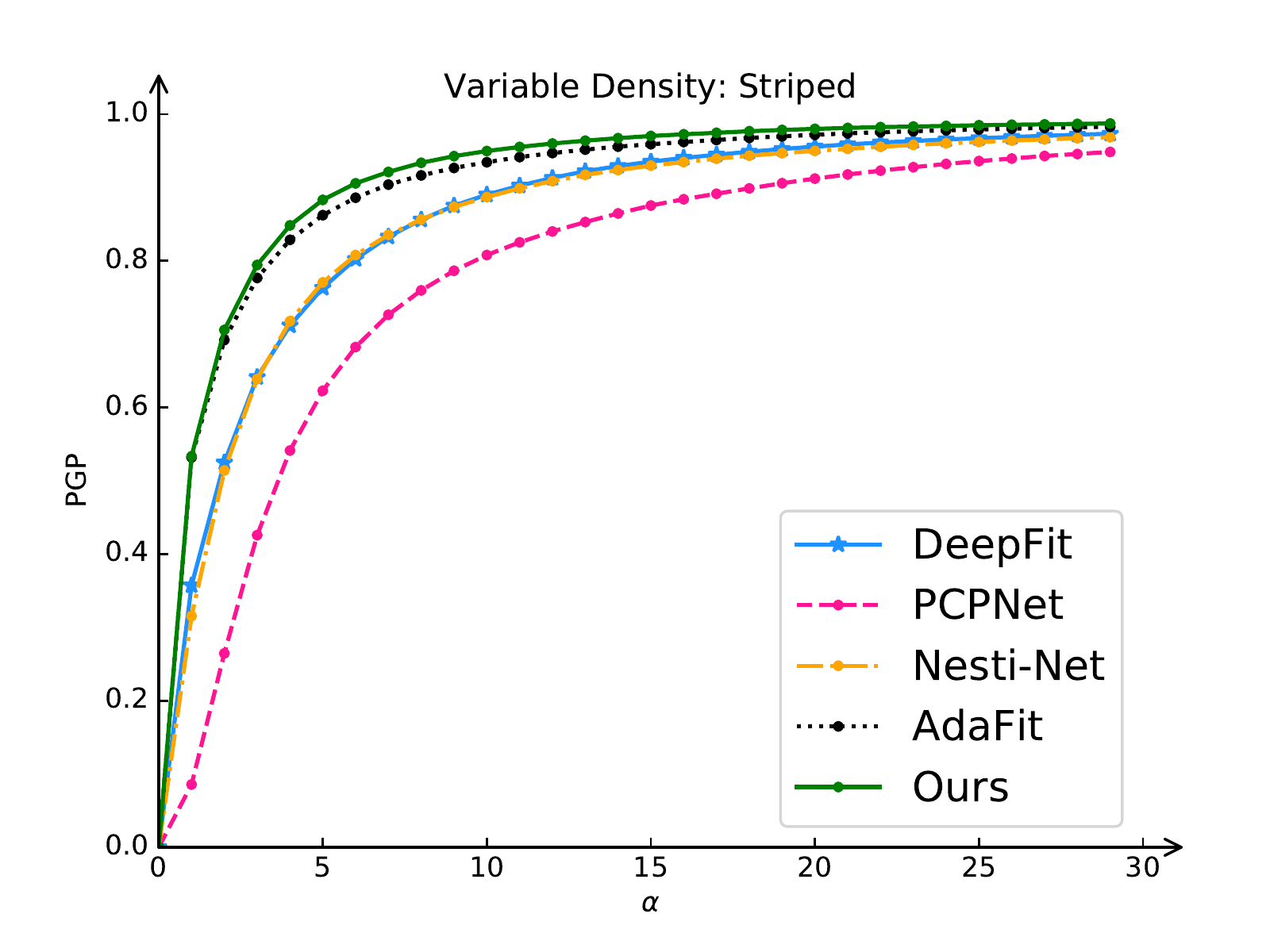}
  \end{subfigure}
  \caption{Comparison of the proposed method with state-of-the-art learning-based approaches regarding the percentage of good points $\left(PGP \alpha \right)$.
}
\label{fig:pgp}
\end{figure}

\noindent{\textbf{{Synthetic test.}}}
We conduct a series of quantitative experiments to compare the performance of all methods on the PCPNet dataset~\cite{guerrero2018pcpnet}, in which the point clouds are contaminated by different noise levels with varying standard deviation $\sigma$. Besides, we perform experiments on two different types of point density to simulate the effects of distance from the sensor (Gradient) and the local occlusions (Striped). The results are reported in \cref{tab:quan1}. As observed, the proposed method attains the highest accuracy under different noise levels and varying point density, demonstrating the effectiveness of our proposed graph-structure-based normal estimation. AdaFit shows satisfactory results but it is relatively sensitive to density variations. Compared with learning-based approaches, traditional methods including Jet and PCA show more deviations.

\begin{figure}[t]
  \centering
  \begin{subfigure}{0.485\linewidth}
    \includegraphics[scale=0.4]{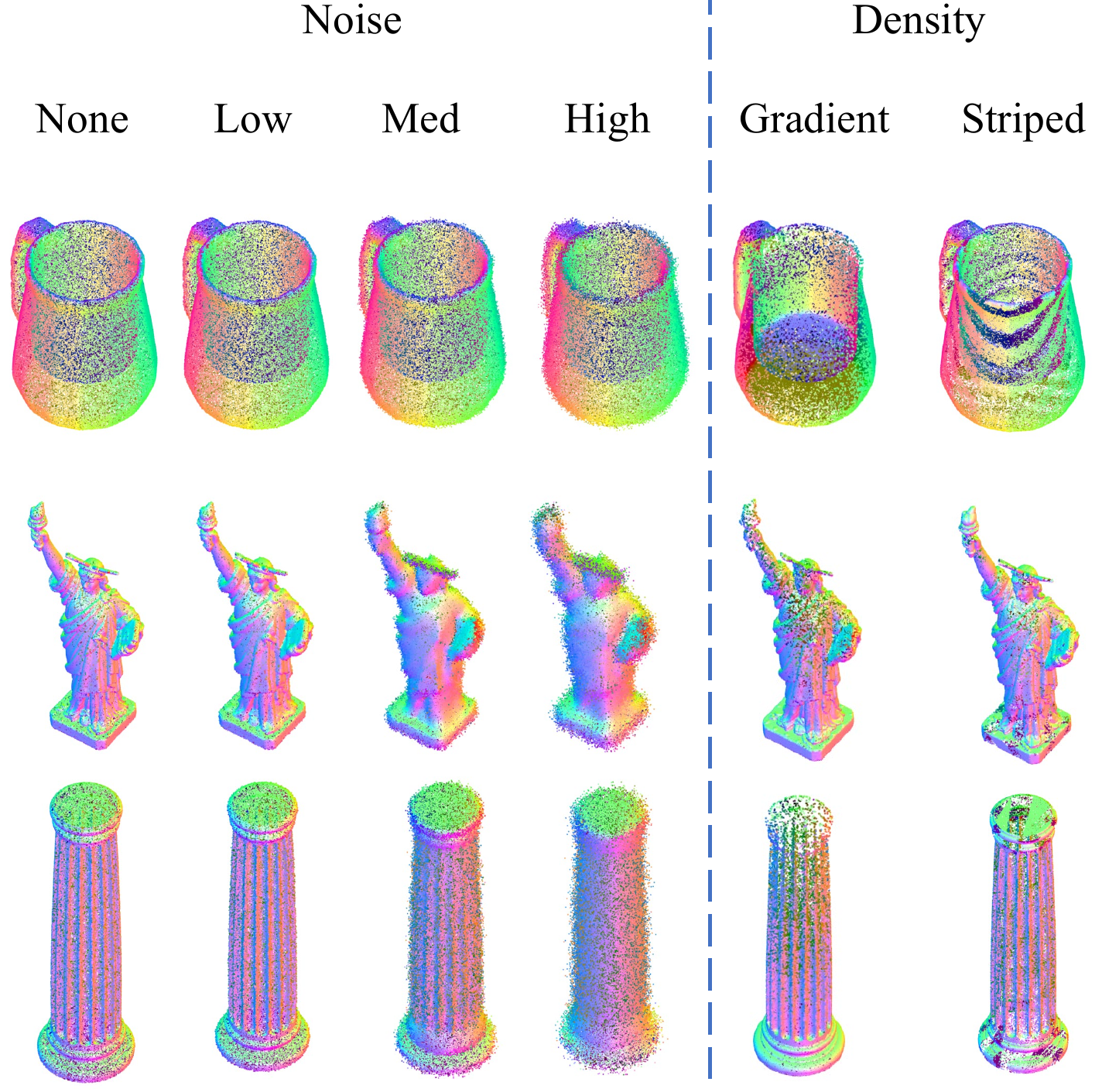}
  \end{subfigure}
  \begin{subfigure}{0.493\linewidth}
    \includegraphics[scale=0.38]{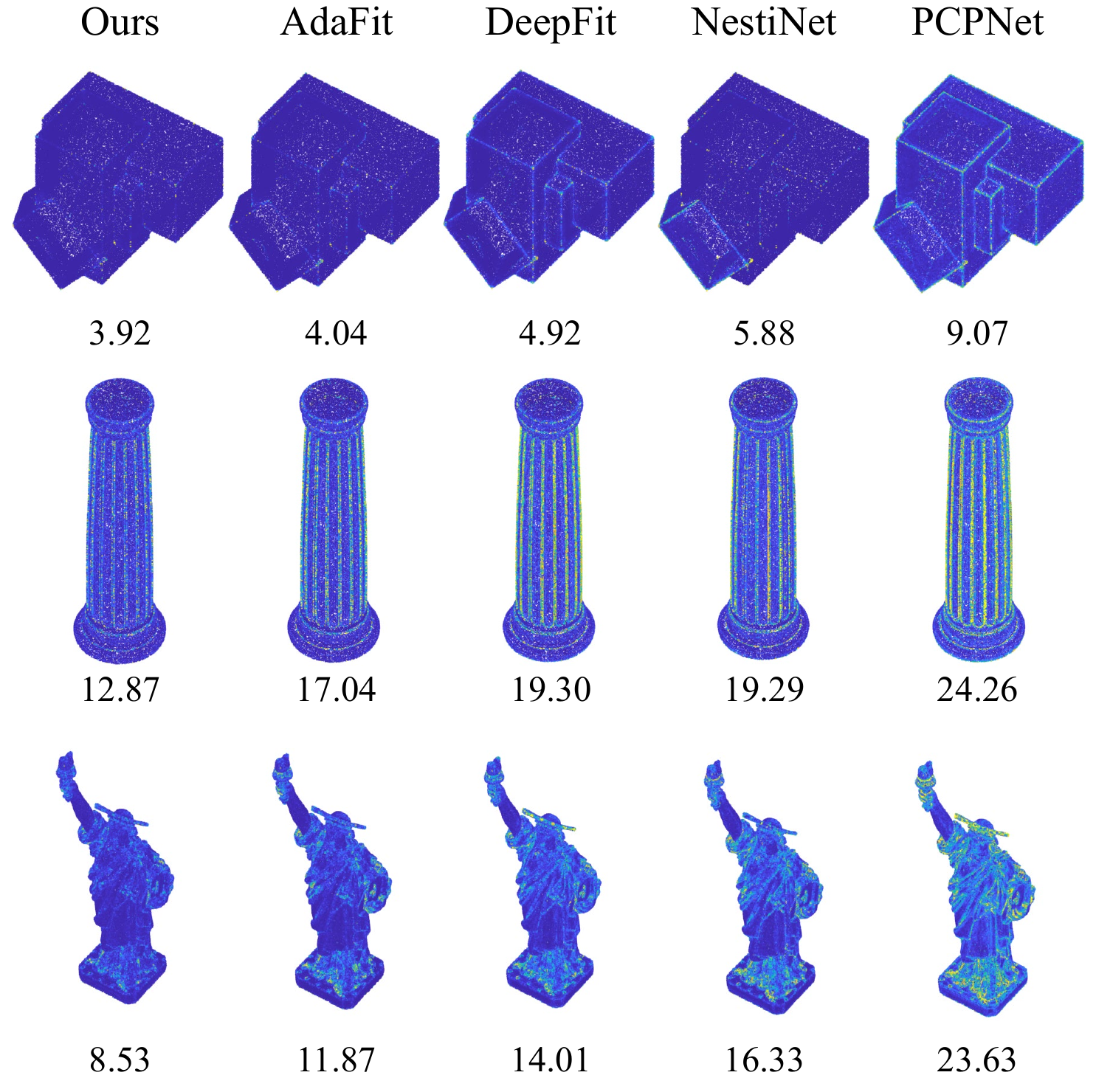}
  \end{subfigure}
    \caption{Left: Normal estimation under four different
noise levels (Columns one--four), and varying point density (Columns
five--six). We map the normals to RGB for easy visualization. Right: Illustration of the normal estimation errors for three different point clouds. The errors are mapped to a heatmap ranging from $0^\circ$ to $60^\circ$. Quantities under the point clouds are the corresponding RMSE.}
    \label{fig:quali1}
\end{figure}

We further evaluate the normal estimation performance on the PCPNet using the percentage of good points $\left(PGP \alpha \right)$ metric. The results are reported in \cref{fig:pgp}. It can be noted that the proposed
method has the overall best performance under different test settings, meaning that it achieves more accurate normal estimation. The normal prediction results of our method is visualized in the left panel of \cref{fig:quali1}. The right panel of \cref{fig:quali1} exhibits the angular error in each test point cloud for all methods, where it can be seen that our proposed method obtains lower RMSE. Besides, it is quite robust against challenging regions, such as sharp edges and corners.

\begin{table}[t]
\centering
\caption{Statistics of the angle RMSE on the real-world SceneNN dataset~\cite{hua-pointwise-cvpr18}.}
\setlength{\tabcolsep}{1.5mm}{
\begin{tabular}{@{}ccccc@{}}
\toprule
 \textbf{Method} &Ours & AdaFit & DeepFit& PCPNet \\ \midrule
\textbf{Average} & \textbf{21.33} & 22.61  & 24.59   & 27.27  \\ \bottomrule
\end{tabular}}
\label{tab:scenenn}
\end{table}

\begin{figure}[t]
  \centering
  \begin{subfigure}{0.19\linewidth}
    \includegraphics[scale=0.08]{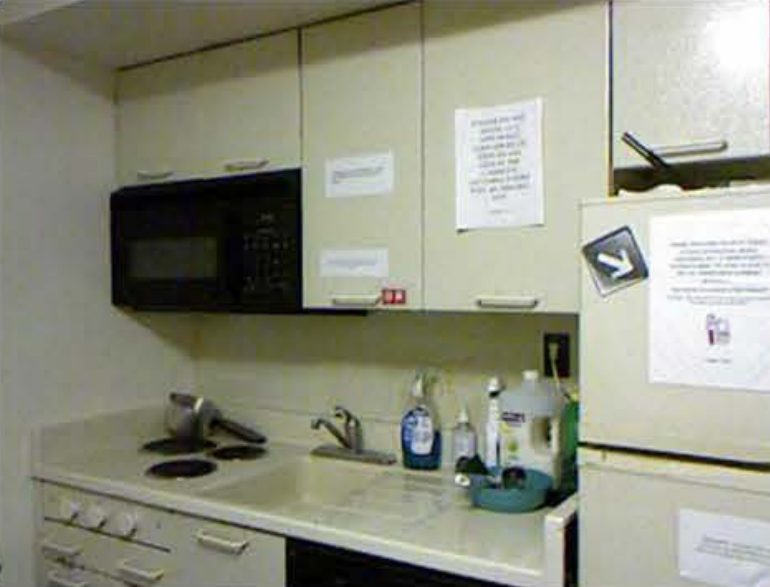}
  \end{subfigure}
  \begin{subfigure}{0.19\linewidth}
    \includegraphics[scale=0.15]{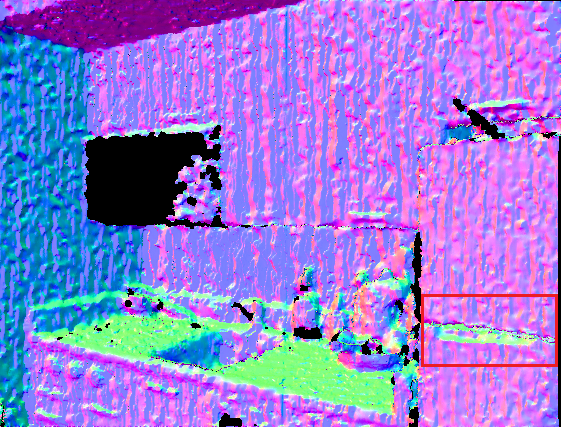}
  \end{subfigure}
  \begin{subfigure}{0.19\linewidth}
    \includegraphics[scale=0.15]{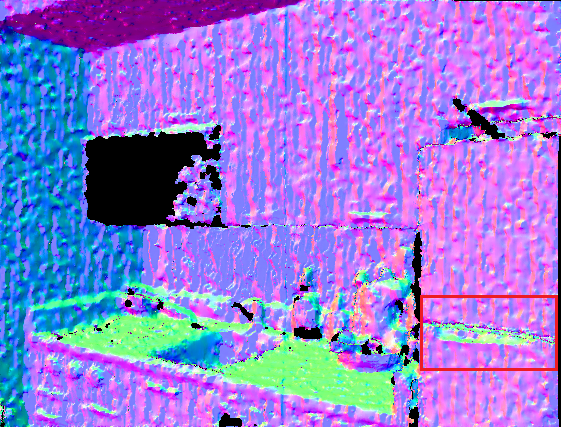}
  \end{subfigure}
  \begin{subfigure}{0.19\linewidth}
    \includegraphics[scale=0.15]{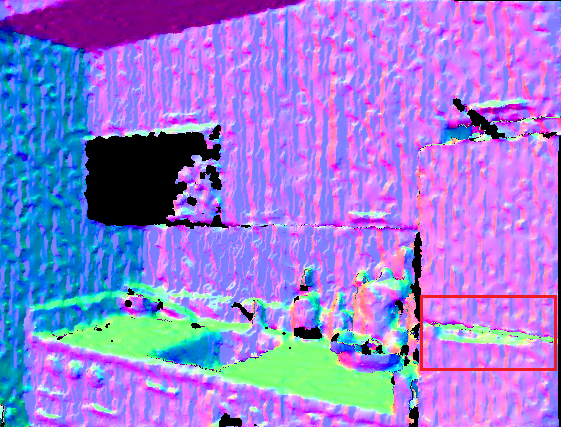}
  \end{subfigure}
  \begin{subfigure}{0.19\linewidth}
    \includegraphics[scale=0.15]{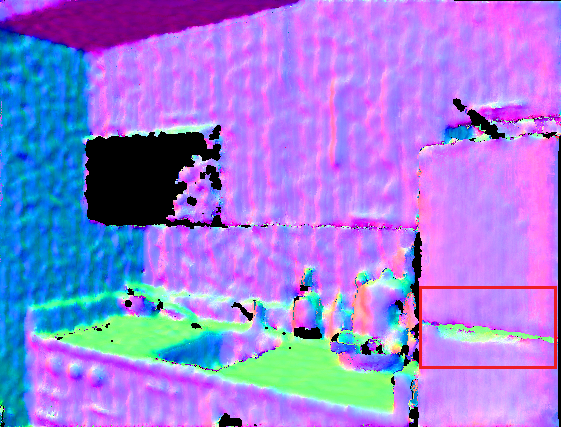}
  \end{subfigure}
    \begin{subfigure}{0.19\linewidth}
    \transparent{0.001}\includegraphics[scale=0.07]{nyu/nyu2.png}
  \end{subfigure}
  \begin{subfigure}{0.19\linewidth}
    \includegraphics[scale=0.6]{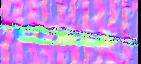}
    \caption{Ours}
  \end{subfigure}
  \begin{subfigure}{0.19\linewidth}
    \includegraphics[scale=0.6]{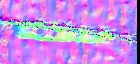}
    \caption{AdaFit}
  \end{subfigure}
  \begin{subfigure}{0.19\linewidth}
    \includegraphics[scale=0.6]{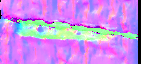}
    \caption{DeepFit}
  \end{subfigure}
  \begin{subfigure}{0.19\linewidth}
    \includegraphics[scale=0.6]{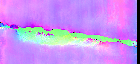}
    \caption{PCPNet}
  \end{subfigure}
  \caption{Visualization of the normal estimation result on the NYU Depth V2 dataset. The second row is the zoom-in images of the red box in the first row. Our method embraces better generalization to this dataset while retains more details and sharp edges than others, yet the scanning artifacts are also kept and visible. We map normals to RGB and then project them to the image.
}
  \label{fig:nyu}
\end{figure}

\begin{figure}[]
  \centering

  \begin{subfigure}{0.19\linewidth}
    \includegraphics[scale=0.13]{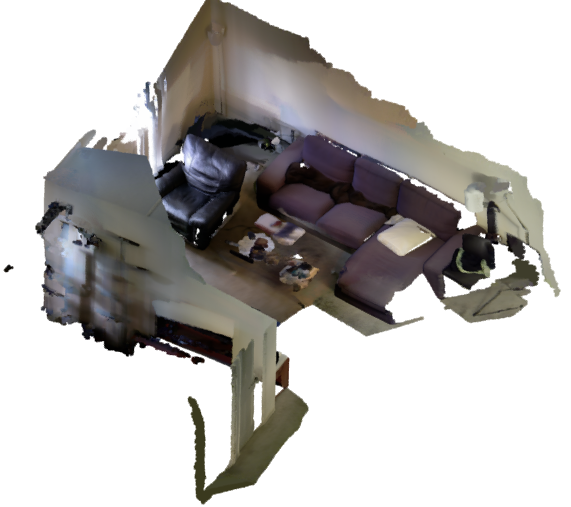}
    \caption{Scene}
  \end{subfigure}
       \begin{subfigure}{0.19\linewidth}
    \includegraphics[scale=0.2]{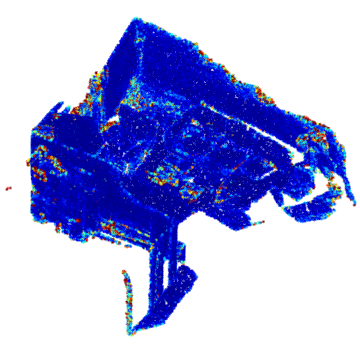}
    \caption{Ours}
  \end{subfigure}
  \begin{subfigure}{0.19\linewidth}
    \includegraphics[scale=0.2]{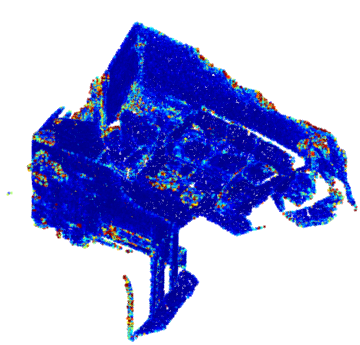}
    \caption{AdaFit}
  \end{subfigure}
  \begin{subfigure}{0.19\linewidth}
    \includegraphics[scale=0.2]{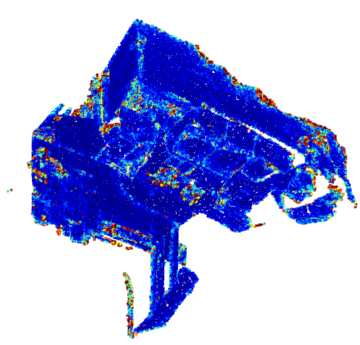}
    \caption{DeepFit}
  \end{subfigure}
  \begin{subfigure}{0.19\linewidth}
    \includegraphics[scale=0.2]{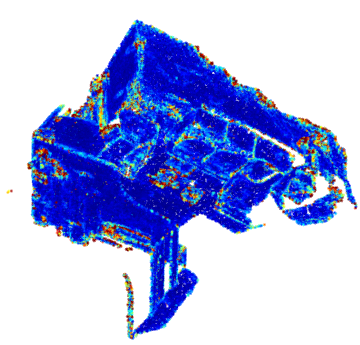}
    \caption{PCPNet}
  \end{subfigure}
  \caption{Normal estimation for a real-world scene.
  The errors are mapped to a heatmap ranging from $0^\circ$ to $60^\circ$. For (b)-(e), the normal errors are \textbf{26.03}, 27.08, 28.86 and 30.83, respectively.
}
  \label{fig:quali2}
\end{figure}
\begin{table}[t]
\begin{center}
\caption{Complexity comparison of learning-based normal estimators.}
\begin{tabular}{@{}cccc@{}}
\toprule
\multicolumn{1}{l}{\textbf{Aug.}}   & params(M) & Model size(MB) & Avg error             \\ \midrule
\multicolumn{1}{l}{PCPNet} & 21.30           & 85.41    & 14.56                     \\
Nesti-Net              & 170.10          & 2,010.00     & 12.41                     \\
DeepFit                  & \textbf{3.36}          & \textbf{13.53}    & 11.80                      \\
AdaFit                  & 4.36           & 18.74    & 10.76                     \\
Ours                       & 4.06           & 16.38    & \textbf{10.26} \\ \bottomrule
\end{tabular}

\label{tab:eff}
\end{center}
\end{table}

\begin{table}[h]
\caption{Left: RMSE comparison of the proposed method with DGCNN~\cite{wang2019dynamic} on the PCPNet dataset. Right: Effect of the adaptive module.}
\begin{center}
\begin{subtable}[t]{0.42\linewidth}
{
\centering
\resizebox{0.9\textwidth}{0.3\textwidth}{
\begin{tabular}{@{}ccccccccc@{}}
\toprule
\textbf{Aug.}                     & Ours           & DGCNN  \\ \midrule
w/o Noise                 & \textbf{4.49}  & 5.47   \\
$\sigma$ = 0.125\%          & \textbf{8.80} & 8.90   \\
$\sigma$ = 0.6\%            & \textbf{16.54} & 16.57 \\
$\sigma$ = 1.2\%            & \textbf{22.69} & 22.85  \\
Gradient & \textbf{5.15}  & 6.19   \\
Striped    & \textbf{5.28}  & 6.52  \\
Average                  & \textbf{10.49} & 11.08  \\ \bottomrule
\end{tabular}}
\label{tab:dgcnn}}
\end{subtable}
\begin{subtable}[t]{0.43\linewidth}{
\centering
\resizebox{1\textwidth}{0.292\textwidth}{
\begin{tabular}{@{}cccc@{}}
\toprule
\textbf{Aug.}               & \multicolumn{2}{c}{Graph-block}         \\ \midrule
 with adaptive module                  &  & $\color{red}\checkmark$ &       \\
w/o Noise           & 4.87                & \textbf{4.56}               \\
$\sigma$ = 0.125\% & 8.94                & \textbf{8.87}               \\
$\sigma$ = 0.6\%   & 16.58               & \textbf{16.57}            \\
$\sigma$ = 1.2\%   & 22.83               & \textbf{22.77}             \\
Gradient           & 5.60                & \textbf{5.33}               \\
Striped              & 5.77                & \textbf{5.38}               \\
Average            & 10.74               & \textbf{10.58}           \\ \bottomrule
\end{tabular}
}
\label{tab:graph}
}
\end{subtable}
\end{center}
\end{table}

\noindent{\textbf{{Real-world test.}}} We also assess the effectiveness of our method on real-world datasets to demonstrate its generalization and stability. We adopt the NYU Depth V2 dataset~\cite{silberman2012indoor} for test. It contains 1,449 aligned and preprocessed RGBD frames, which are transformed to  point clouds before applying our method. Note that all compared methods are only trained once on the PCPNet dataset. Compared with synthetic data, normal estimation of real-world scanning data are more challenging due to the occlusion and varying noise patterns. In particular, the noise often has the same magnitude as some of the features. Like most real datasets, there is no ground truth for each point. 

As shown in \cref{fig:nyu}, our proposed method is comparable to AdaFit on preserving fine details, meanwhile its performance is better than other compared approaches that typically result in over-smoothing results. However, to a certain extent, this also leads to the sharp extraction of scanning artifacts, as seen on the walls of the scanned room and the refrigerator surface in \cref{fig:nyu}.

We further validate the proposed model on the SceneNN dataset~\cite{hua-pointwise-cvpr18}, which contains 76 scenes re-annotated with 40 NYU-D v2 classes collected by a \emph{depth camera} with ground-truth reconstructed meshes. We obtain the sampled point clouds and compute ground-truth normals from the meshes. The statistical angle RMSE of all methods are reported in \cref{tab:scenenn}. 
Thanks to the integration of more local neighbor features, our method significantly outperforms all competitors and achieves the state-of-the-art performance. We visualize the normal error of a random scene in \cref{fig:quali2}.

\noindent{\textbf{{Computational complexity.}}} We further compare the complexity of our model with state-of-the-art approaches, where the number of parameters of each method, their model size, and the average RMSE are reported. \cref{tab:eff} indicates that 
DeepFit has the smallest computational complexity, but its RMSE is relatively large. Instead, the proposed method has the lowest average error along with comparable complexity of DeepFit, thereby it achieves a good balance between accuracy and complexity.

\noindent{\textbf{{Comparison with DGCNN.}}} We compare our network with the baseline backbone DGCNN~\cite{wang2019dynamic}, which can be seen as a standard graph convolution version of our proposed method. To be fair, we train DGCNN on the PCPNet dataset to predict point-wise weights and offsets. The neighborhood size and the jet order of the two methods are the same and equal to 256 and 3, respectively. The left panel of \cref{tab:dgcnn} summarizes the comparison results, where our method outperforms DGCNN in various settings, such as different noise levels and varying point density, demonstrating the overall advantages of our designed network.

\begin{table}[t]
\begin{center}
\caption{RMSE comparison with different Jet order $n$ on the PCPNet dataset.}
\setlength{\tabcolsep}{2mm}{
\begin{tabular}{@{}ccccccccc@{}}
\toprule
\textbf{Aug.}                      & \multicolumn{2}{c}{$n=1$}                           & \multicolumn{2}{c}{$n=2$}                           & \multicolumn{2}{c}{$n=3$}                           & \multicolumn{2}{c}{$n=4$}                           \\ \midrule
Graph-block               & $\color{red}\checkmark$ & $\color{red}\checkmark$ & $\color{red}\checkmark$ & $\color{red}\checkmark$ & $\color{red}\checkmark$ & $\color{red}\checkmark$ & $\color{red}\checkmark$ & $\color{red}\checkmark$ \\
Multi-scale               &                         & $\color{red}\checkmark$ &                         & $\color{red}\checkmark$ &                         & $\color{red}\checkmark$ &                         & $\color{red}\checkmark$ \\
w/o Noise                  & 5.02                    & 4.60                    & 5.26                    & 4.40                    & 4.56                    & 4.49                    & 4.80                    & 4.34                    \\
$\sigma$ = 0.125\% & 9.15                    & 8.97                    & 9.08                    & 8.76                    & 8.87                    & 8.80                    & 9.00                    & 8.85                    \\
$\sigma$ = 0.6\%   & 16.65                   & 16.59                  & 16.60                   & 16.54                   & 16.57                   & 16.54                   & 16.61                   & 16.48                   \\
$\sigma$ = 1.2\%   & 22.75                   & 22.74                   & 22.88                   & 22.73                   & 22.77                   & 22.69                   & 22.87                   & 22.69                   \\
Gradient & 5.85                    & 5.36                    & 5.74                    & 5.23                    & 5.33                    & 5.15                    & 5.60                    & 5.06                    \\
Striped    & 5.93                    & 5.49                    & 5.95                    & 5.36                    & 5.38                    & 5.28                    & 5.59                    & 5.22                    \\
Average                   & 10.91                   & \textbf{10.63}                   & 10.92                   & \textbf{10.50}                   & 10.58                   & \textbf{10.49}                   & 10.75                   &\textbf{ 10.44}                   \\ \bottomrule
\end{tabular}}
\label{tab:njet}
\end{center}
\end{table}
\noindent{\textbf{{Robustness against the Jet orders.}}} We also investigate the influence of the Jet order for normal estimation in our method. To this end, we set $n=1, 2, 3, 4$, and the neighborhood size is fixed as 256 points. \cref{tab:njet} shows the RMSE on the PCPNet dataset. As observed, with Jet order increasing, our method gradually produces more accurate normal estimation. Even under lower Jet order, the estimated normals are still comparable to DeepFit~\cite{ben2020deepfit} and AdaFit~\cite{zhu2021adafit} as reported in \cref{tab:quan1}, which shows our method is quite robust against Jet orders. \cref{tab:njet} also records the ablation study results of our proposed multi-scale layer, from which we conclude that the multi-scale layer effectively fuses richer geometric features hence assures more accurate normal estimation.

\begin{table}[!htbp]
\caption{Investigation on the neighbor size $k$ on normal estimation.}
\begin{center}
\begin{tabular}{@{}ccccccc@{}}
\toprule
\textbf{Aug.}        & \multicolumn{2}{c}{$k=256$}                           & \multicolumn{2}{c}{$k=500$}                           & \multicolumn{2}{c}{$k=700$}                           \\ \midrule
Graph-block & $\color{red}\checkmark$ & $\color{red}\checkmark$ & $\color{red}\checkmark$ & $\color{red}\checkmark$ & $\color{red}\checkmark$ & $\color{red}\checkmark$ \\
Multi-scale &                         & $\color{red}\checkmark$ &                         & $\color{red}\checkmark$ &                         & $\color{red}\checkmark$ \\
w/o Noise    & 4.56                    & 4.49                    & 4.80                    & 4.45                    & 4.78                    & 4.83                    \\
$\sigma$ = 0.125\%   & 8.87                    & 8.80                    & 8.90                    & 8.74                   & 8.77                    & 8.70                    \\
$\sigma$ = 0.6\%   & 16.57                   & 16.54                   & 16.16                   & 16.05                   & 15.99                   & 16.04                   \\
$\sigma$ = 1.2\%  & 22.77                   & 22.69                   & 21.80                   & 21.64                   & 21.45                   & 21.36                   \\
Gradient    & 5.33                    & 5.15                    & 5.45                    & 5.22                    & 5.50                    & 5.51                    \\
Striped       & 5.38                    & 5.28                    & 5.90                    & 5.48                    & 5.78                    & 5.61                    \\
Average     & 10.58                   & \textbf{10.49}                   & 10.50                   & \textbf{10.26}                   & 10.38                   & \textbf{10.34}                   \\ \bottomrule
\end{tabular}
\label{tab:scale}
\end{center}
\end{table}

\noindent{\textbf{{Robustness against the neighborhood size.}}}
We also conduct experiments on the PCPNet dataset to test the robustness of the proposed method against the neighborhood size $k = 256, 500, 700$. The default Jet order $n=3$ is used. From \cref{tab:scale}, we see that the average RMSE is relatively stable and changes around 10.50. There is no significant fluctuation even under different noise levels and varying point density. Hence we can use $k = 256$ for general normal estimation tasks. Note that AdaFit~\cite{zhu2021adafit} uses a fixed large patch size $\left(k = 700\right)$ for training, which usually consumes more time and computing resource. The ablation study of  the multi-scale layer in \cref{tab:scale} again evidences its advantages. 

\noindent{\textbf{{Effect of the proposed adaptive module.}}} The designed adaptive module on the basis of attention mechanism can effectively improve the normal estimation precision. We implement ablation study to verify this and the results are reported in the right panel of  \cref{tab:graph}, in which the neighborhood size is set as 256 and the jet order is equal to 3. It can be seen that by adding an adaptive module, the network returns more accurate normals, particularly, it effectively reduces the influence of the uneven point density, which can be concluded from the cases of Gradient and Striped.

\section{Conclusion}
\label{sec:conclu}
We presented an accurate and robust pipeline for normal estimation of unstructured 3D point clouds, which achieves state-of-the-art performance compared with competing approaches. Our contribution is highlighting the local neighborhood relationships for normal estimation which are usually neglected by previous methods, meanwhile, we invoke graph convolutional learning to efficiently encode such situation. Moreover, based on the attention mechanism, we introduce an adaptive module on top of the graph block, to effectively combine the point features with their local neighbor features. As demonstrated, this operation significantly improves the network's robustness against point density variations. Together, we leverage the multi-scale layer to extract richer geometric features and consequently enhance the normal estimation precision.

Extensive experiments are conducted on a wide variety of datasets from synthetic to real-world data. Results demonstrate that our method achieves the state-of-the-art performance in accuracy and robustness on the benchmark PCPNet dataset, and shows quite stable generalization ability on the real-world NYU Depth V2 scenes, which suggests its potential as a fast normal estimation technique. In the future, we will customize the proposed method for CAD models, especially for surface reconstruction and denosing from scanned point clouds.


\clearpage
%
%
\bibliographystyle{splncs04}
\bibliography{5419}

\begin{thebibliography}{10}
\providecommand{\url}[1]{\texttt{#1}}
\providecommand{\urlprefix}{URL }
\providecommand{\doi}[1]{https://doi.org/#1}

\bibitem{alliez2007voronoi}
Alliez, P., Cohen-Steiner, D., Tong, Y., Desbrun, M.: Voronoi-based variational
  reconstruction of unoriented point sets. In: Proceedings of the 5th
  Eurographics Symposium on Geometry Processing. pp. 39--48 (2007)

\bibitem{amenta1999surface}
Amenta, N., Bern, M.: Surface reconstruction by voronoi filtering. Discrete \&
  Computational Geometry.  \textbf{22}(4),  481--504 (1999)

\bibitem{ben2020deepfit}
Ben-Shabat, Y., Gould, S.: {DeepFit}: {3D} surface fitting via neural network
  weighted least squares. In: Proceedings of the European Conference on
  Computer Vision. pp. 20--34 (2020)

\bibitem{ben20183dmfv}
Ben-Shabat, Y., Lindenbaum, M., Fischer, A.: {3DMFV}: Three-dimensional point
  cloud classification in real-time using convolutional neural networks. IEEE
  Robotics and Automation Letters.  \textbf{3}(4),  3145--3152 (2018)

\bibitem{ben2019nesti}
Ben-Shabat, Y., Lindenbaum, M., Fischer, A.: {Nesti-Net}: Normal estimation for
  unstructured {3D} point clouds using convolutional neural networks. In:
  Proceedings of the IEEE/CVF Conference on Computer Vision and Pattern
  Recognition. pp. 10112--10120 (2019)

\bibitem{boulch2012fast}
Boulch, A., Marlet, R.: Fast and robust normal estimation for point clouds with
  sharp features. Computer Graphics Forum.  \textbf{31}(5),  1765--1774 (2012)

\bibitem{boulch2016deep}
Boulch, A., Marlet, R.: Deep learning for robust normal estimation in
  unstructured point clouds. Computer Graphics Forum  \textbf{35}(5),  281--290
  (2016)

\bibitem{castillo2013point}
Castillo, E., Liang, J., Zhao, H.: Point cloud segmentation and denoising via
  constrained nonlinear least squares normal estimates, pp. 283--299 (2013)

\bibitem{cazals2005estimating}
Cazals, F., Pouget, M.: Estimating differential quantities using polynomial
  fitting of osculating jets. Computer Aided Geometric Design.  \textbf{22}(2),
   121--146 (2005)

\bibitem{che2018multi}
Che, E., Olsen, M.J.: Multi-scan segmentation of terrestrial laser scanning
  data based on normal variation analysis. ISPRS Journal of Photogrammetry and
  Remote Sensing  \textbf{143},  233--248 (2018)

\bibitem{comino2018sensor}
Comino, M., Andujar, C., Chica, A., Brunet, P.: Sensor-aware normal estimation
  for point clouds from {3D} range scans. Computer Graphics Forum
  \textbf{37}(5),  233--243 (2018)

\bibitem{dey2006provable}
Dey, T.K., Goswami, S.: Provable surface reconstruction from noisy samples.
  Computational Geometry.  \textbf{35}(1-2),  124--141 (2006)

\bibitem{fan2021scf}
Fan, S., Dong, Q., Zhu, F., Lv, Y., Ye, P., Wang, F.Y.: {SCF-Net}: Learning
  spatial contextual features for large-scale point cloud segmentation. In:
  Proceedings of the IEEE/CVF Conference on Computer Vision and Pattern
  Recognition. pp. 14504--14513 (2021)

\bibitem{fleishman2005robust}
Fleishman, S., Cohen-Or, D., Silva, C.T.: Robust moving least-squares fitting
  with sharp features. ACM Transactions on Graphics.  \textbf{24}(3),  544--552
  (2005)

\bibitem{giraudot2013noise}
Giraudot, S., Cohen-Steiner, D., Alliez, P.: Noise-adaptive shape
  reconstruction from raw point sets. Computer Graphics Forum  \textbf{32}(5),
  229--238 (2013)

\bibitem{guennebaud2007algebraic}
Guennebaud, G., Gross, M.: Algebraic point set surfaces. ACM Transactions On
  Graphics.  \textbf{26},  23--es (2007)

\bibitem{guerrero2018pcpnet}
Guerrero, P., Kleiman, Y., Ovsjanikov, M., Mitra, N.J.: Pcpnet learning local
  shape properties from raw point clouds. Computer Graphics Forum.
  \textbf{37}(2),  75--85 (2018)

\bibitem{hashimoto2019normal}
Hashimoto, T., Saito, M.: Normal estimation for accurate {3D} mesh
  reconstruction with point cloud model incorporating spatial structure. In:
  Proceedings of the IEEE/CVF Conference on Computer Vision and Pattern
  Recognition Workshops. pp. 54--63 (2019)

\bibitem{hermosilla2019total}
Hermosilla, P., Ritschel, T., Ropinski, T.: Total denoising: Unsupervised
  learning of {3D} point cloud cleaning. In: Proceedings of the IEEE/CVF
  International Conference on Computer Vision. pp. 52--60 (2019)

\bibitem{hoppe1992surface}
Hoppe, H., DeRose, T., Duchamp, T., McDonald, J., Stuetzle, W.: Surface
  reconstruction from unorganized points. In: Proceedings of the 19th Annual
  Conference on Computer Graphics and Interactive Techniques. pp. 71--78 (1992)

\bibitem{hu2018squeeze}
Hu, J., Shen, L., Sun, G.: {Squeeze-and-Excitation} networks. In: Proceedings
  of the IEEE/CVF Conference on Computer Vision and Pattern Recognition. pp.
  7132--7141 (2018)

\bibitem{hua-pointwise-cvpr18}
Hua, B.S., Tran, M.K., Yeung, S.K.: Pointwise convolutional neural networks.
  In: Proceedings of the IEEE/CVF conference on Computer Vision and Pattern
  Recognition. pp. 984--993 (2018)

\bibitem{kazhdan2006poisson}
Kazhdan, M., Bolitho, M., Hoppe, H.: Poisson surface reconstruction. In:
  Proceedings of the 4th Eurographics Symposium on Geometry Processing. vol.~7
  (2006)

\bibitem{khaloo2017robust}
Khaloo, A., Lattanzi, D.: Robust normal estimation and region growing
  segmentation of infrastructure {3D} point cloud models. Advanced Engineering
  Informatics  \textbf{34},  1--16 (2017)

\bibitem{kingma2014adam}
Kingma, D.P., Ba, J.: Adam: A method for stochastic optimization. In:
  Proceedings of the International Conference on Learning Representations.
  (2015)

\bibitem{lenssen2020deep}
Lenssen, J.E., Osendorfer, C., Masci, J.: Deep iterative surface normal
  estimation. In: Proceedings of the IEEE/CVF Conference on Computer Vision and
  Pattern Recognition. pp. 11247--11256 (2020)

\bibitem{levin1998approximation}
Levin, D.: The approximation power of moving least-squares. Mathematics of
  Computation.  \textbf{67}(224),  1517--1531 (1998)

\bibitem{lu2020deep}
Lu, D., Lu, X., Sun, Y., Wang, J.: Deep feature-preserving normal estimation
  for point cloud filtering. Computer-Aided Design.  \textbf{125},  102860
  (2020)

\bibitem{lu2020low}
Lu, X., Schaefer, S., Luo, J., Ma, L., He, Y.: Low rank matrix approximation
  for {3D} geometry filtering. IEEE Transactions on Visualization and Computer
  Graphics.  \textbf{28}(04),  1835--1847 (2022)

\bibitem{merigot2010voronoi}
M{\'e}rigot, Q., Ovsjanikov, M., Guibas, L.J.: Voronoi-based curvature and
  feature estimation from point clouds. IEEE Transactions on Visualization and
  Computer Graphics.  \textbf{17}(6),  743--756 (2010)

\bibitem{mitra2003estimating}
Mitra, N.J., Nguyen, A.: Estimating surface normals in noisy point cloud data.
  In: Proceedings of the 19th Annual Symposium on Computational Geometry. pp.
  322--328 (2003)

\bibitem{nurunnabi2014robust}
Nurunnabi, A., Belton, D., West, G.: Robust statistical approaches for local
  planar surface fitting in {3D} laser scanning data. ISPRS Journal of
  Photogrammetry and Remote Sensing  \textbf{96},  106--122 (2014)

\bibitem{nurunnabi2015outlier}
Nurunnabi, A., West, G., Belton, D.: Outlier detection and robust
  normal-curvature estimation in mobile laser scanning {3D} point cloud data.
  Pattern Recognition  \textbf{48}(4),  1404--1419 (2015)

\bibitem{pistilli2020learning}
Pistilli, F., Fracastoro, G., Valsesia, D., Magli, E.: Learning
  graph-convolutional representations for point cloud denoising. In:
  Proceedings of the European Conference on Computer Vision. pp. 103--118
  (2020)

\bibitem{qi2017pointnet}
Qi, C.R., Su, H., Mo, K., Guibas, L.J.: {PointNet}: Deep learning on point sets
  for {3D} classification and segmentation. In: Proceedings of the IEEE/CVF
  Conference on Computer Vision and Pattern Recognition. pp. 652--660 (2017)

\bibitem{qi2017pointnet++}
Qi, C.R., Yi, L., Su, H., Guibas, L.J.: {PointNet++}: Deep hierarchical feature
  learning on point sets in a metric space. pp. 5100--5109 (2017)

\bibitem{rakotosaona2020pointcleannet}
Rakotosaona, M.J., La~Barbera, V., Guerrero, P., Mitra, N.J., Ovsjanikov, M.:
  {PointCleanNet}: Learning to denoise and remove outliers from dense point
  clouds. Computer Graphics Forum.  \textbf{39}(1),  185--203 (2020)

\bibitem{silberman2012indoor}
Silberman, N., Hoiem, D., Kohli, P., Fergus, R.: Indoor segmentation and
  support inference from rgbd images. In: Proceedings of the European
  Conference on Computer Vision. pp. 746--760 (2012)

\bibitem{wang2019dynamic}
Wang, Y., Sun, Y., Liu, Z., Sarma, S.E., Bronstein, M.M., Solomon, J.M.:
  Dynamic graph {CNN} for learning on point clouds. ACM Transactions On
  Graphics.  \textbf{38}(5),  1--12 (2019)

\bibitem{wang2020neighbourhood}
Wang, Z., Prisacariu, V.A.: Neighbourhood-insensitive point cloud normal
  estimation network (2020)

\bibitem{yu2018ec}
Yu, L., Li, X., Fu, C.W., Cohen-Or, D., Heng, P.A.: {EC-Net}: an edge-aware
  point set consolidation network. In: Proceedings of the European Conference
  on Computer Vision. pp. 386--402 (2018)

\bibitem{zhang2020pointfilter}
Zhang, D., Lu, X., Qin, H., He, Y.: {PointFilter}: Point cloud filtering via
  encoder-decoder modeling. IEEE Transactions on Visualization and Computer
  Graphics.  \textbf{27}(3),  2015--2027 (2020)

\bibitem{zhang2022geometry}
Zhang, J., Cao, J.J., Zhu, H.R., Yan, D.M., Liu, X.P.: Geometry guided deep
  surface normal estimation. Computer-Aided Design.  \textbf{142},  103119
  (2022)

\bibitem{zhou2020geometry}
Zhou, H., Chen, H., Feng, Y., Wang, Q., Qin, J., Xie, H., Wang, F.L., Wei, M.,
  Wang, J.: Geometry and learning co-supported normal estimation for
  unstructured point cloud. In: Proceedings of the IEEE/CVF Conference on
  Computer Vision and Pattern Recognition. pp. 13238--13247 (2020)

\bibitem{zhu2021adafit}
Zhu, R., Liu, Y., Dong, Z., Wang, Y., Jiang, T., Wang, W., Yang, B.: {AdaFit}:
  Rethinking learning-based normal estimation on point clouds. In: Proceedings
  of the IEEE/CVF International Conference on Computer Vision. pp. 6118--6127
  (2021)

\end{thebibliography}

\newpage
\appendix
\section{Appendix}

In this document, we provide \emph{additional details}, \emph{experiments} and \emph{applications} to support the original paper. Below is a summary of the contents:
\begin{itemize}
    \item A detailed description of the adopted datasets and training settings, as well as the architecture of the proposed network are provided.
    \item We demonstrate the practical applications of our method for \emph{point cloud denoising} and \emph{3D surface reconstruction} based on normal estimation.
    \item More qualitative results, including angle RMSE and PGP5/10, are reported.
\end{itemize}

\subsection{Implementation Details}
\label{sec:implement}
We use the benchmark dataset PCPNet \cite{guerrero2018pcpnet} to train the proposed network and test it. The training set contains eight shapes: four CAD objects (boxunion, cup, fandisk and flower) and four high-quality scans of figurines (armadillo, bunny, dragon and turtle). All shapes are modeled as triangular meshes and densely sampled with 100 k points. To augment the training set, we add Gaussian noise with zero mean and varying standard deviation $\sigma\in \{0.012, 0.006, 0.00125\}$, with respect to the diagonal length of the bounding box for each model. Then we attain 32 point clouds for training. The test set has 19 shapes, including CAD objects, figurines, and analytic shapes.

\begin{figure}[]
  \centering
 \includegraphics[scale=0.45]{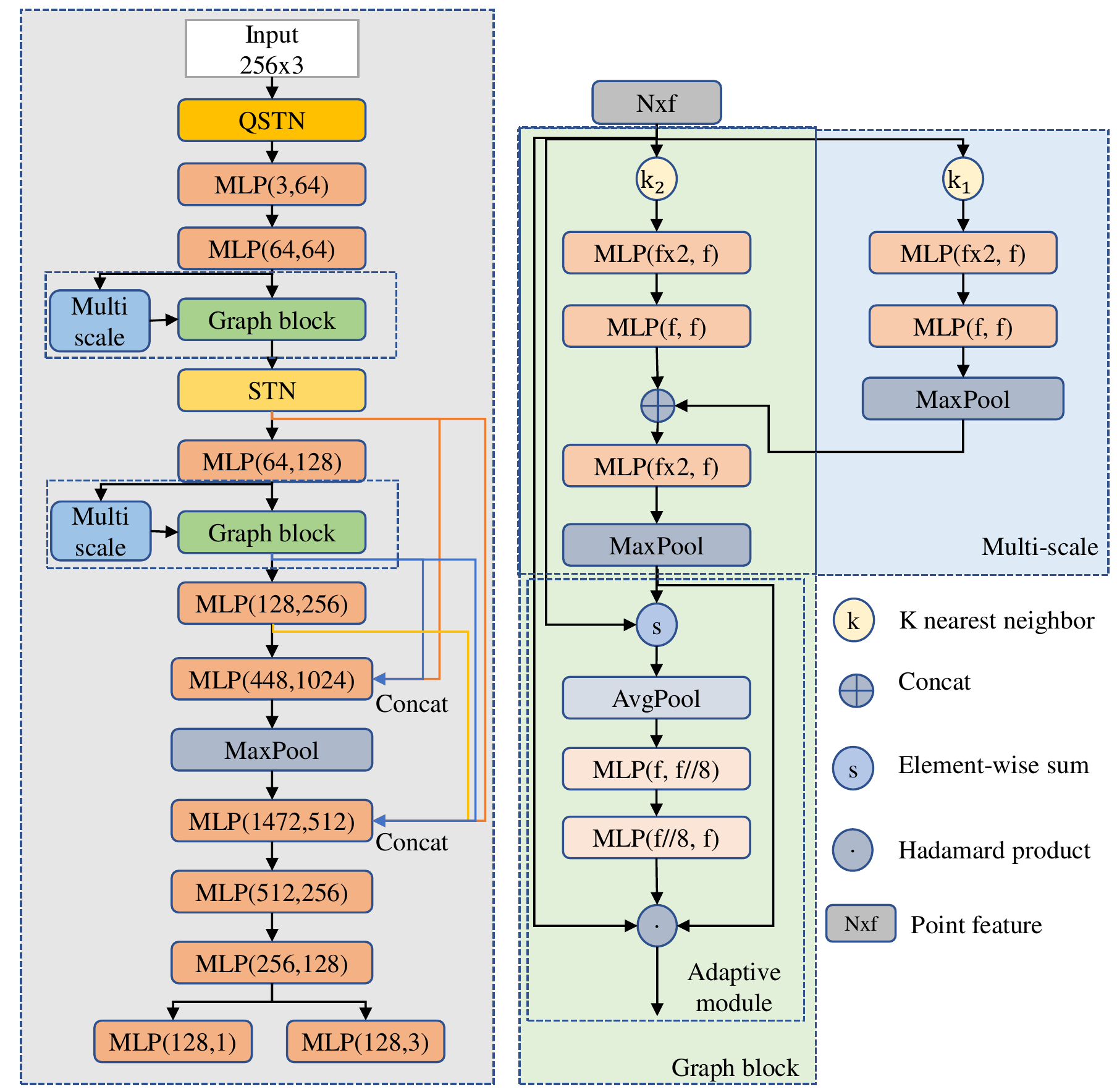}
  \caption{Architecture details of our proposed network GraphFit.}
  \label{fig:GraphFit}
\end{figure}
The proposed network is trained and implemented by the Pytorch framework on a Nvidia Tesla v100 GPU, using the Adam optimizer\cite{kingma2014adam}. The batch size and initial learning rate are equal to 256 and $1e\mbox{-}3$, respectively. We train the network 600 epochs in total, and the learning rate decays to $10\%$ of the initial value at epoch 200 and 500. The overall training loss is defined as
\\
\begin{equation}
\mathcal{L}_{\text{tol}}=\left|\mathbf{n}_{gt} \times  \hat{\mathbf{n}}\right|+\mathcal{L}_{\text {con }}+\lambda_{3} \mathcal{L}_{\text {reg1}}+\lambda_{4} \mathcal{L}_{\text {reg2}},
\end{equation}
\begin{equation}
\mathcal{L}_{\text {con }}=\frac{1}{N_{p_{i}}}\left[-\lambda_{1} \sum_{j=1}^{N_{p_{i}}} \log \left(w_{j}\right)+\lambda_{2} \sum_{j=1}^{N_{p_{i}}} w_{j}\left|\mathbf{n}_{gt, j} \times \hat{\mathbf{n}}_{j}\right|\right],
\end{equation}

\begin{equation}
\mathcal{L}_{\text{reg1}}=\left|I-A_{1} A_{1}^{T}\right|,
\end{equation}
\begin{equation}
\mathcal{L}_{\text{reg2}}=\left|I-A_{2} A_{2}^{T}\right|,
\end{equation}
where we use the default settings $\lambda_{1} = 0.05$, $\lambda_{2} = 0.25$, $\lambda_{3}=0.1$, and $\lambda_{4} = 0.01$.
\subsection{Details of Our Proposed Network Architecture}
\cref{fig:GraphFit} presents the details of our proposed network architecture. An MLP unit consists of a $1\times1 $ Conv, a BatchNorm layer and a ReLU in the left of \cref{fig:GraphFit}. For MLP in the Graph block and the multi-scale layer, we use the Leaky ReLU (LReLU) as its activation function.
\subsection{Results on the SceneNN Dataset}
\begin{figure}[h]
  \centering
     \includegraphics[scale=0.36]{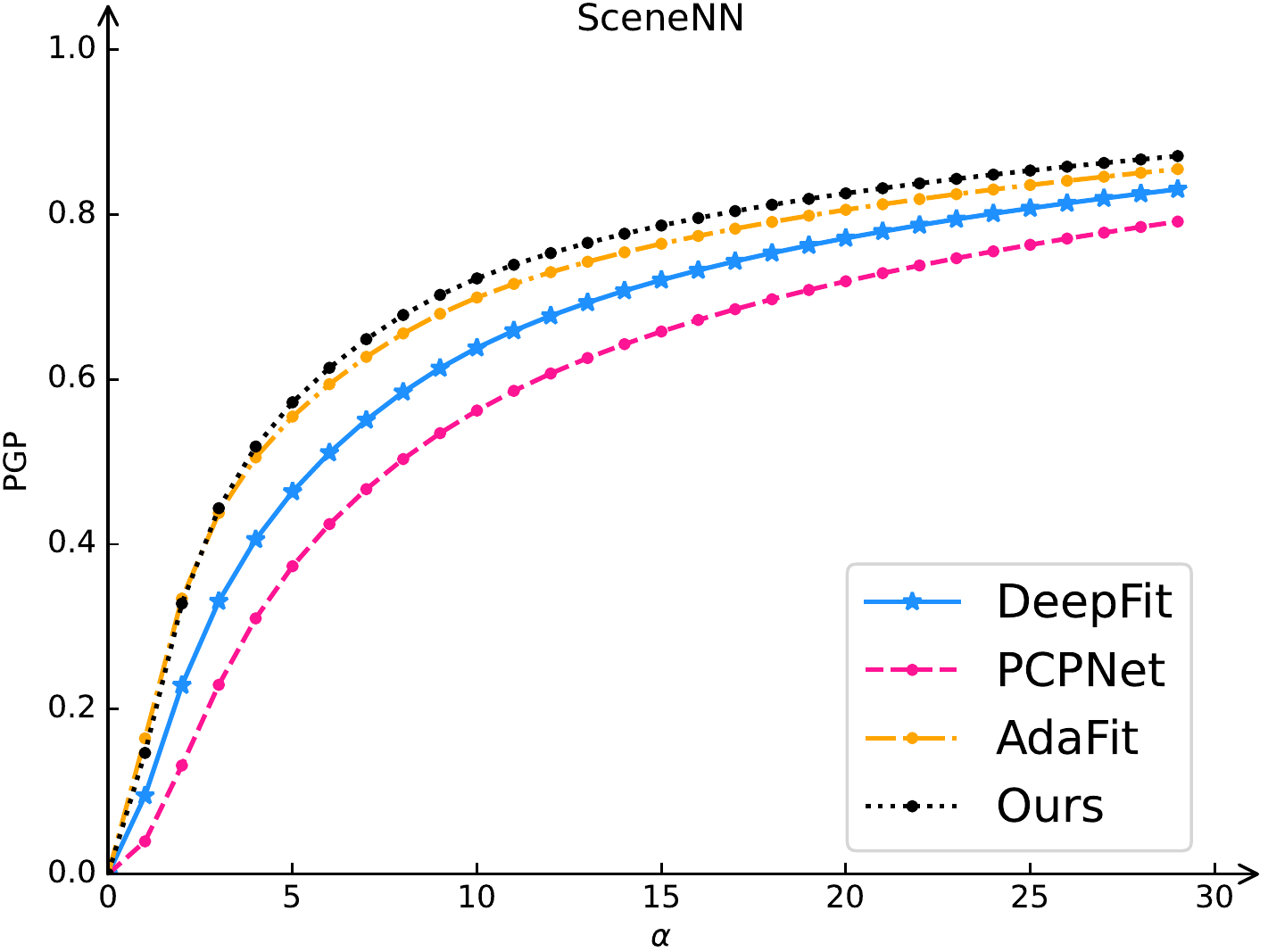}
   \caption{Comparison of $PGP \alpha $ for unoriented normal estimation on the SceneNN dataset~\cite{hua-pointwise-cvpr18}.
   }
   \label{fig:scenenn}
\end{figure}

 The $PGP \alpha$ results are shown in \cref{fig:scenenn},  our method significantly outperforms all competitors and achieves the state-of-the-art performance.
\subsection{Ablation Study of Different Graph Blocks}
To demonstrate the effectiveness of different graph blocks in the proposed network, we further implement ablation study and report the average RMSE in \cref{tab:graph}, in which the neighborhood size and the jet order are equal to 256 and 3, respectively. As can be seen, the network achieves higher accuracy for both noise and varying point density by adding graph blocks, where the average RMSE decreases from 10.76 to 10.52. Nevertheless, with the number of graph blocks increasing, such as from two to three, the angle RMSE of unoriented normal vectors does not reduce significantly, thereby we adopt one graph block for use.
\begin{table}[t]
\caption{Investigation on the influence of different number of graph blocks on normal estimation.}
\begin{center}
\setlength{\tabcolsep}{1mm}
\begin{tabular}{@{}cccc@{}}
\toprule
Aug.               & 1 Graph-block & 2 Graph-blocks & 3 Graph-blocks \\ \midrule
No Noise           & 4.90          & 4.56          & 4.43          \\
$\sigma$ = 0.125\% & 9.06          & 8.87          & 8.88          \\
$\sigma$ = 0.6\%   & 16.58         & 16.57         & 16.53         \\
$\sigma$ = 1.2\%   & 22.87         & 22.77         & 22.74         \\
Gradient           & 5.52          & 5.33          & 5.22          \\
Strip              & 5.61          & 5.38          & 5.31          \\
Average            & 10.76         & 10.58         & 10.52         \\ \bottomrule
\end{tabular}
\end{center}
\label{tab:graph}
\end{table}
\begin{figure}
	\begin{center}
	\begin{subfigure}{0.15\linewidth}
		\includegraphics[scale=0.1]{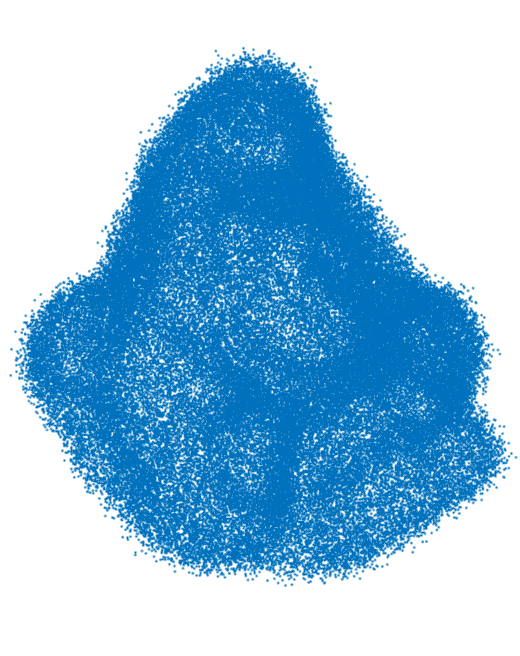}
		\caption{Input}
	\end{subfigure}
	\begin{subfigure}{0.15\linewidth }
		\includegraphics[scale=0.1]{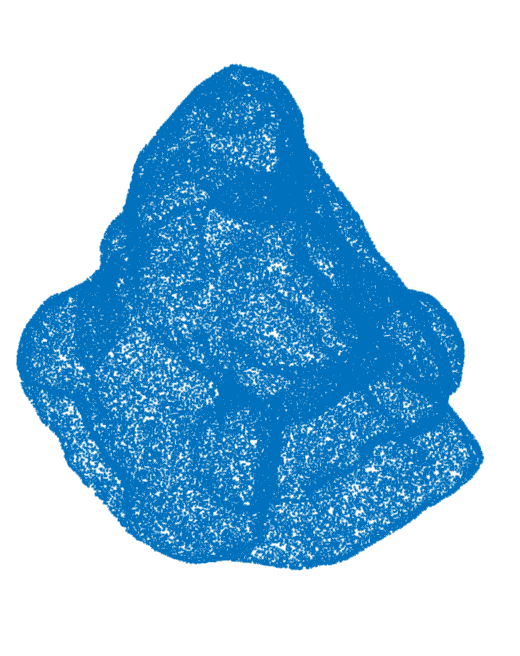}
		\caption{Ours}
	\end{subfigure}
	\begin{subfigure}{0.15\linewidth}
		\includegraphics[scale=0.1]{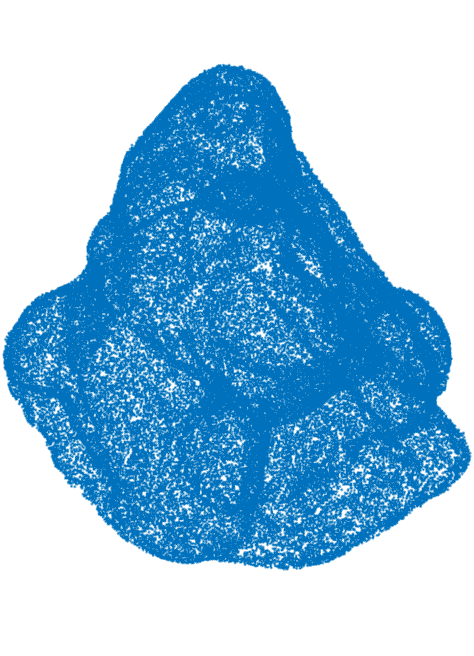}
		\caption{GT}
	\end{subfigure}
	\begin{subfigure}{0.15\linewidth}
		\includegraphics[scale=0.1]{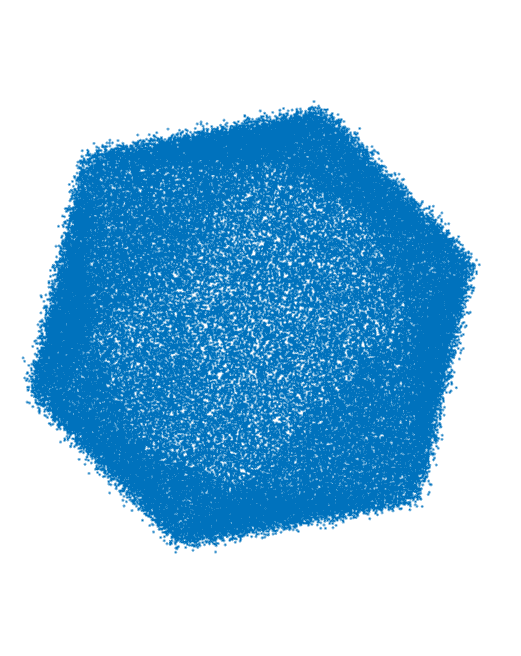}
		\caption{Input}
	\end{subfigure}
	\begin{subfigure}{0.15\linewidth}
		\includegraphics[scale=0.1]{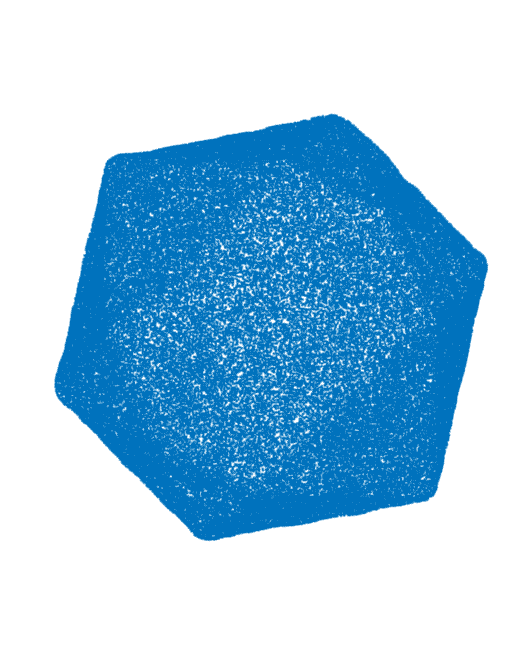}
		\caption{Ours}
	\end{subfigure}
	\begin{subfigure}{0.15   \linewidth}
		\includegraphics[scale=0.1]{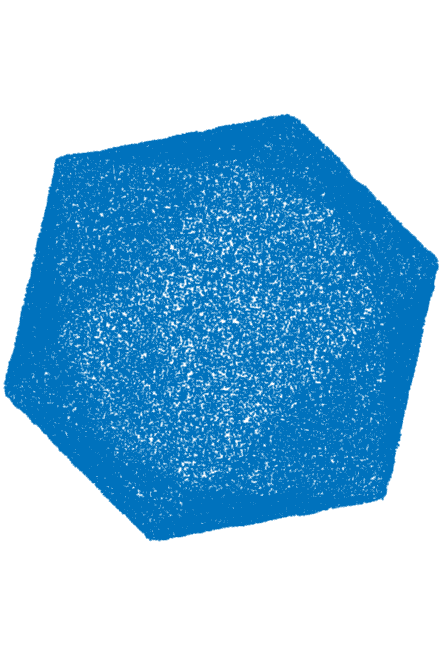}
		\caption{GT}
	\end{subfigure}
	\caption{Point cloud denoising using our proposed normal estimator. 
	}
	\label{fig:denoisy}
	\end{center}
\end{figure}

\subsection{Practical Applications}
\begin{figure}
\centering
  \begin{subfigure}{0.3\linewidth}
  \centering
    \includegraphics[scale=0.1]{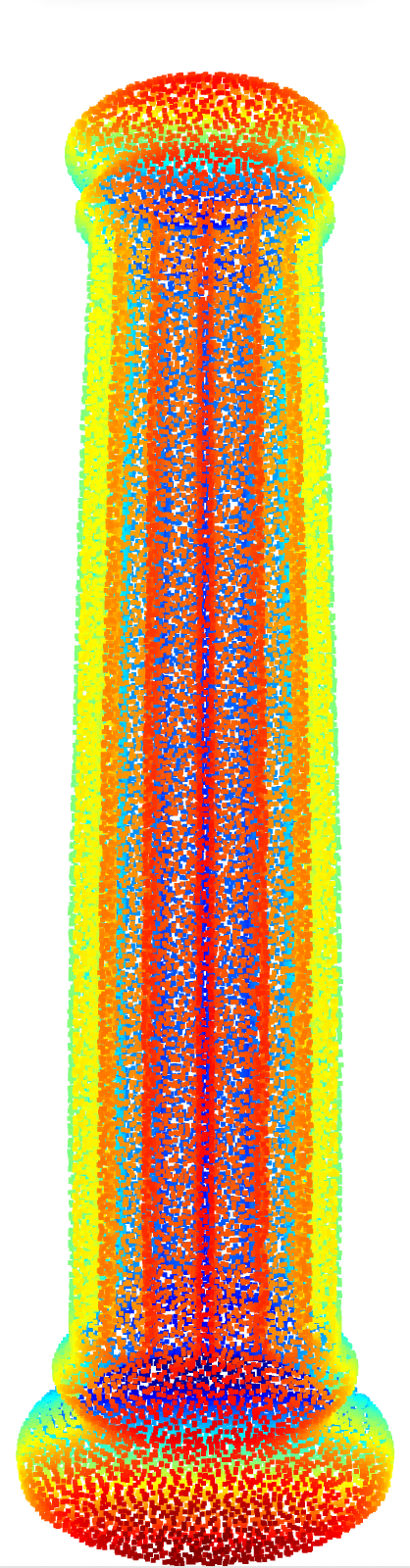}
  \end{subfigure}
  \begin{subfigure}{0.3\linewidth}
  \centering
    \includegraphics[scale=0.1]{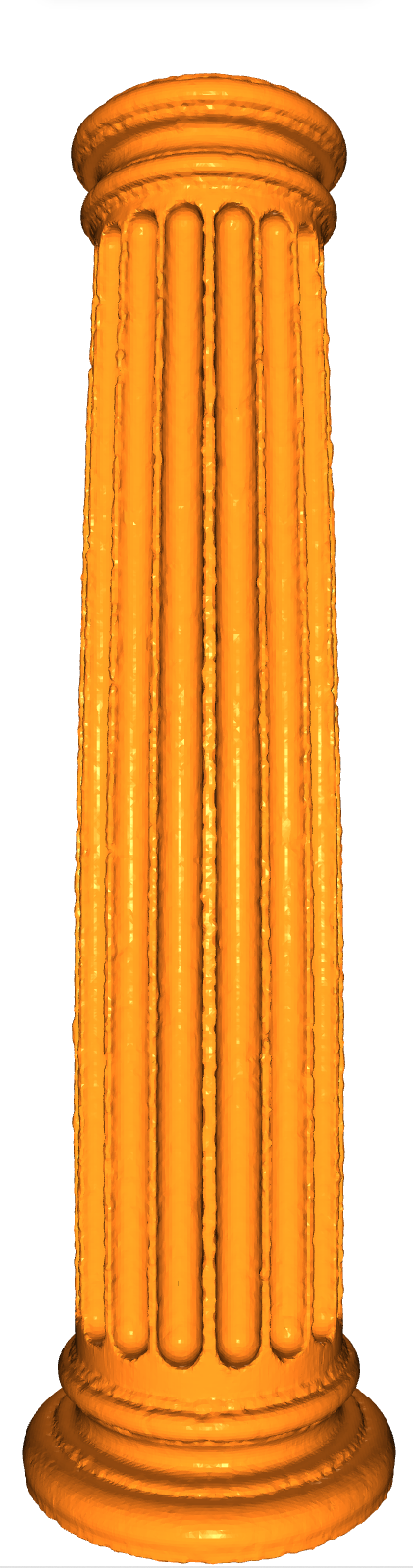}
  \end{subfigure}
  \begin{subfigure}{0.3\linewidth}
  \centering
    \includegraphics[scale=0.1]{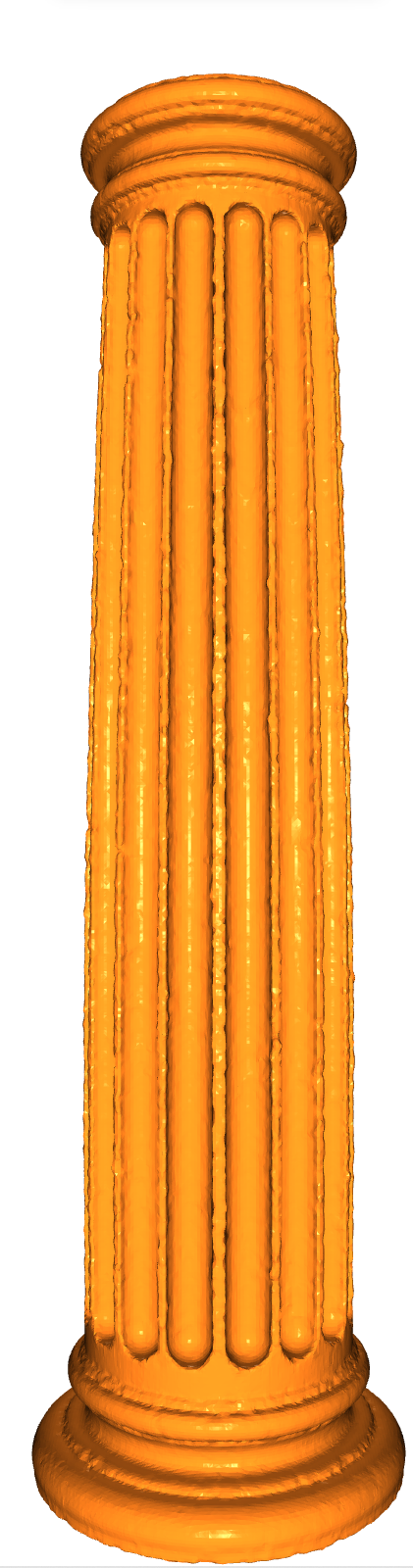}
  \end{subfigure}
  
    \begin{subfigure}{0.3\linewidth}
    \centering
    \includegraphics[scale=0.1]{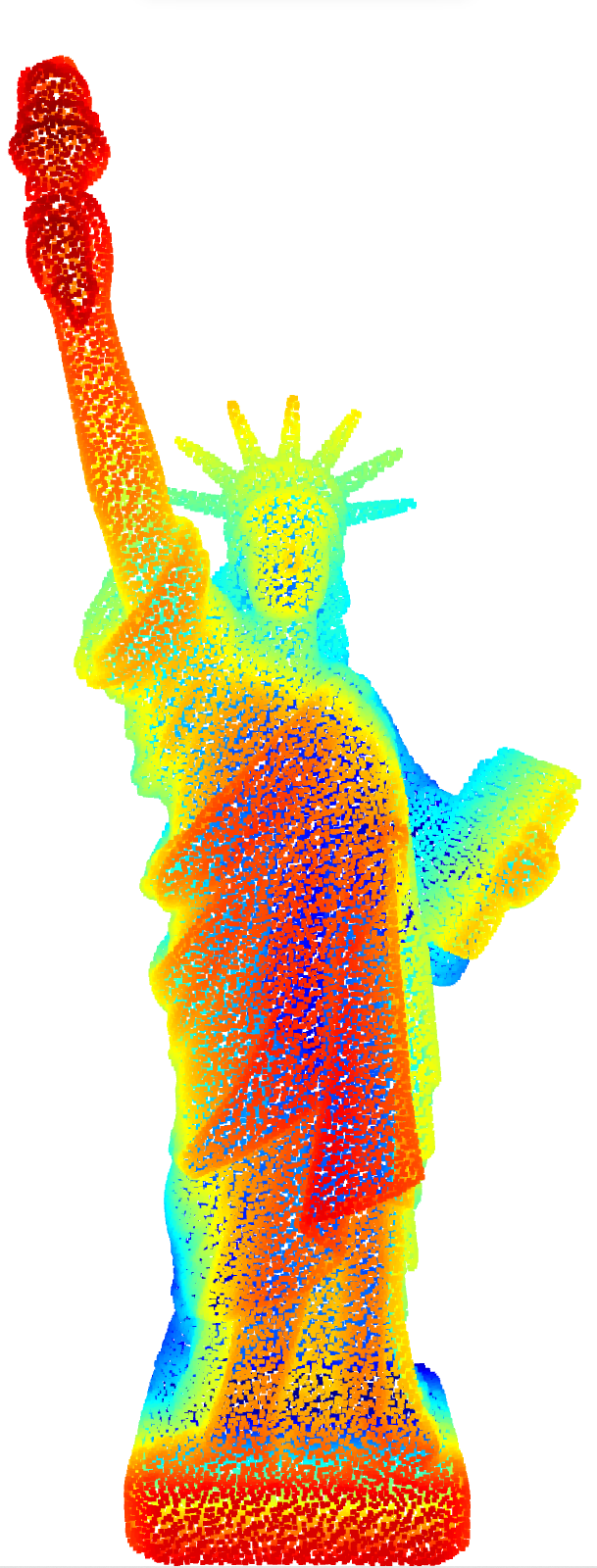}
  \end{subfigure}
  \begin{subfigure}{0.3\linewidth}
  \centering
    \includegraphics[scale=0.1]{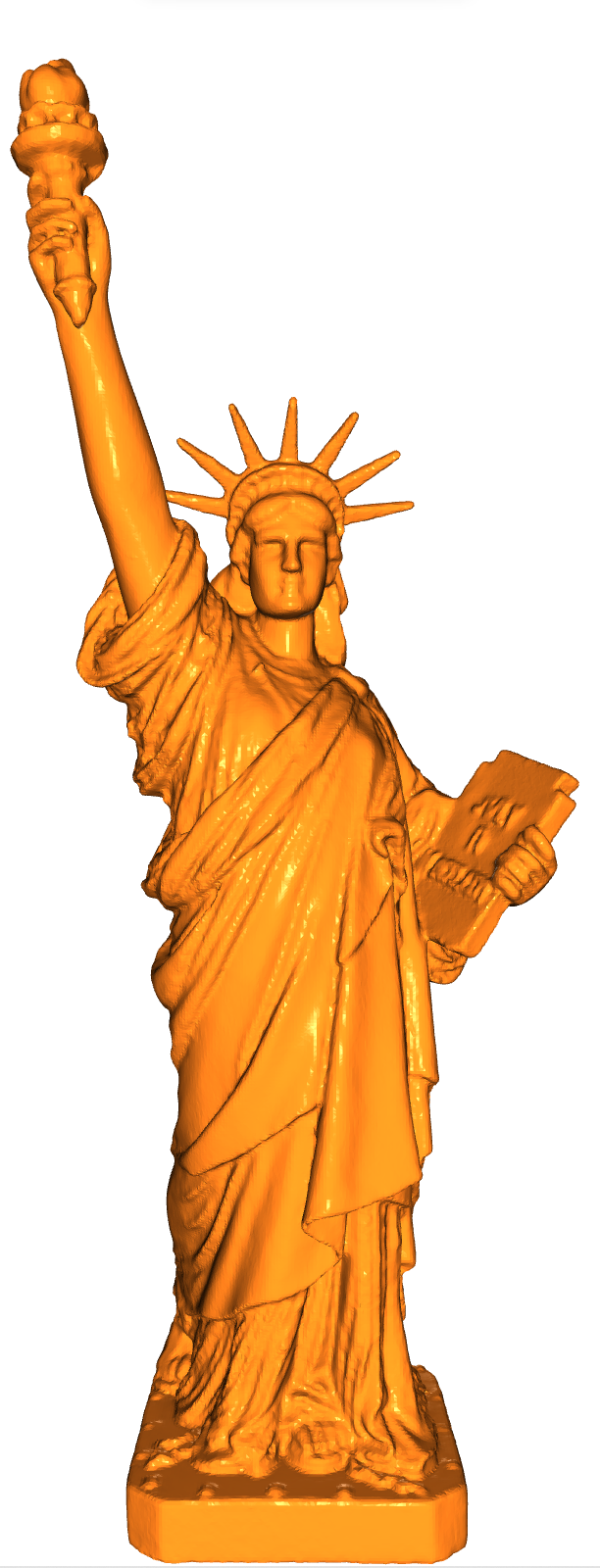}
  \end{subfigure}
  \begin{subfigure}{0.3\linewidth}
  \centering
    \includegraphics[scale=0.1]{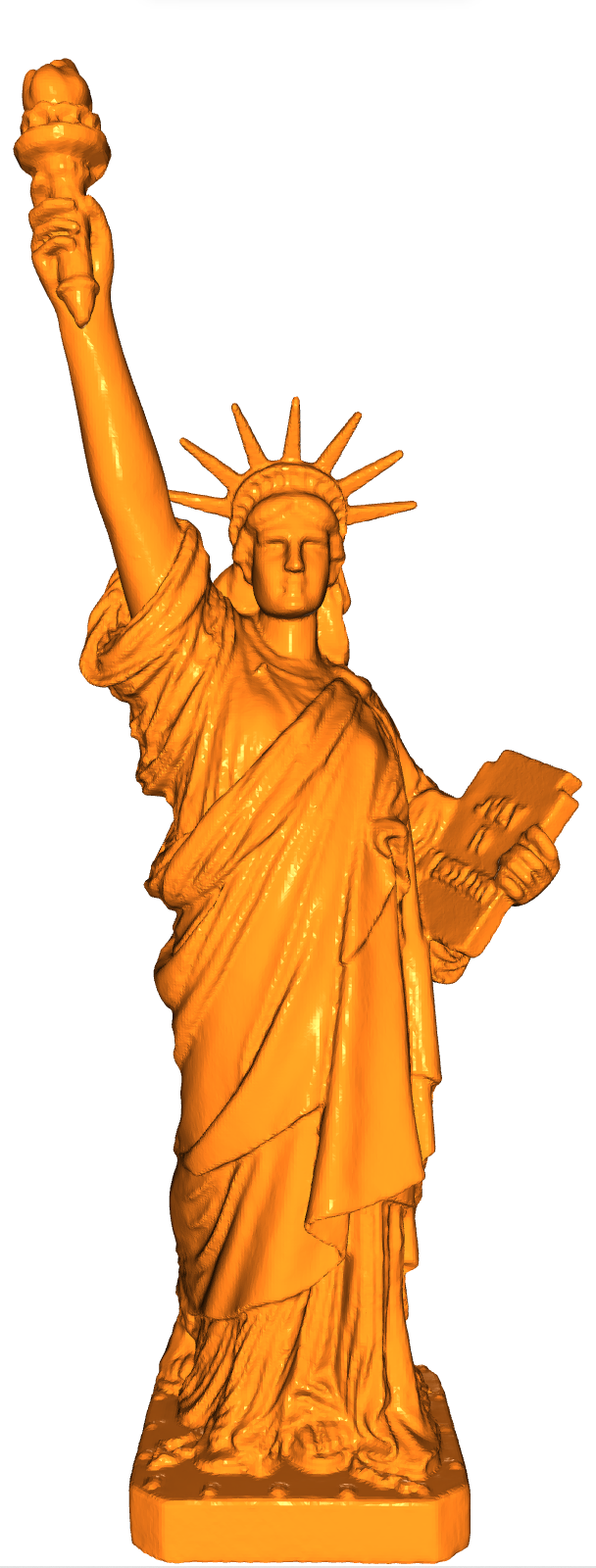}
  \end{subfigure}
      \begin{subfigure}{0.3\linewidth}
      \centering
    \includegraphics[scale=0.05]{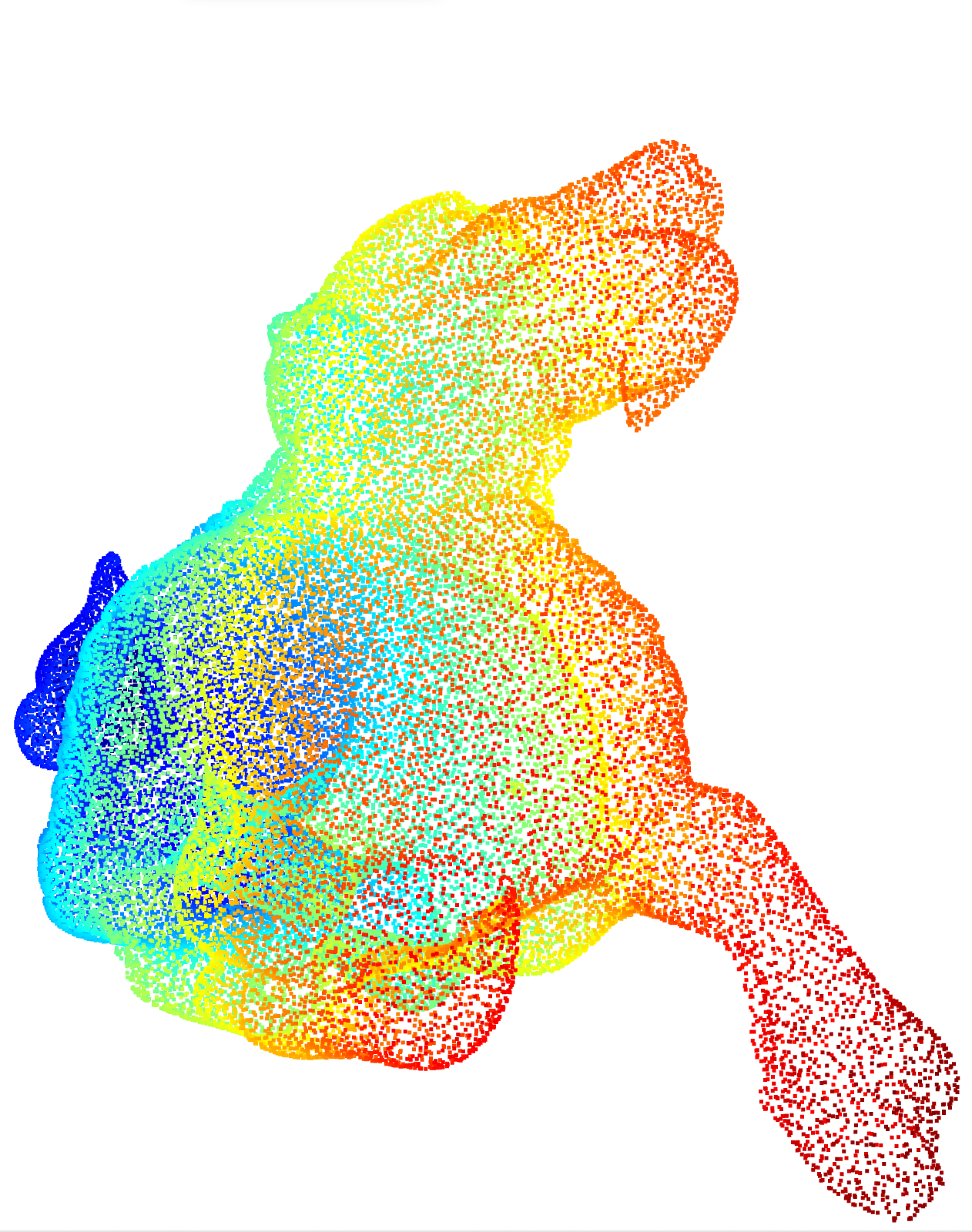}
    \caption{Input}
  \end{subfigure}
  \begin{subfigure}{0.3\linewidth}
  \centering
    \includegraphics[scale=0.05]{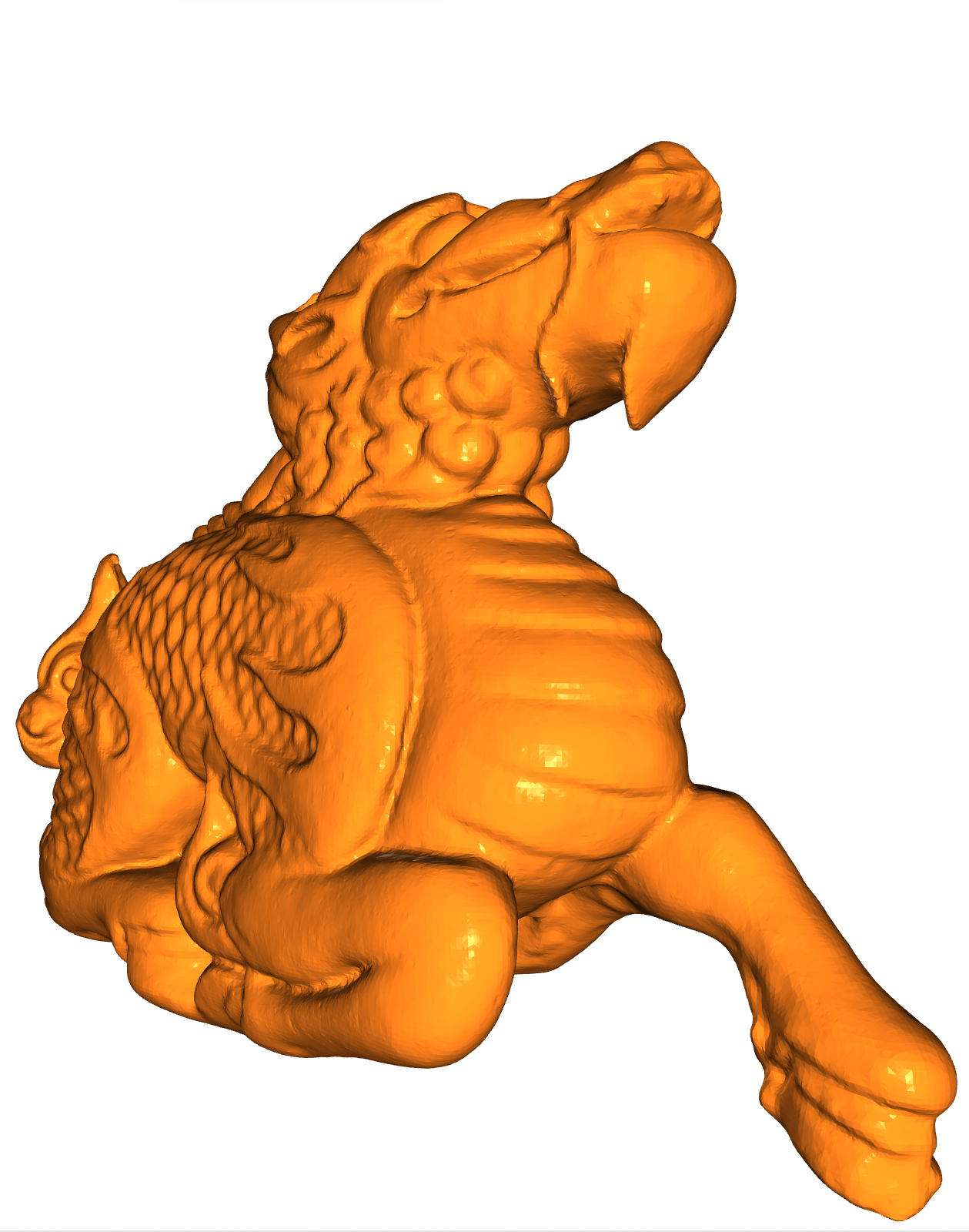}
    \caption{Ours}
  \end{subfigure}
  \begin{subfigure}{0.3\linewidth}
  \centering
    \includegraphics[scale=0.05]{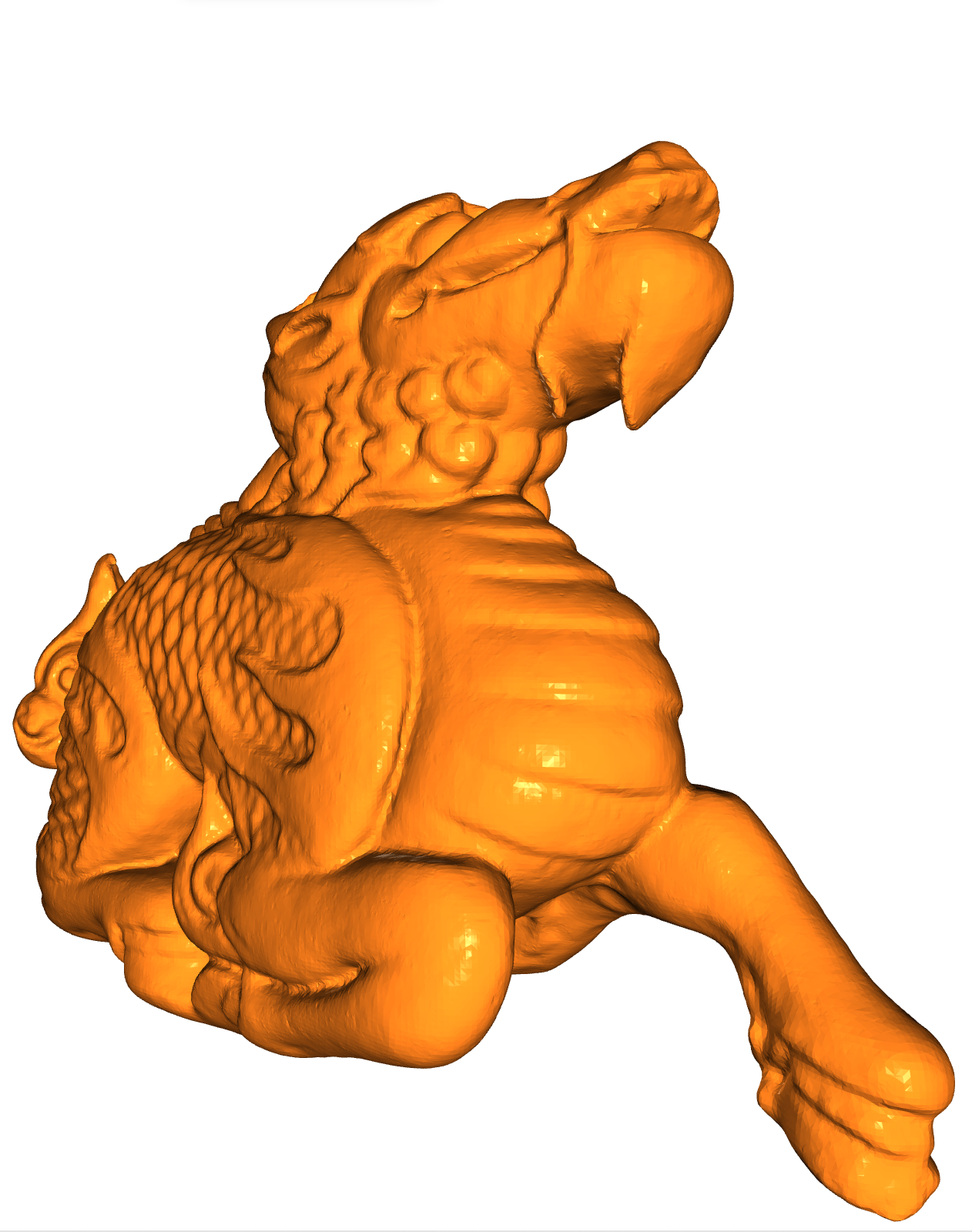}
    \caption{Groundtruth}
  \end{subfigure}
  
  \caption{Poisson surface reconstruction via the proposed normal estimator. 
}
  \label{fig:recon}
\end{figure}

We also deploy the proposed normal estimation method for point cloud denoising and 3D surface reconstruction. 
\\
\noindent\textbf{Point cloud denoising.} It is an important task in the 3D vision field to remove noise disturbance for point clouds. We combine the proposed normal estimation method with the \emph{modified edge recovery algorithm} in \cite{lu2020low} to update the point positions $\mathbf{p}_i$. The new position $\mathbf{p}_{i}^{\prime}$ is 
calculated by
\begin{equation}
\mathbf{p}_{i}^{\prime}=\mathbf{p}_{i}+\gamma_{i} \sum_{j \in \mathcal{N}_i}\left(\mathbf{p}_{j}-\mathbf{p}_{i}\right)\left(\mathbf{n}_{i}^{T} \mathbf{n}_{i} + \mathbf{n}_{j}^{T} \mathbf{n}_{j}\right)
\end{equation}
where $\mathcal{N}_i$ are the neighboring points of $\mathbf{p}_i$, $\mathbf{n}_i, \mathbf{n}_j$ are the estimated normals. \cref{fig:denoisy} shows sample denosing results. As observed, our method attains highly promising denosing performance compared with the ground truths, which are recovered via the ground truth normals.
\begin{table}[t]
	\centering
	\caption{Comparison of the 
		surface reconstruction precision (RMSE) of our proposed method and AdaFit, DeepFit, PCPNet.}
	\begin{tabular}{@{}ccccc@{}}
		\toprule
		& Ours             & AdaFit  & DeepFit & PCPNet  \\ \midrule
		Liberty & \textbf{0.00039} & 0.00064 & 0.00084 & 0.00165 \\
		Column  & \textbf{0.00592} & 0.00885 & 0.01077 & 0.01667 \\
		Netsuke & \textbf{0.00495} & 0.00501 & 0.00993 & 0.01573 \\
		Average & \textbf{0.00375} & 0.00483 & 0.00718 & 0.01135 \\ \bottomrule
	\end{tabular}
	\label{tab:recon}
\end{table}
\\
\noindent\textbf{3D surface reconstruction.} One common application of normal estimation for point clouds is to reconstruct the potential surface. We apply the proposed normal estimation method and adopt \emph{Poisson reconstruction} \cite{kazhdan2006poisson} for this purpose. Tab. \ref{tab:recon} reports the RMSE of all compared approaches in the sense of \emph{Hausdorff distance}. Results demonstrate that our method achieves high-quality reconstruction on all test cases. It is more accurate than baseline competitors. We present several reconstructed surface in \cref{fig:recon}.

\section{More Qualitative Results}
\begin{figure*}
	\centering
	\begin{subfigure}{0.15\linewidth}
		\centering
		\includegraphics[scale=0.1]{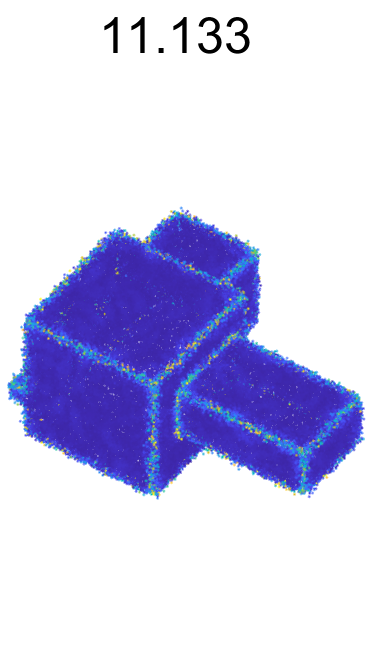}
	\end{subfigure}
	\hspace{0.1cm}
	\begin{subfigure}{0.15\linewidth}
		\centering
		\includegraphics[scale=0.1]{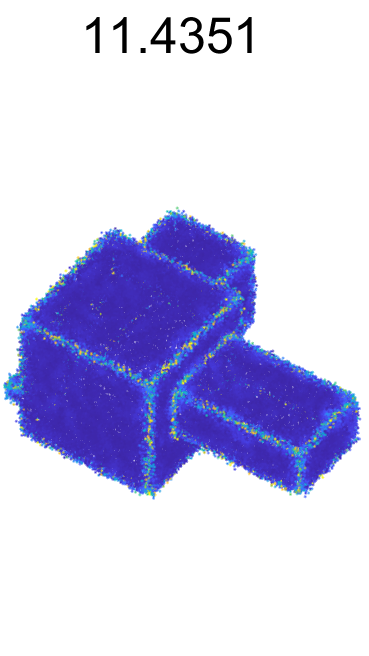}
	\end{subfigure}
	\hspace{0.1cm}
	\begin{subfigure}{0.17\linewidth}
		\centering
		\includegraphics[scale=0.1]{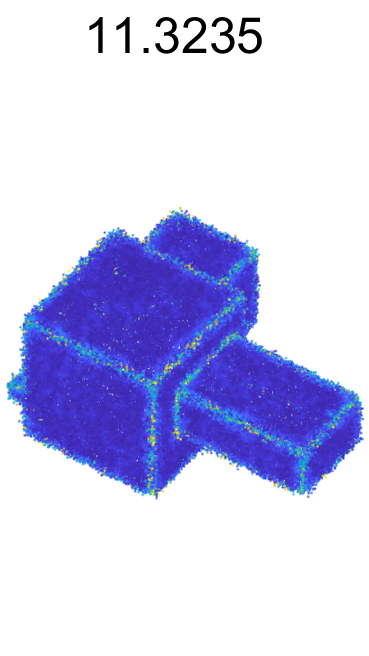}
	\end{subfigure}
	\hspace{0.1cm}
	\begin{subfigure}{0.17\linewidth}
		\centering
		\includegraphics[scale=0.1]{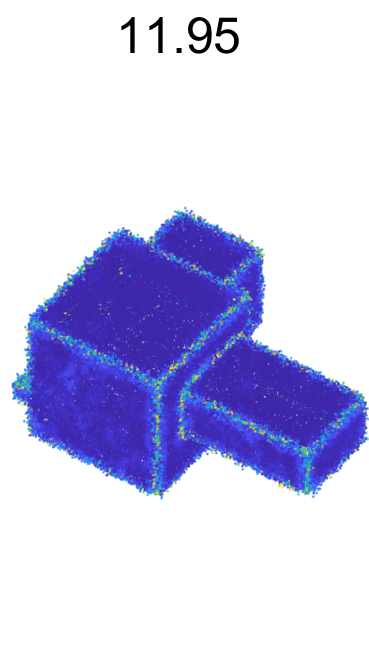}
	\end{subfigure}
	\hspace{0.1cm}
	\begin{subfigure}{0.17\linewidth}
		\centering
		\includegraphics[scale=0.1]{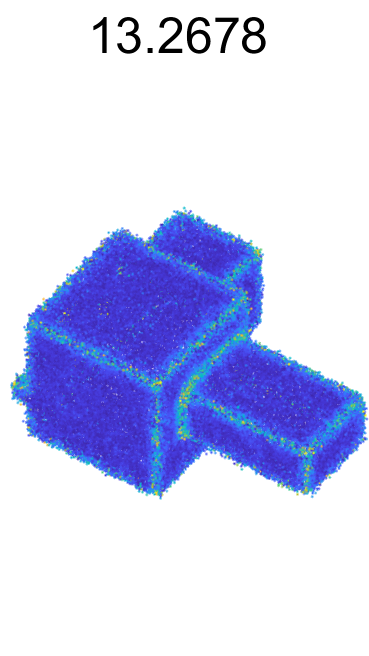}
	\end{subfigure}
	
	\begin{subfigure}{0.15\linewidth}
		\centering
		\includegraphics[scale=0.1]{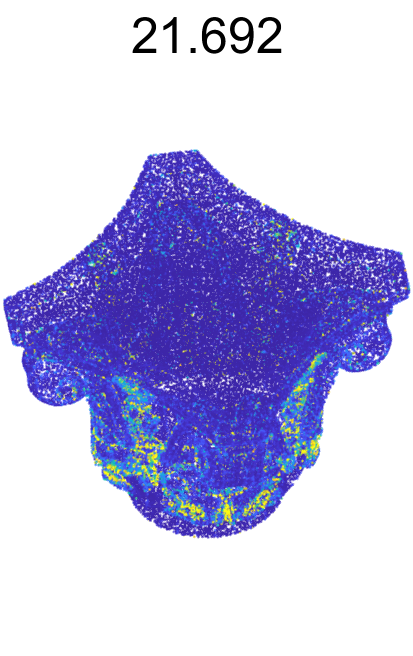}
	\end{subfigure}
	\hspace{0.1cm}
	\begin{subfigure}{0.15\linewidth}
		\centering
		\includegraphics[scale=0.1]{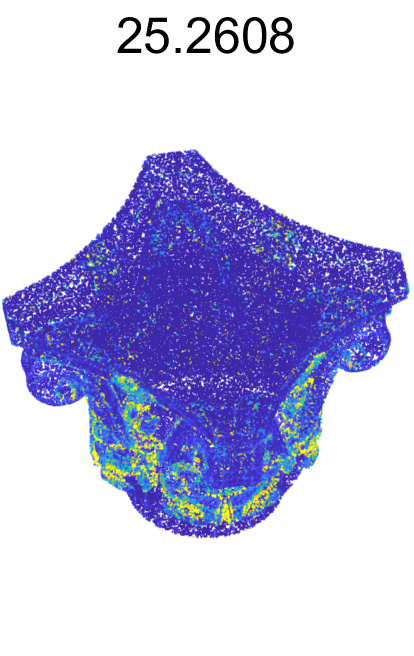}
	\end{subfigure}
	\hspace{0.1cm}
	\begin{subfigure}{0.17\linewidth}
		\centering
		\includegraphics[scale=0.1]{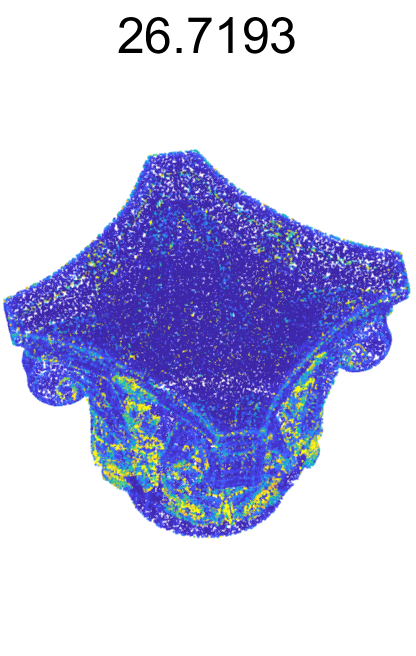}
	\end{subfigure}
	\hspace{0.1cm}
	\begin{subfigure}{0.17\linewidth}
		\centering
		\includegraphics[scale=0.1]{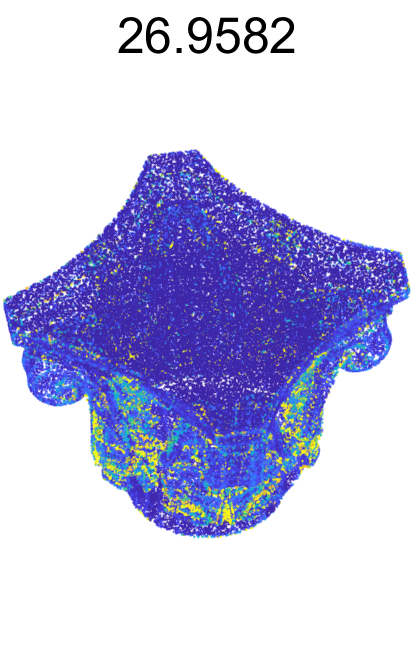}
	\end{subfigure}
	\hspace{0.1cm}
	\begin{subfigure}{0.17\linewidth}
		\centering
		\includegraphics[scale=0.1]{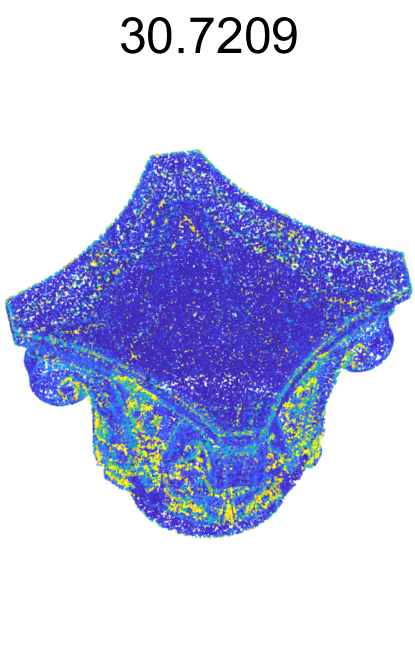}
	\end{subfigure}
	
	\begin{subfigure}{0.15\linewidth}
		\centering
		\includegraphics[scale=0.1]{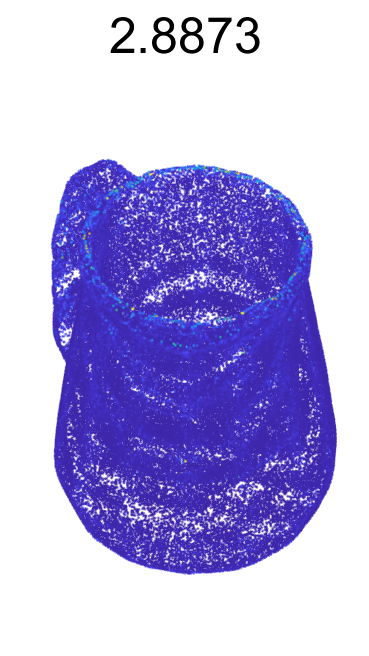}
	\end{subfigure}
	\hspace{0.1cm}
	\begin{subfigure}{0.15\linewidth}
		\centering
		\includegraphics[scale=0.1]{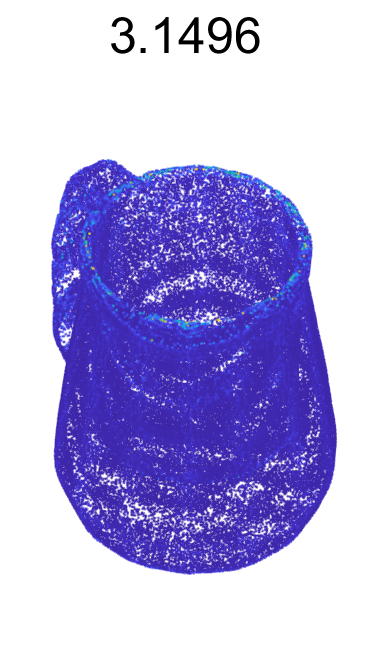}
	\end{subfigure}
	\hspace{0.1cm}
	\begin{subfigure}{0.17\linewidth}
		\centering
		\includegraphics[scale=0.1]{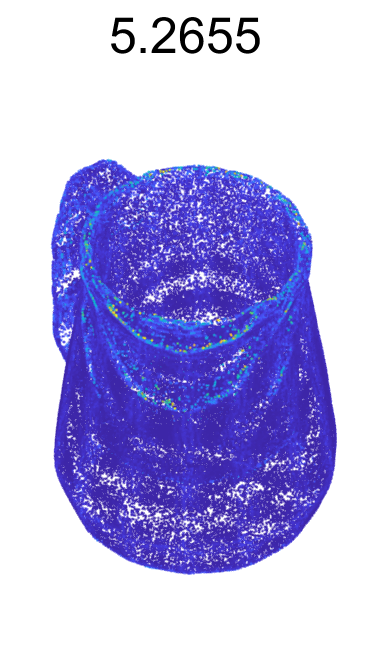}
	\end{subfigure}
	\hspace{0.1cm}
	\begin{subfigure}{0.17\linewidth}
		\centering
		\includegraphics[scale=0.1]{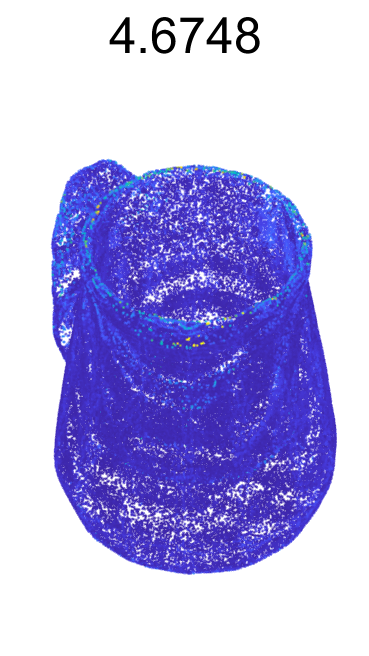}
	\end{subfigure}
	\hspace{0.1cm}
	\begin{subfigure}{0.17\linewidth}
		\centering
		\includegraphics[scale=0.1]{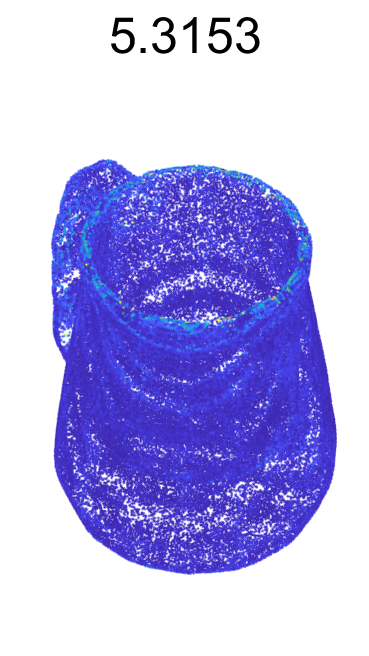}
	\end{subfigure}
	
	\begin{subfigure}{0.15\linewidth}
		\centering
		\includegraphics[scale=0.1]{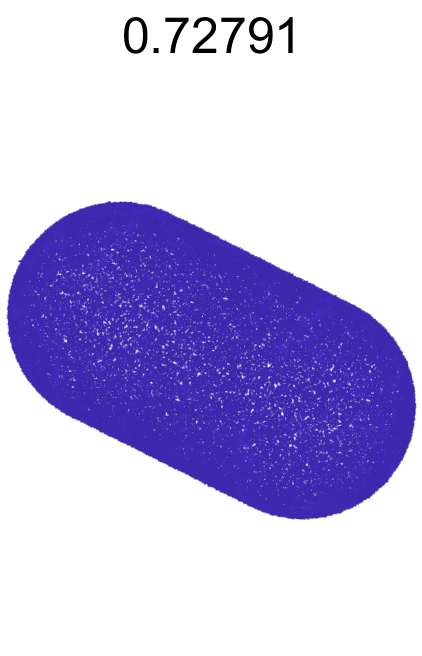}
	\end{subfigure}
	\hspace{0.1cm}
	\begin{subfigure}{0.15\linewidth}
		\centering
		\includegraphics[scale=0.1]{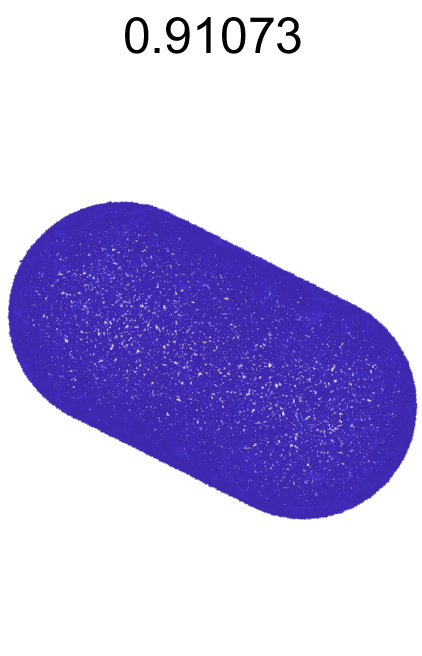}
	\end{subfigure}
	\hspace{0.1cm}
	\begin{subfigure}{0.17\linewidth}
		\centering
		\includegraphics[scale=0.1]{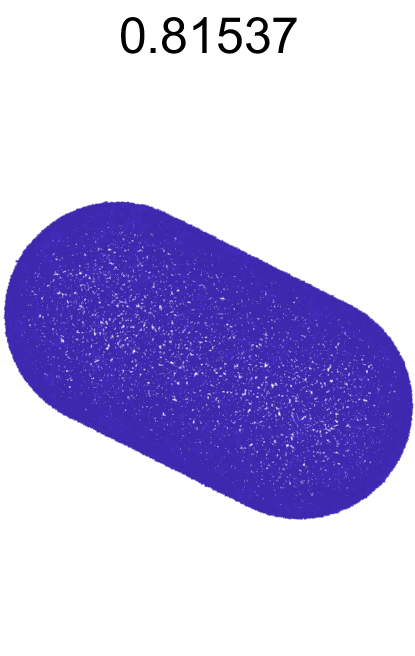}
	\end{subfigure}
	\hspace{0.1cm}
	\begin{subfigure}{0.17\linewidth}
		\centering
		\includegraphics[scale=0.1]{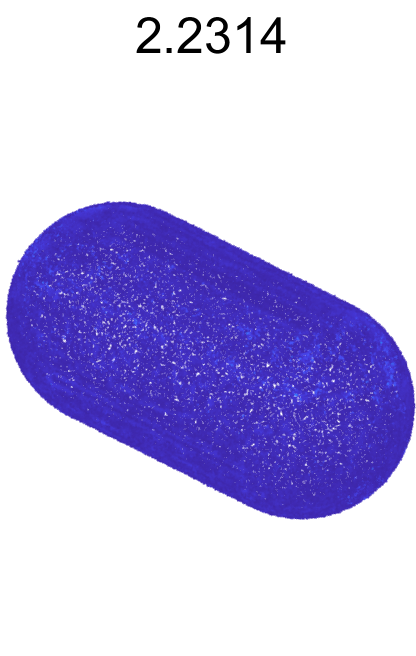}
	\end{subfigure}
	\hspace{0.1cm}
	\begin{subfigure}{0.17\linewidth}
		\centering
		\includegraphics[scale=0.1]{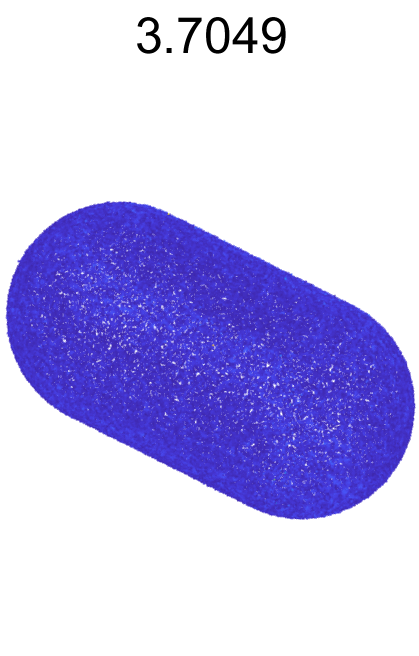}
	\end{subfigure}
	
	\begin{subfigure}{0.15\linewidth}
		\centering
		\includegraphics[scale=0.1]{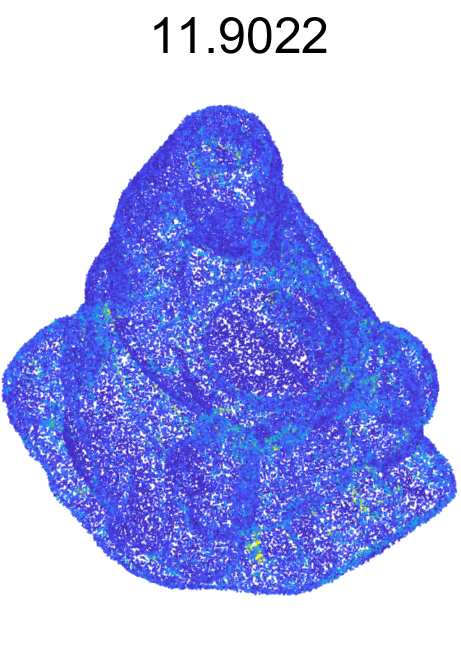}
	\end{subfigure}
	\hspace{0.1cm}
	\begin{subfigure}{0.15\linewidth}
		\centering
		\includegraphics[scale=0.1]{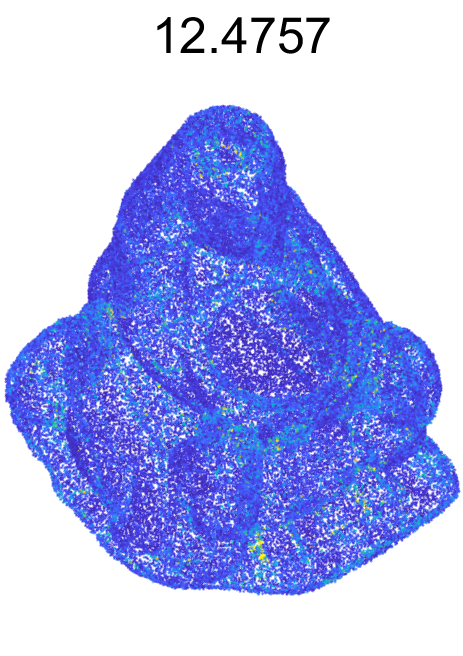}
	\end{subfigure}
	\hspace{0.1cm}
	\begin{subfigure}{0.17\linewidth}
		\centering
		\includegraphics[scale=0.1]{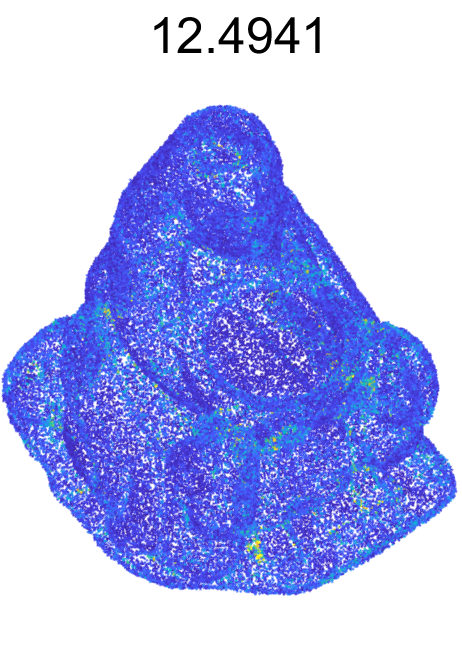}
	\end{subfigure}
	\hspace{0.1cm}
	\begin{subfigure}{0.17\linewidth}
		\centering
		\includegraphics[scale=0.1]{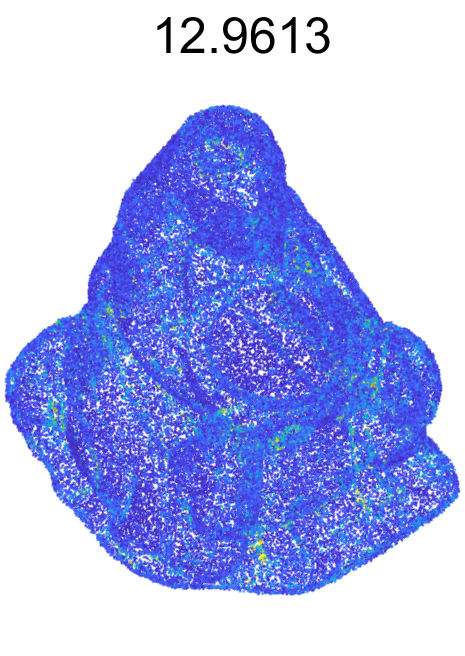}
	\end{subfigure}
	\hspace{0.1cm}
	\begin{subfigure}{0.17\linewidth}
		\centering
		\includegraphics[scale=0.1]{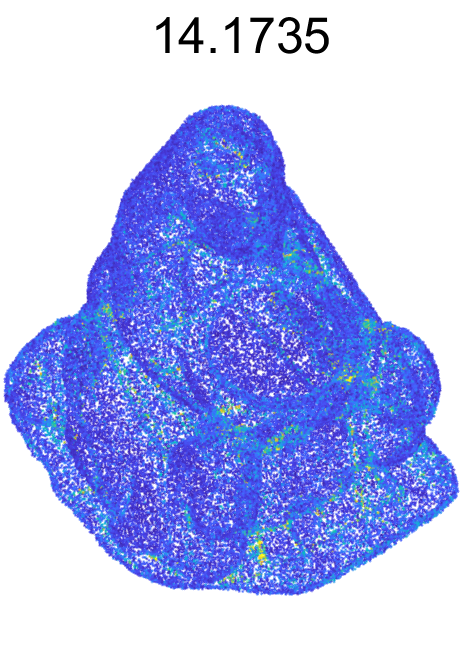}
	\end{subfigure}
	
	\begin{subfigure}{0.15\linewidth}
		\centering
		\includegraphics[scale=0.1]{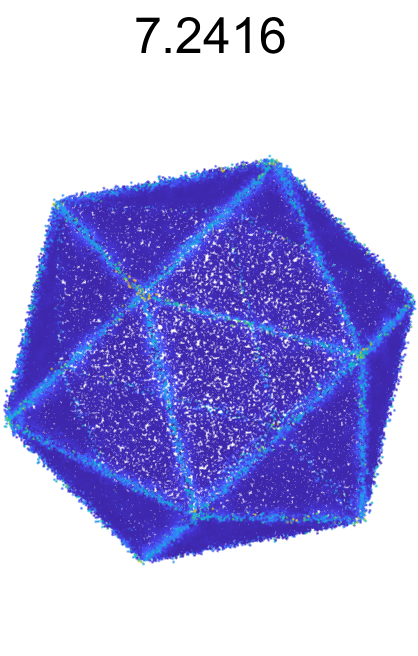}
	\end{subfigure}
	\hspace{0.1cm}
	\begin{subfigure}{0.15\linewidth}
		\centering
		\includegraphics[scale=0.1]{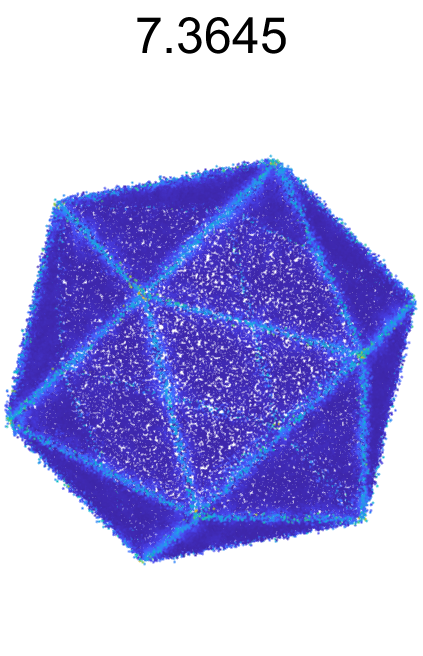}
	\end{subfigure}
	\hspace{0.1cm}
	\begin{subfigure}{0.17\linewidth}
		\centering
		\includegraphics[scale=0.1]{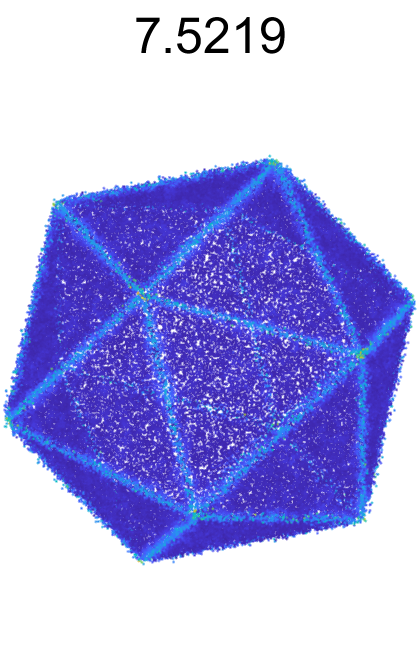}
	\end{subfigure}
	\hspace{0.1cm}
	\begin{subfigure}{0.17\linewidth}
		\centering
		\includegraphics[scale=0.1]{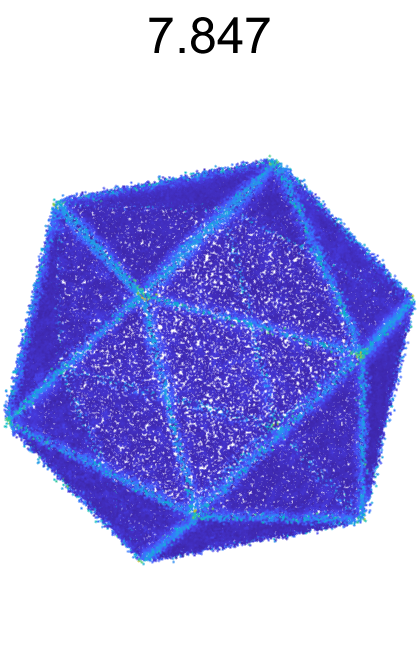}
	\end{subfigure}
	\hspace{0.1cm}
	\begin{subfigure}{0.17\linewidth}
		\centering
		\includegraphics[scale=0.1]{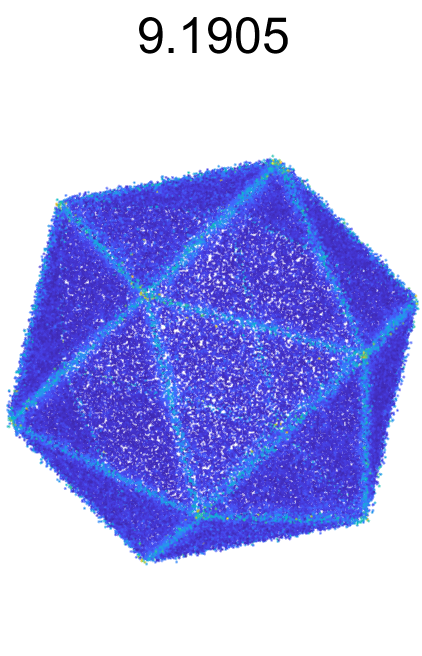}
	\end{subfigure}
	
	\begin{subfigure}{0.15\linewidth}
		\centering
		\includegraphics[scale=0.1]{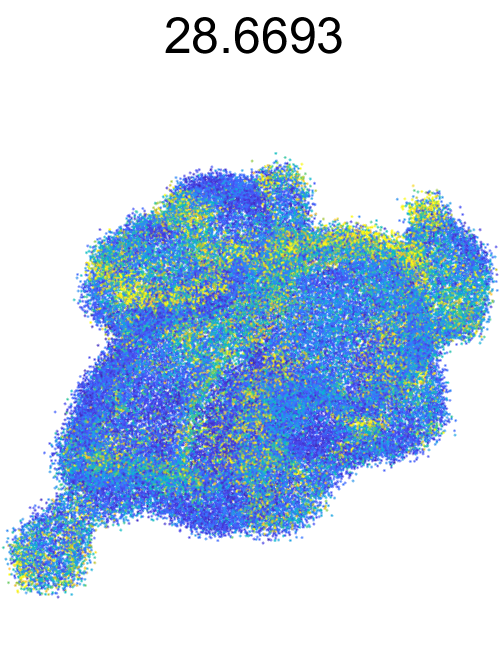}
	\end{subfigure}
	\hspace{0.1cm}
	\begin{subfigure}{0.15\linewidth}
		\centering
		\includegraphics[scale=0.1]{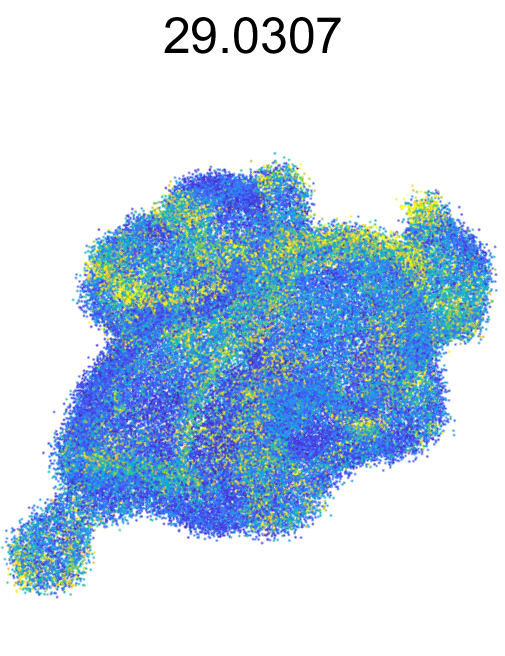}
	\end{subfigure}
	\hspace{0.1cm}
	\begin{subfigure}{0.17\linewidth}
		\centering
		\includegraphics[scale=0.1]{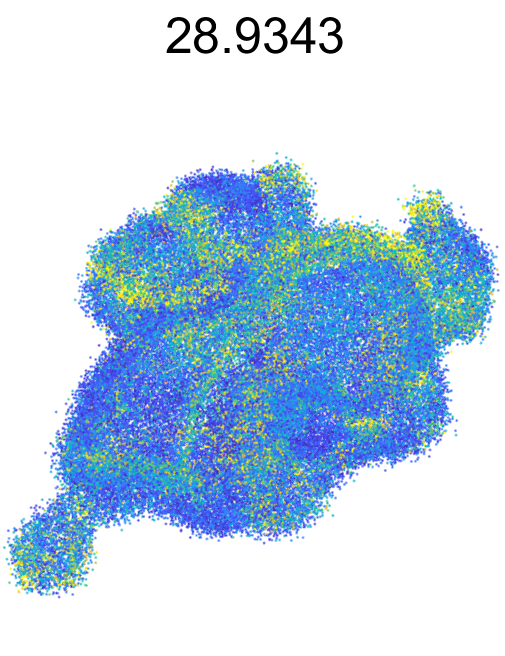}
	\end{subfigure}
	\hspace{0.1cm}
	\begin{subfigure}{0.17\linewidth}
		\centering
		\includegraphics[scale=0.1]{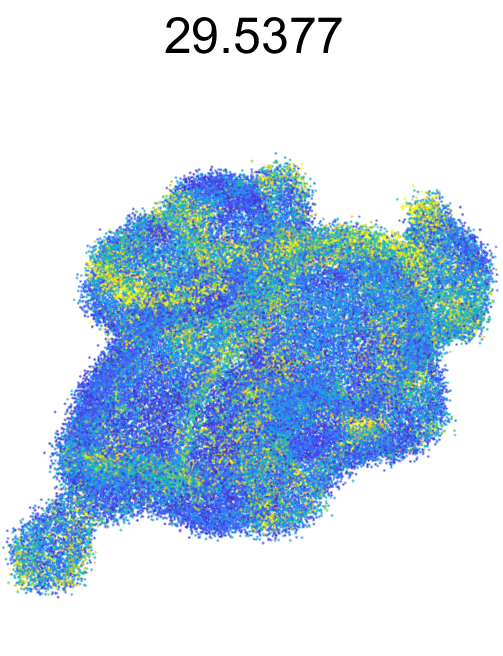}
	\end{subfigure}
	\hspace{0.1cm}
	\begin{subfigure}{0.17\linewidth}
		\centering
		\includegraphics[scale=0.1]{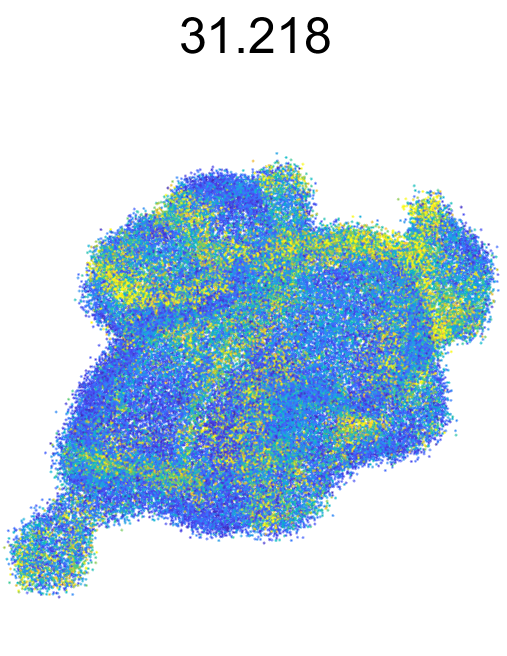}
	\end{subfigure}
	
	\begin{subfigure}{0.15\linewidth}
		\centering
		\includegraphics[scale=0.1]{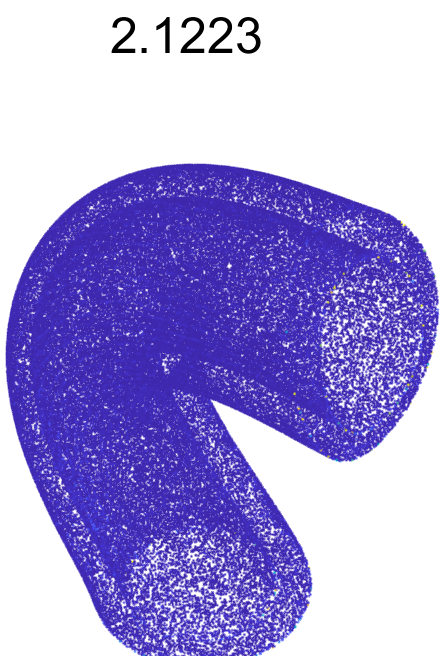}
	\end{subfigure}
	\hspace{0.1cm}
	\begin{subfigure}{0.15\linewidth}
		\centering
		\includegraphics[scale=0.1]{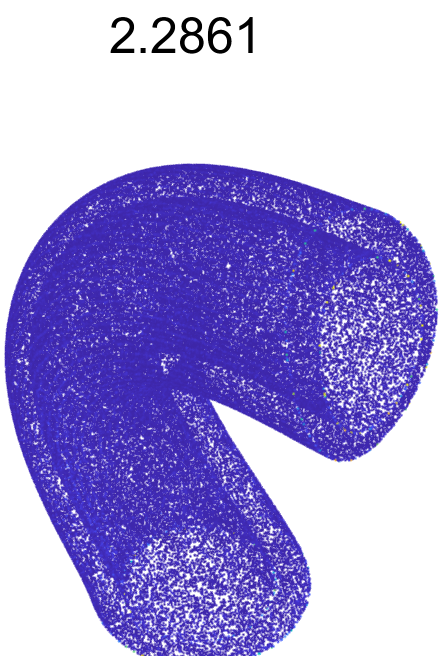}
	\end{subfigure}
	\hspace{0.1cm}
	\begin{subfigure}{0.17\linewidth}
		\centering
		\includegraphics[scale=0.1]{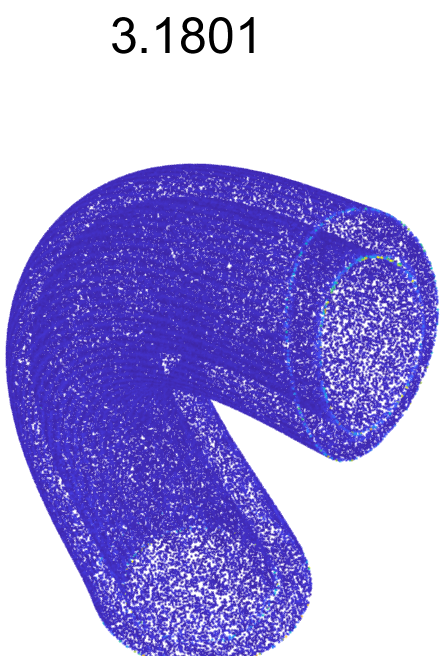}
	\end{subfigure}
	\hspace{0.1cm}
	\begin{subfigure}{0.17\linewidth}
		\centering
		\includegraphics[scale=0.1]{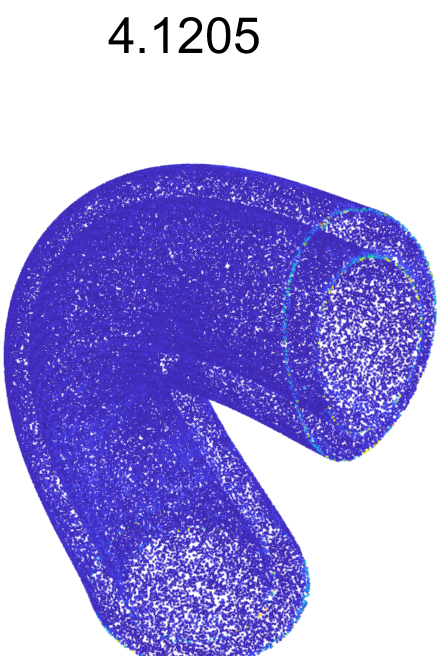}
	\end{subfigure}
	\hspace{0.1cm}
	\begin{subfigure}{0.17\linewidth}
		\centering
		\includegraphics[scale=0.1]{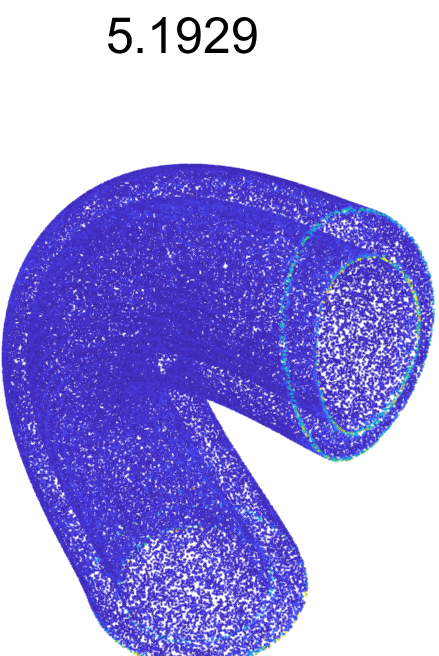}
	\end{subfigure}
	
	\begin{subfigure}{0.15\linewidth}
		\centering
		\includegraphics[scale=0.1]{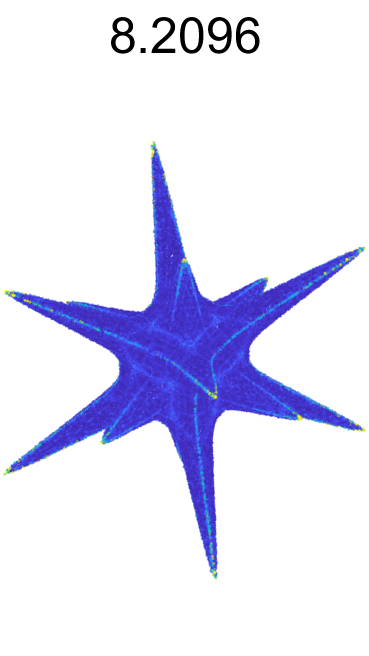}
	\end{subfigure}
	\hspace{0.1cm}
	\begin{subfigure}{0.15\linewidth}
		\centering
		\includegraphics[scale=0.1]{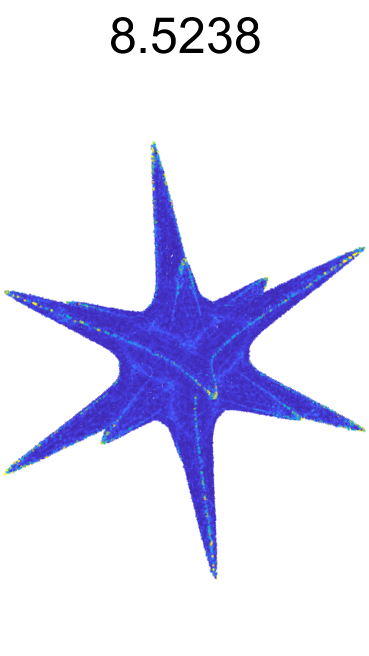}
	\end{subfigure}
	\hspace{0.1cm}
	\begin{subfigure}{0.17\linewidth}
		\centering
		\includegraphics[scale=0.1]{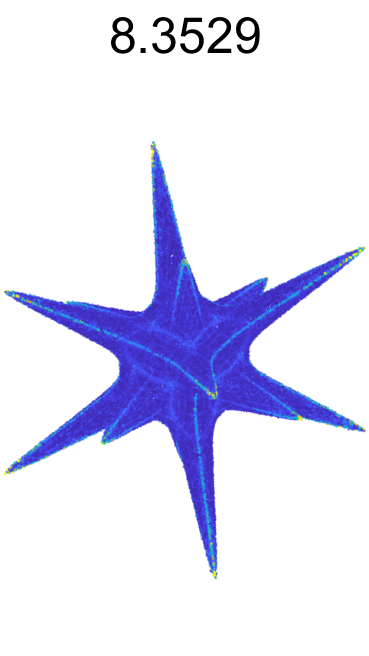}
	\end{subfigure}
	\hspace{0.1cm}
	\begin{subfigure}{0.17\linewidth}
		\centering
		\includegraphics[scale=0.1]{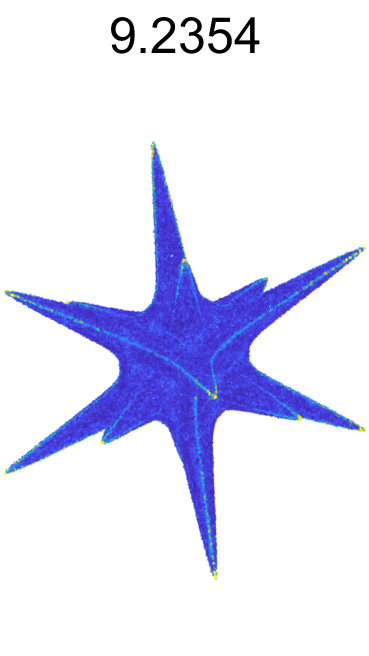}
	\end{subfigure}
	\hspace{0.1cm}
	\begin{subfigure}{0.17\linewidth}
		\centering
		\includegraphics[scale=0.1]{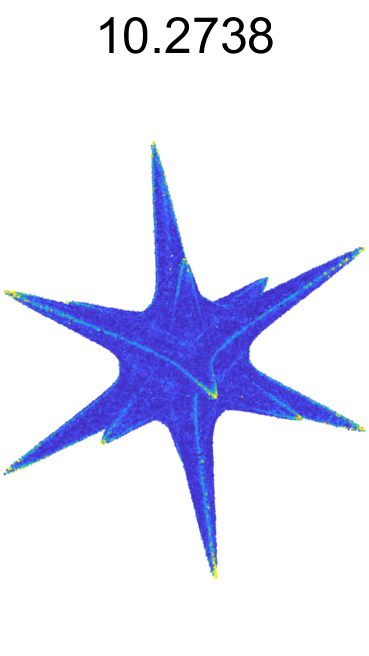}
	\end{subfigure}
	
	\begin{subfigure}{0.15\linewidth}
		\centering
		\includegraphics[scale=0.1]{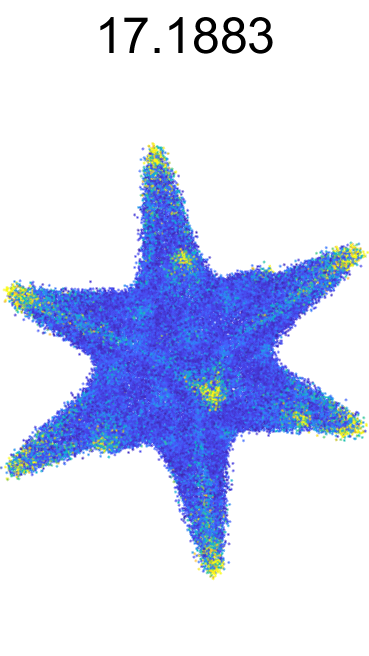}
		\caption{Ours}
	\end{subfigure}
	\hspace{0.1cm}
	\begin{subfigure}{0.15\linewidth}
		\centering
		\includegraphics[scale=0.1]{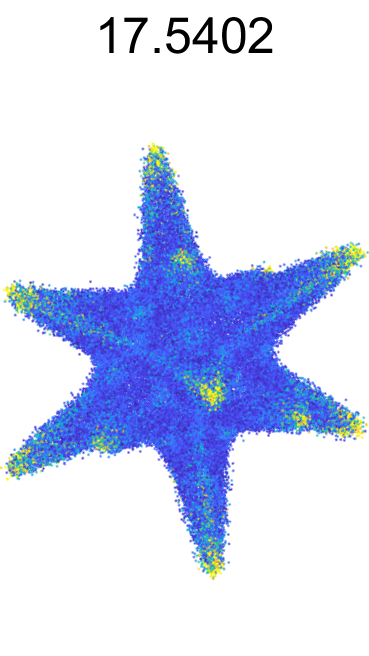}
		\caption{AdaFit}
	\end{subfigure}
	\hspace{0.1cm}
	\begin{subfigure}{0.17\linewidth}
		\centering
		\includegraphics[scale=0.1]{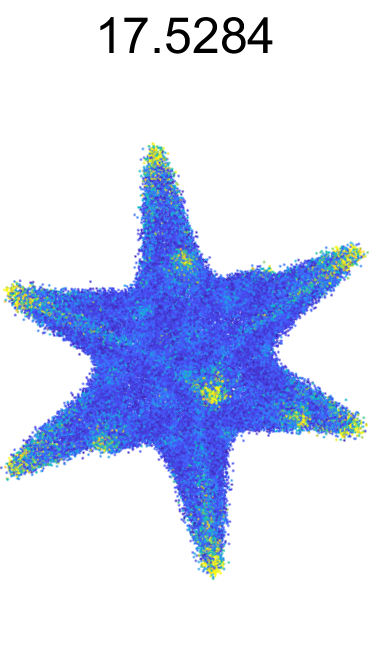}
		\caption{DeepFit}
	\end{subfigure}
	\hspace{0.1cm}
	\begin{subfigure}{0.17\linewidth}
		\centering
		\includegraphics[scale=0.1]{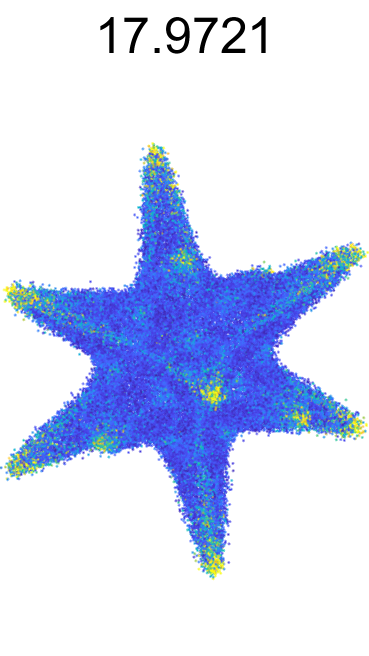}
		\caption{NestiNet}
	\end{subfigure}
	\hspace{0.1cm}
	\begin{subfigure}{0.17\linewidth}
		\centering
		\includegraphics[scale=0.1]{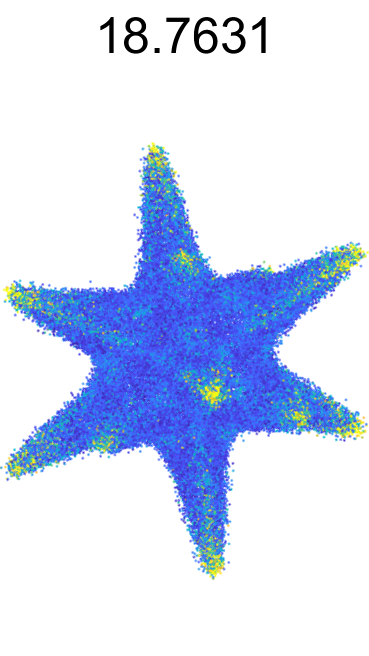}
		\hspace{0.3cm}
		\includegraphics[scale=0.1]{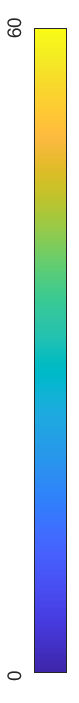}
		\caption{PCPNet}
	\end{subfigure}
	\caption{Illustration of the normal estimation errors.
		The errors are mapped to a heatmap ranging from $0^\circ$ to $60^\circ$. Values above the models are the corresponding RMSE. Our method achieves higher accuracy.
	}
	\label{fig:quali3}
\end{figure*}

\begin{figure*}
	\centering
	\begin{subfigure}{0.15\linewidth}
		\centering
		\includegraphics[scale=0.1]{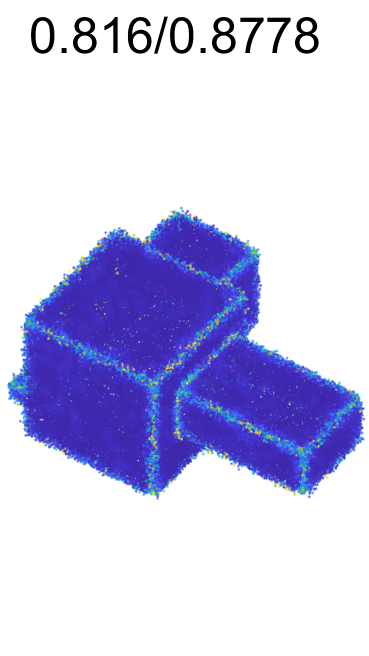}
	\end{subfigure}
	\hspace{0.1cm}
	\begin{subfigure}{0.15\linewidth}
		\centering
		\includegraphics[scale=0.1]{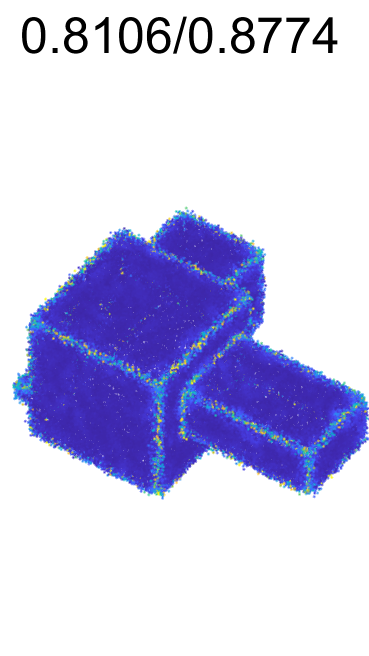}
	\end{subfigure}
	\hspace{0.1cm}
	\begin{subfigure}{0.17\linewidth}
		\centering
		\includegraphics[scale=0.1]{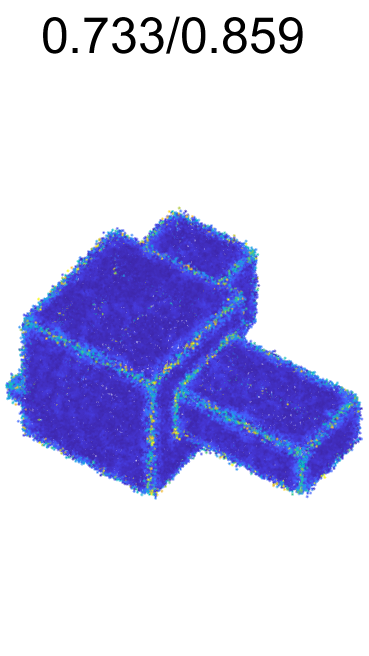}
	\end{subfigure}
	\hspace{0.1cm}
	\begin{subfigure}{0.17\linewidth}
		\centering
		\includegraphics[scale=0.1]{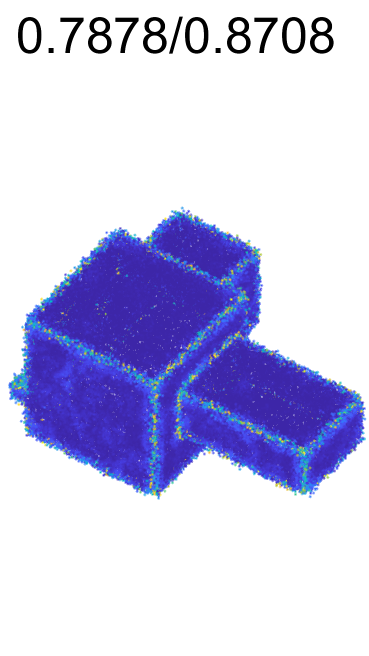}
	\end{subfigure}
	\hspace{0.1cm}
	\begin{subfigure}{0.17\linewidth}
		\centering
		\includegraphics[scale=0.1]{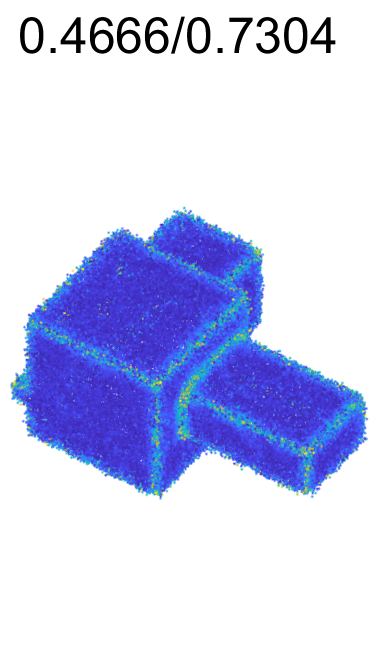}
	\end{subfigure}
	
	\begin{subfigure}{0.15\linewidth}
		\centering
		\includegraphics[scale=0.1]{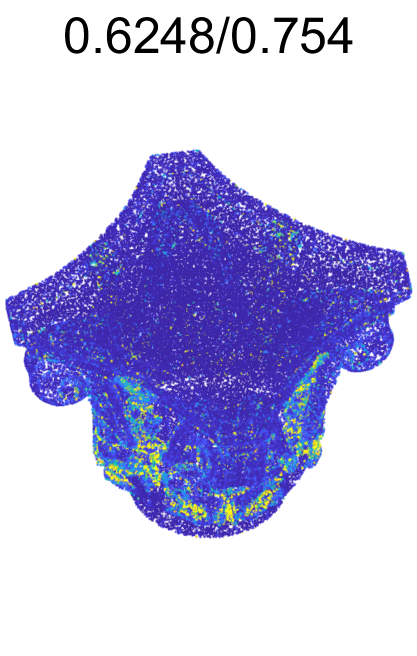}
	\end{subfigure}
	\hspace{0.1cm}
	\begin{subfigure}{0.15\linewidth}
		\centering
		\includegraphics[scale=0.1]{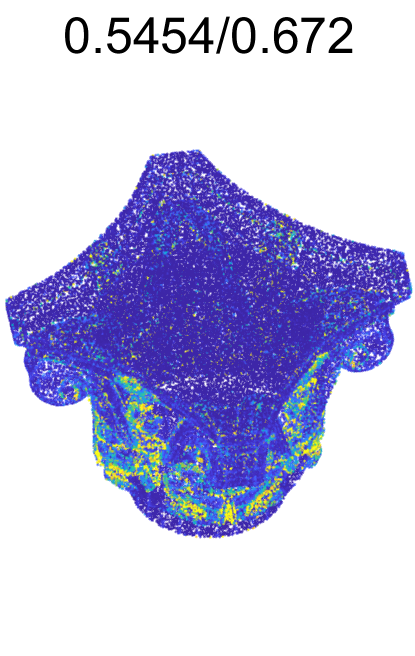}
	\end{subfigure}
	\hspace{0.1cm}
	\begin{subfigure}{0.17\linewidth}
		\centering
		\includegraphics[scale=0.1]{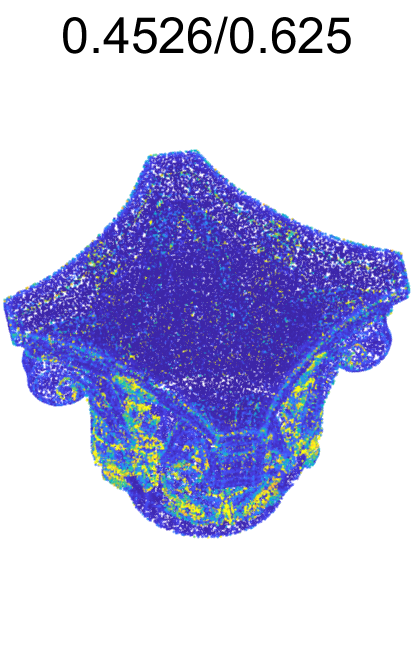}
	\end{subfigure}
	\hspace{0.1cm}
	\begin{subfigure}{0.17\linewidth}
		\centering
		\includegraphics[scale=0.1]{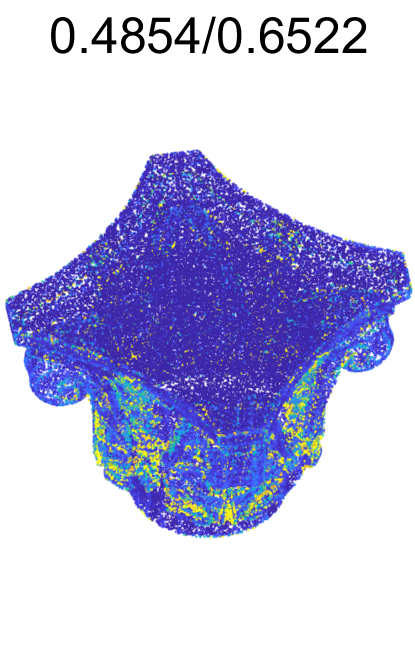}
	\end{subfigure}
	\hspace{0.1cm}
	\begin{subfigure}{0.17\linewidth}
		\centering
		\includegraphics[scale=0.1]{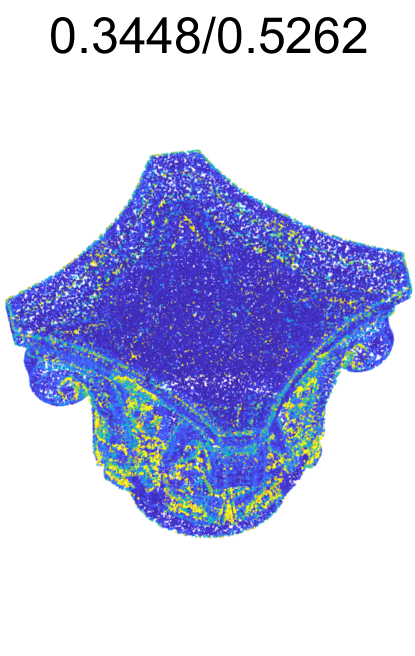}
	\end{subfigure}
	
	\begin{subfigure}{0.15\linewidth}
		\centering
		\includegraphics[scale=0.1]{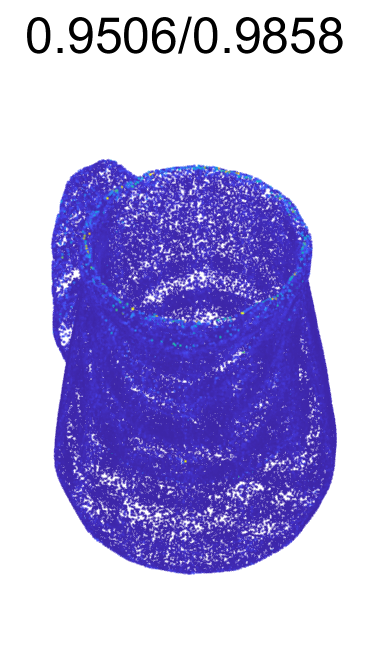}
	\end{subfigure}
	\hspace{0.1cm}
	\begin{subfigure}{0.15\linewidth}
		\centering
		\includegraphics[scale=0.1]{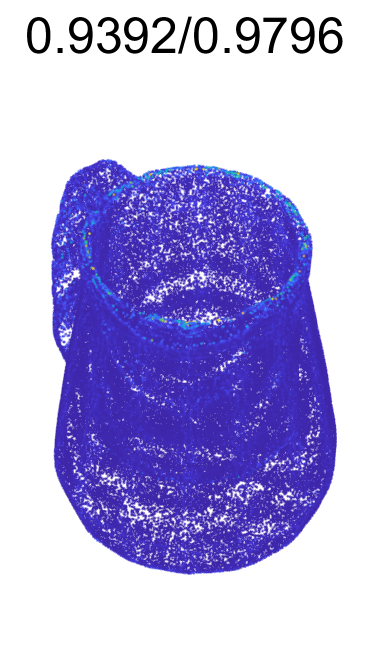}
	\end{subfigure}
	\hspace{0.1cm}
	\begin{subfigure}{0.17\linewidth}
		\centering
		\includegraphics[scale=0.1]{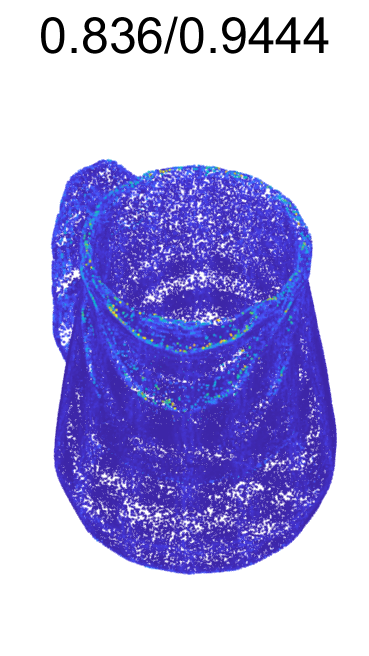}
	\end{subfigure}
	\hspace{0.1cm}
	\begin{subfigure}{0.17\linewidth}
		\centering
		\includegraphics[scale=0.1]{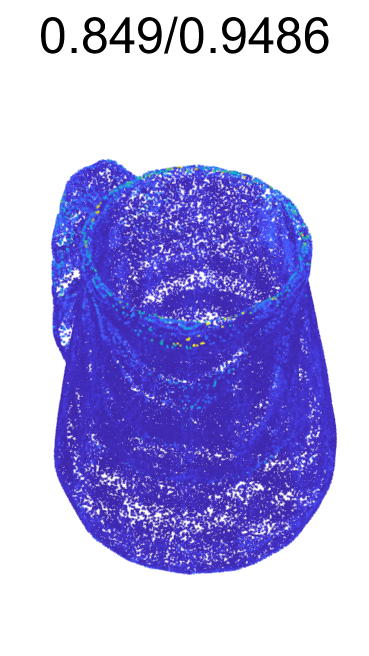}
	\end{subfigure}
	\hspace{0.1cm}
	\begin{subfigure}{0.17\linewidth}
		\centering
		\includegraphics[scale=0.1]{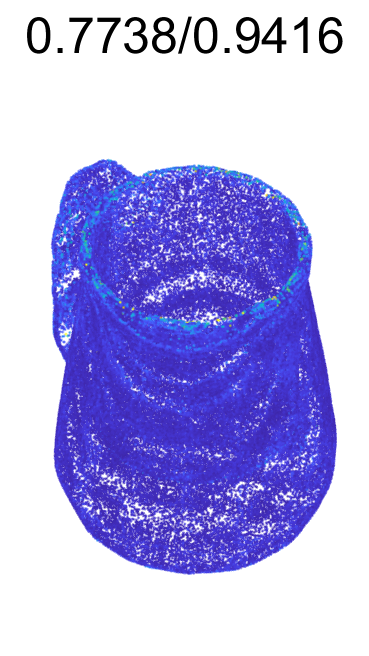}
	\end{subfigure}
	
	\begin{subfigure}{0.15\linewidth}
		\centering
		\includegraphics[scale=0.1]{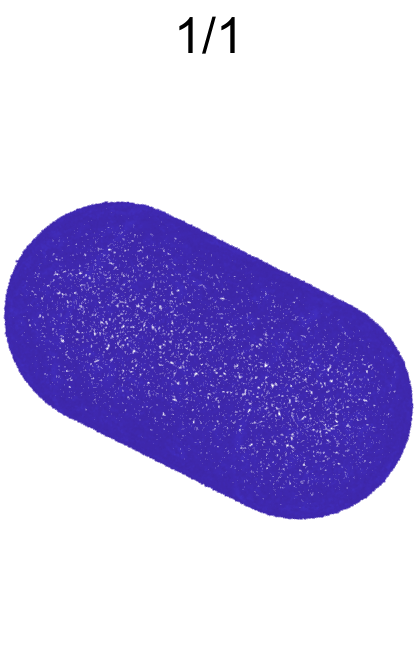}
	\end{subfigure}
	\hspace{0.1cm}
	\begin{subfigure}{0.15\linewidth}
		\centering
		\includegraphics[scale=0.1]{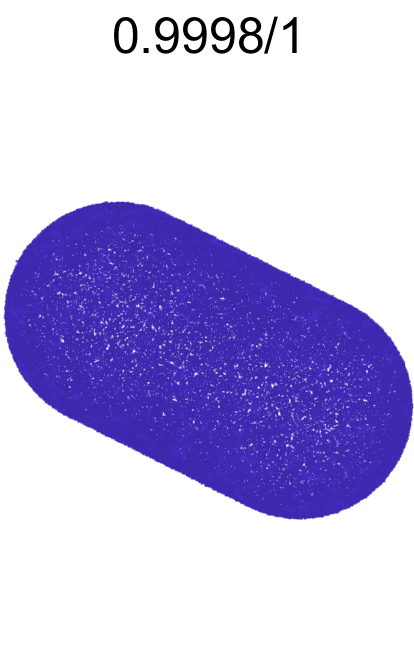}
	\end{subfigure}
	\hspace{0.1cm}
	\begin{subfigure}{0.17\linewidth}
		\centering
		\includegraphics[scale=0.1]{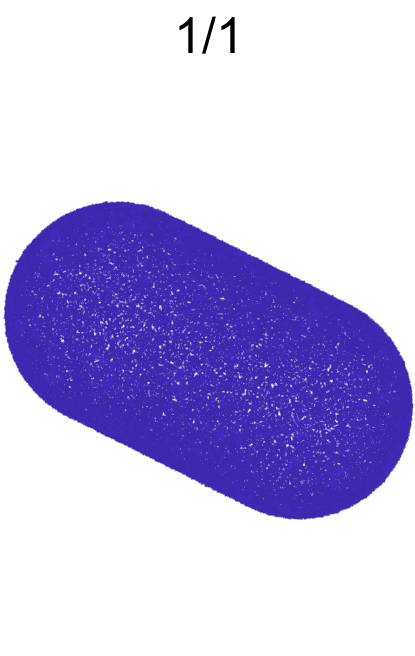}
	\end{subfigure}
	\hspace{0.1cm}
	\begin{subfigure}{0.17\linewidth}
		\centering
		\includegraphics[scale=0.1]{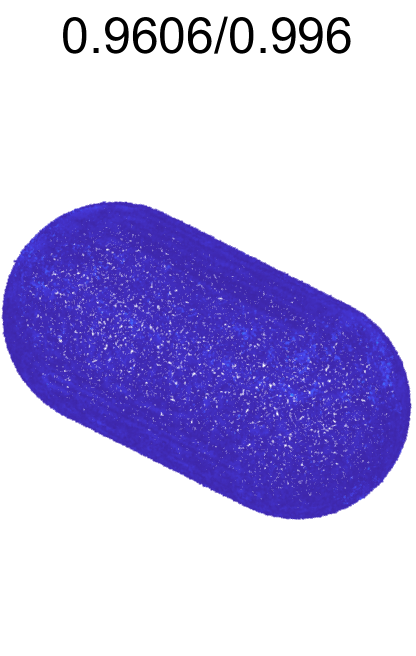}
	\end{subfigure}
	\hspace{0.1cm}
	\begin{subfigure}{0.17\linewidth}
		\centering
		\includegraphics[scale=0.1]{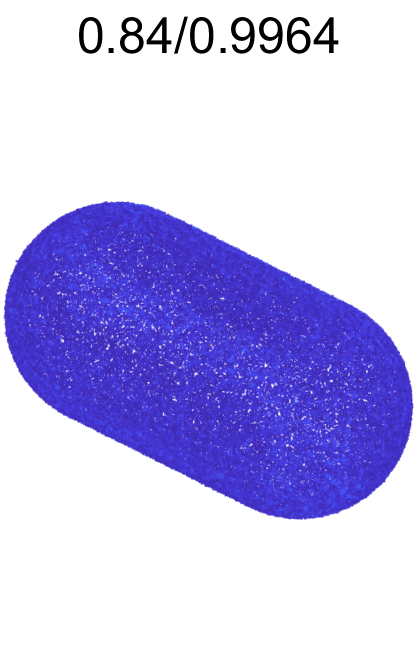}
	\end{subfigure}
	
	\begin{subfigure}{0.15\linewidth}
		\centering
		\includegraphics[scale=0.1]{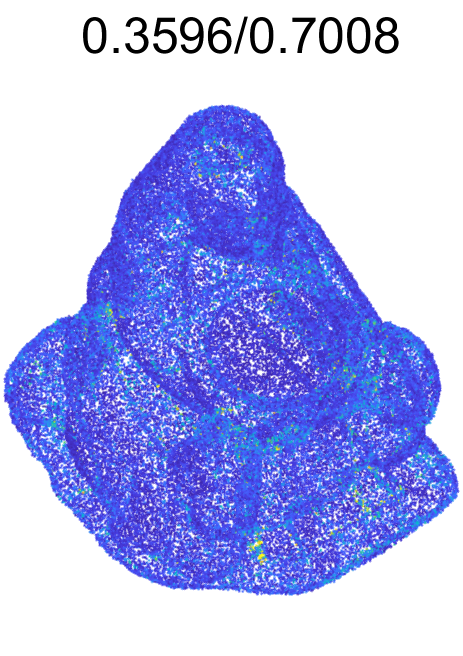}
	\end{subfigure}
	\hspace{0.1cm}
	\begin{subfigure}{0.15\linewidth}
		\centering
		\includegraphics[scale=0.1]{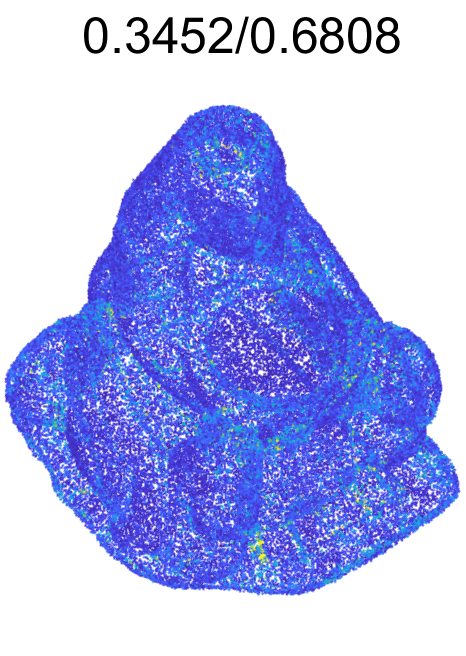}
	\end{subfigure}
	\hspace{0.1cm}
	\begin{subfigure}{0.17\linewidth}
		\centering
		\includegraphics[scale=0.1]{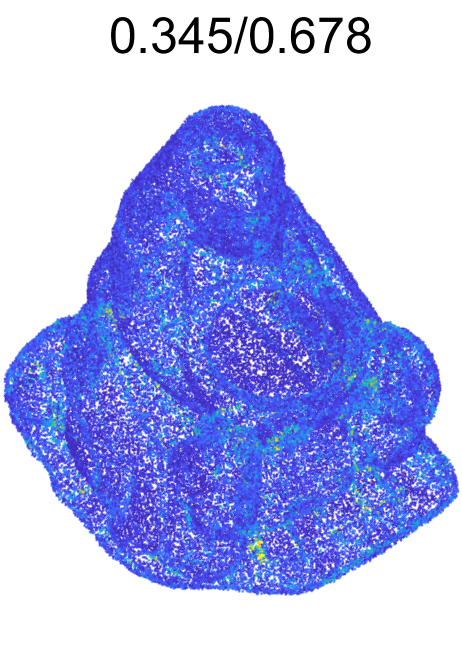}
	\end{subfigure}
	\hspace{0.1cm}
	\begin{subfigure}{0.17\linewidth}
		\centering
		\includegraphics[scale=0.1]{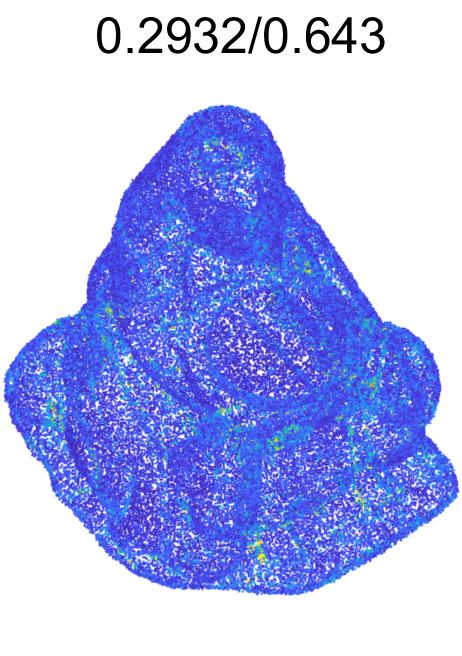}
	\end{subfigure}
	\hspace{0.1cm}
	\begin{subfigure}{0.17\linewidth}
		\centering
		\includegraphics[scale=0.1]{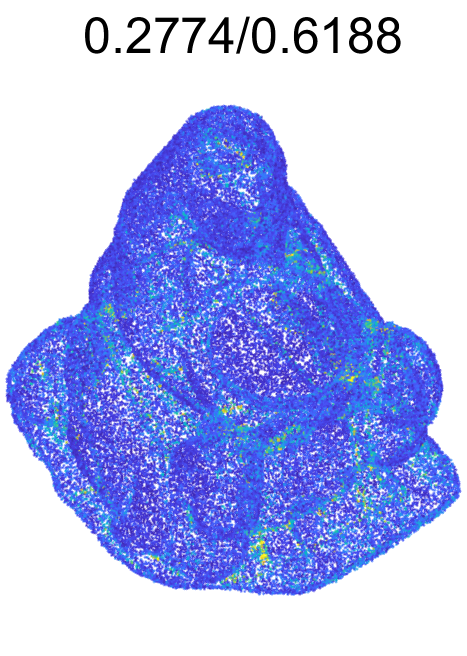}
	\end{subfigure}
	
	\begin{subfigure}{0.15\linewidth}
		\centering
		\includegraphics[scale=0.1]{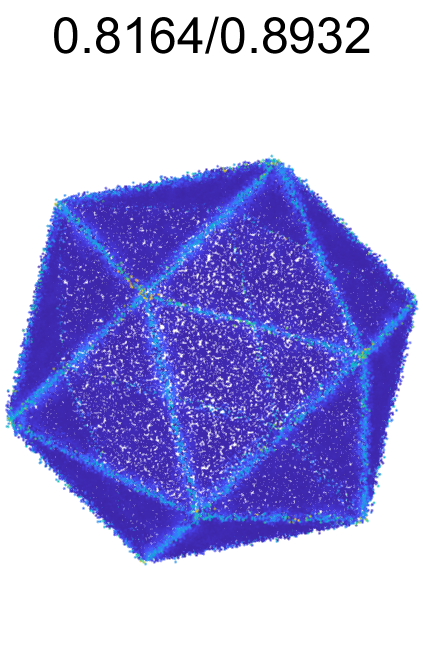}
	\end{subfigure}
	\hspace{0.1cm}
	\begin{subfigure}{0.15\linewidth}
		\centering
		\includegraphics[scale=0.1]{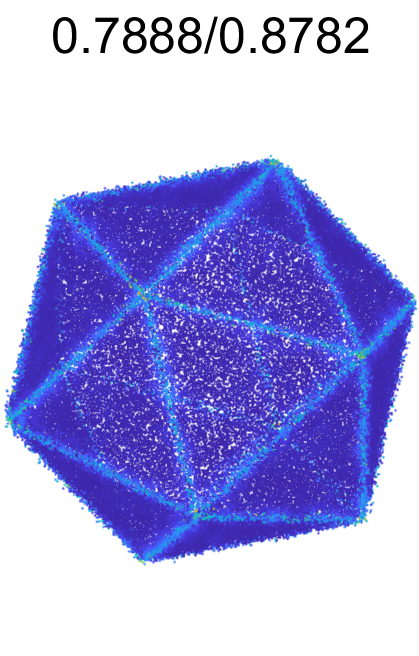}
	\end{subfigure}
	\hspace{0.1cm}
	\begin{subfigure}{0.17\linewidth}
		\centering
		\includegraphics[scale=0.1]{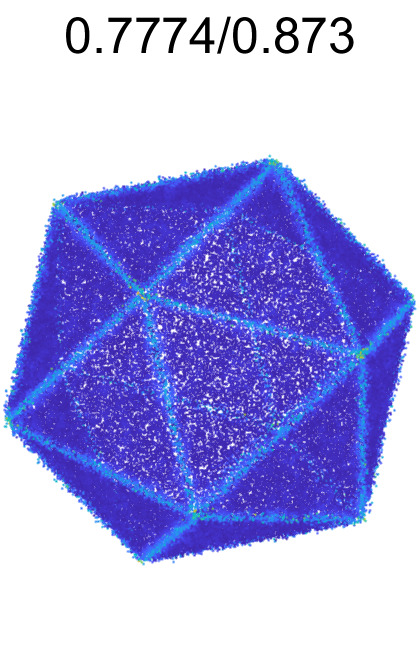}
	\end{subfigure}
	\hspace{0.1cm}
	\begin{subfigure}{0.17\linewidth}
		\centering
		\includegraphics[scale=0.1]{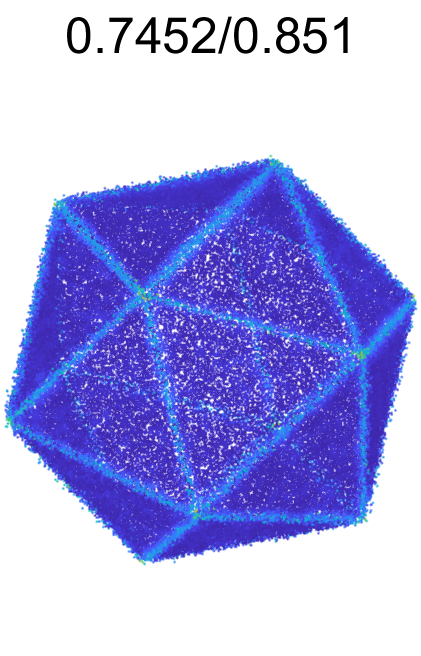}
	\end{subfigure}
	\hspace{0.1cm}
	\begin{subfigure}{0.17\linewidth}
		\centering
		\includegraphics[scale=0.1]{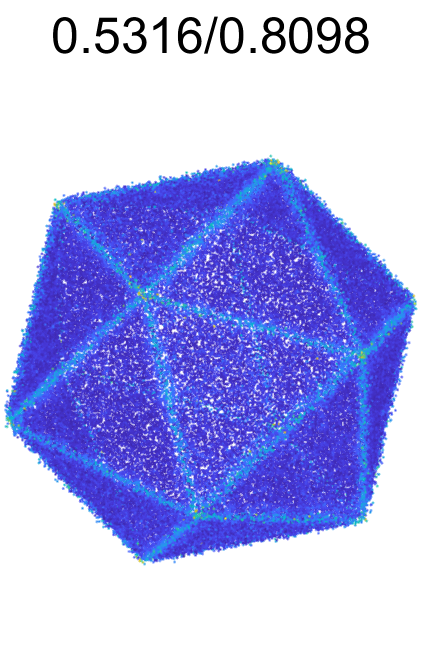}
	\end{subfigure}
	
	\begin{subfigure}{0.15\linewidth}
		\centering
		\includegraphics[scale=0.1]{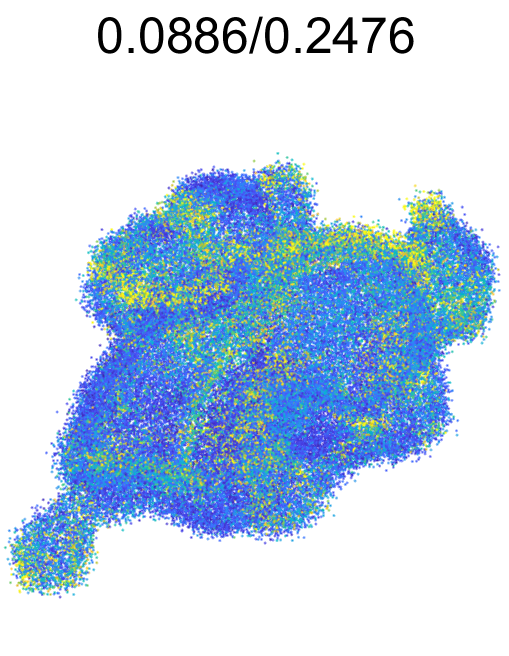}
	\end{subfigure}
	\hspace{0.1cm}
	\begin{subfigure}{0.15\linewidth}
		\centering
		\includegraphics[scale=0.1]{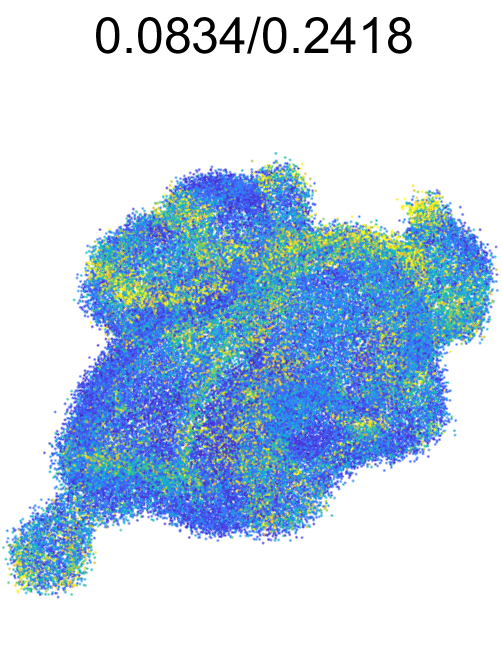}
	\end{subfigure}
	\hspace{0.1cm}
	\begin{subfigure}{0.17\linewidth}
		\centering
		\includegraphics[scale=0.1]{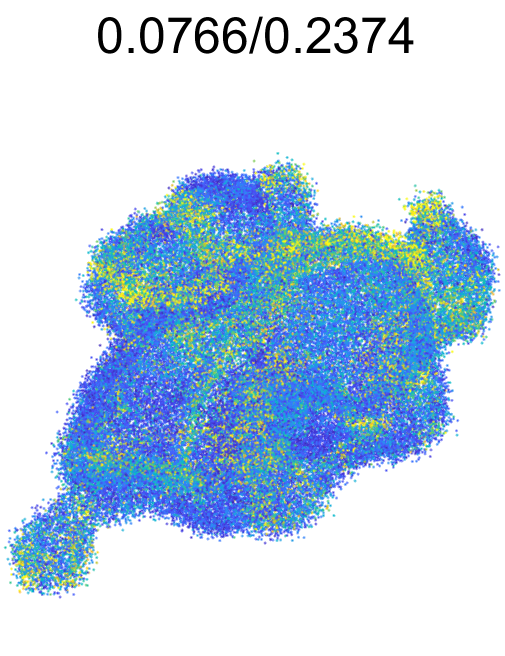}
	\end{subfigure}
	\hspace{0.1cm}
	\begin{subfigure}{0.17\linewidth}
		\centering
		\includegraphics[scale=0.1]{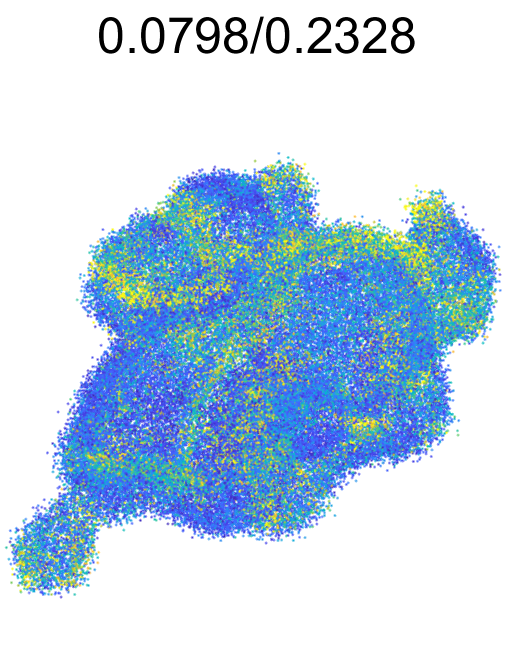}
	\end{subfigure}
	\hspace{0.1cm}
	\begin{subfigure}{0.17\linewidth}
		\centering
		\includegraphics[scale=0.1]{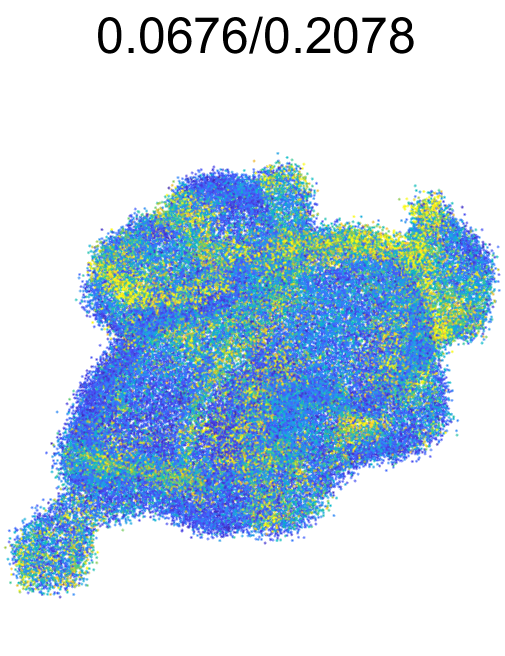}
	\end{subfigure}
	
	\begin{subfigure}{0.15\linewidth}
		\centering
		\includegraphics[scale=0.1]{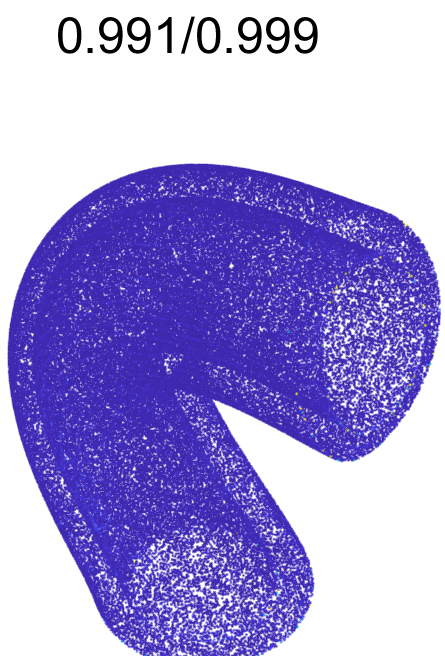}
	\end{subfigure}
	\hspace{0.1cm}
	\begin{subfigure}{0.15\linewidth}
		\centering
		\includegraphics[scale=0.1]{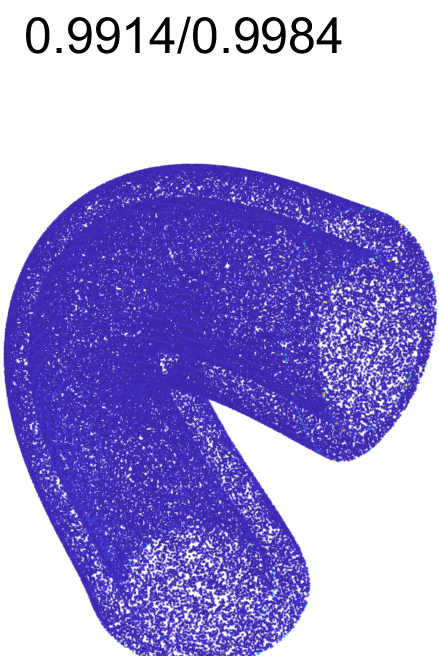}
	\end{subfigure}
	\hspace{0.1cm}
	\begin{subfigure}{0.17\linewidth}
		\centering
		\includegraphics[scale=0.1]{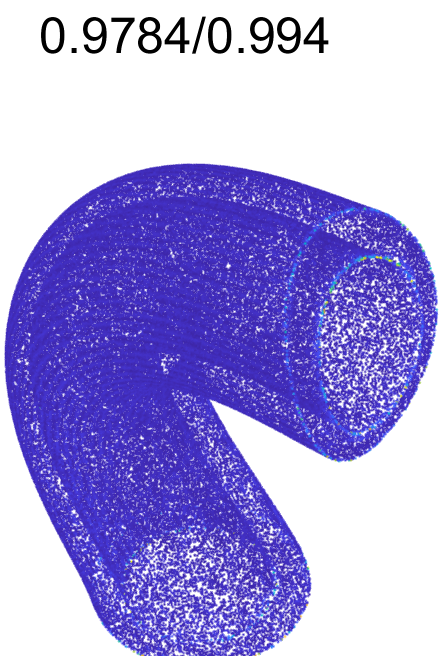}
	\end{subfigure}
	\hspace{0.1cm}
	\begin{subfigure}{0.17\linewidth}
		\centering
		\includegraphics[scale=0.1]{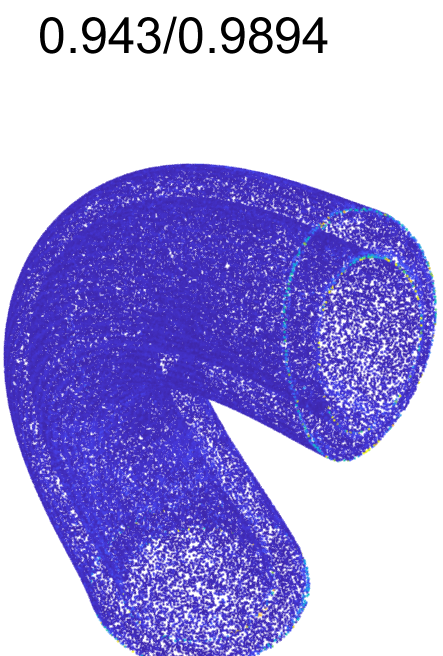}
	\end{subfigure}
	\hspace{0.1cm}
	\begin{subfigure}{0.17\linewidth}
		\centering
		\includegraphics[scale=0.1]{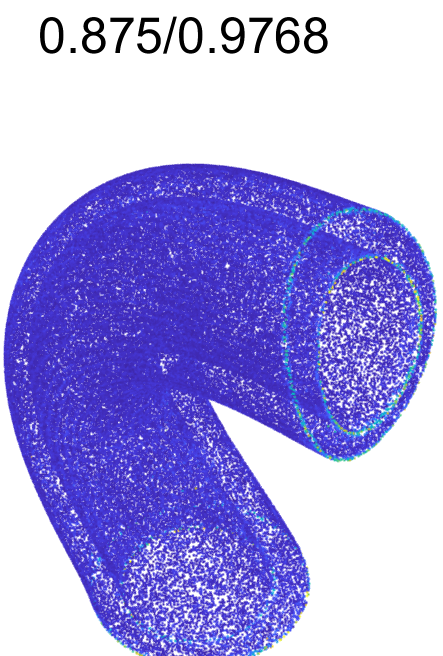}
	\end{subfigure}
	
	\begin{subfigure}{0.15\linewidth}
		\centering
		\includegraphics[scale=0.1]{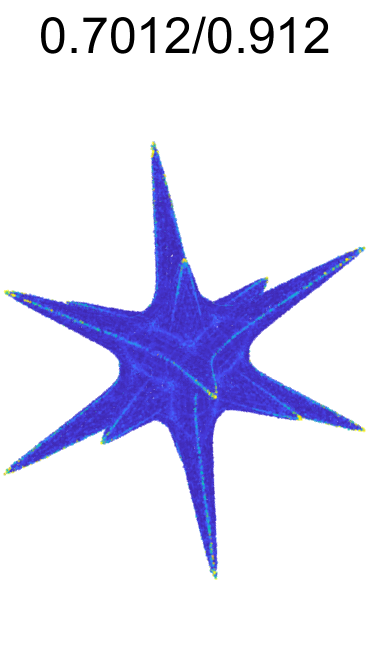}
	\end{subfigure}
	\hspace{0.1cm}
	\begin{subfigure}{0.15\linewidth}
		\centering
		\includegraphics[scale=0.1]{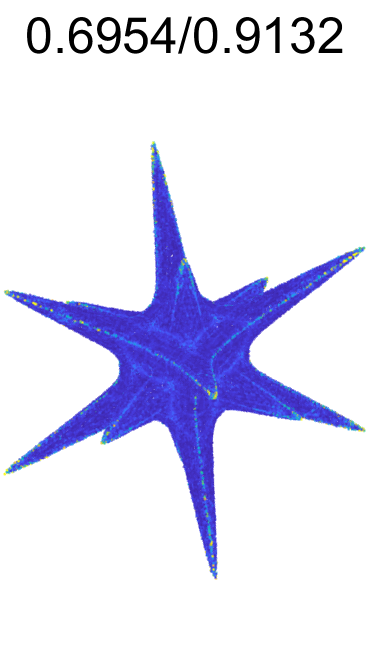}
	\end{subfigure}
	\hspace{0.1cm}
	\begin{subfigure}{0.17\linewidth}
		\centering
		\includegraphics[scale=0.1]{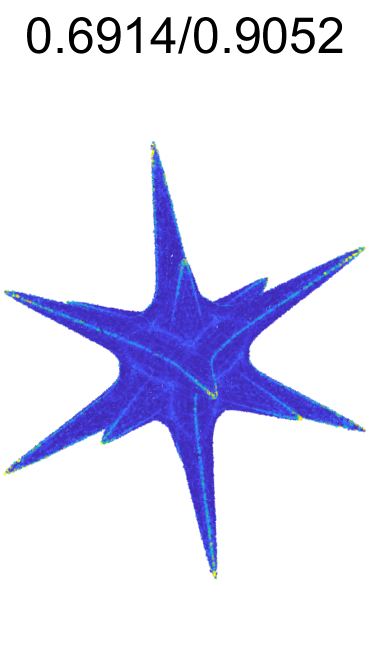}
	\end{subfigure}
	\hspace{0.1cm}
	\begin{subfigure}{0.17\linewidth}
		\centering
		\includegraphics[scale=0.1]{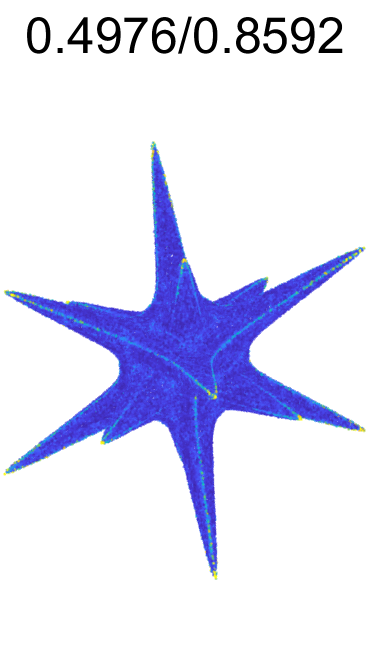}
	\end{subfigure}
	\hspace{0.1cm}
	\begin{subfigure}{0.17\linewidth}
		\centering
		\includegraphics[scale=0.1]{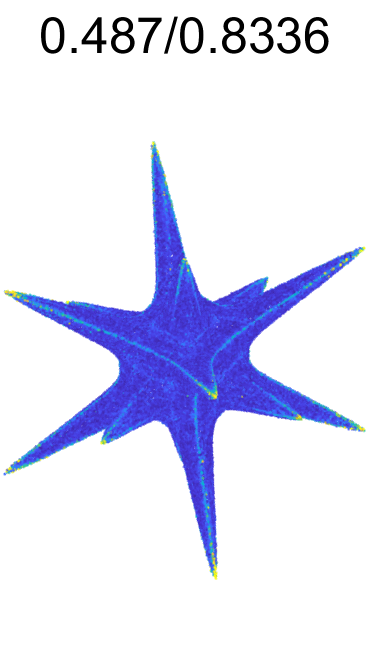}
	\end{subfigure}
	
	\begin{subfigure}{0.15\linewidth}
		\centering
		\includegraphics[scale=0.1]{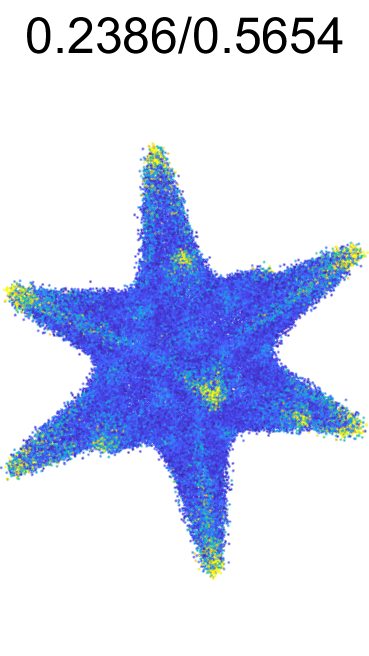}
		\caption{Ours}
	\end{subfigure}
	\hspace{0.1cm}
	\begin{subfigure}{0.15\linewidth}
		\centering
		\includegraphics[scale=0.1]{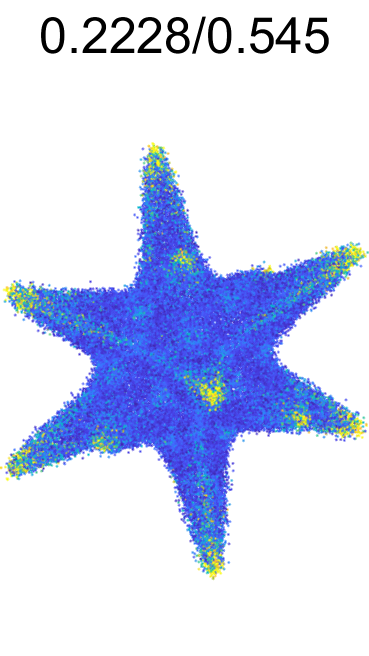}
		\caption{AdaFit}
	\end{subfigure}
	\hspace{0.1cm}
	\begin{subfigure}{0.17\linewidth}
		\centering
		\includegraphics[scale=0.1]{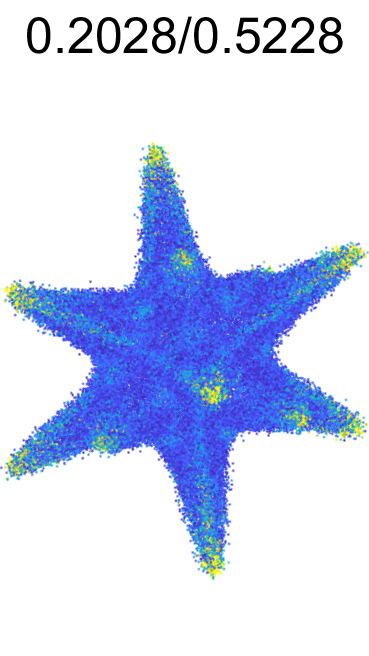}
		\caption{DeepFit}
	\end{subfigure}
	\hspace{0.1cm}
	\begin{subfigure}{0.17\linewidth}
		\centering
		\includegraphics[scale=0.1]{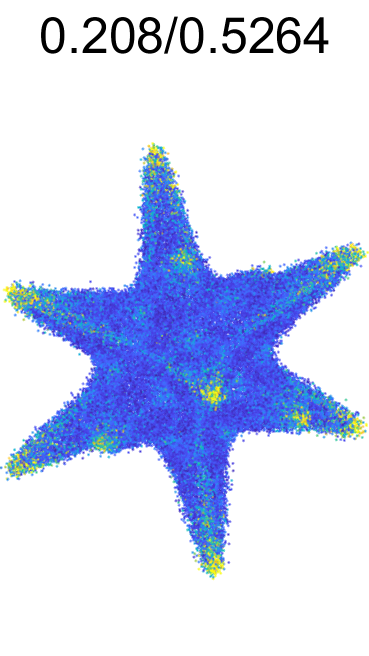}
		\caption{NestiNet}
	\end{subfigure}
	\hspace{0.1cm}
	\begin{subfigure}{0.17\linewidth}
		\centering
		\includegraphics[scale=0.1]{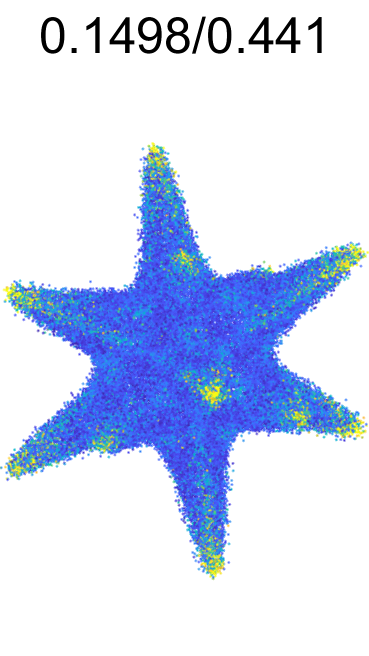}
		\caption{PCPNet}
	\end{subfigure}
	\caption{Qualitative results with respect to $\operatorname{PGP}(5)$ and $\operatorname{PGP}(10)$. Values above the models are the corresponding $\operatorname{PGP5/10}$ errors. Our method attains more accurate normal estimation.
	}
	\label{fig:quali4}
\end{figure*}

We present more test results in \cref{fig:quali3} and \cref{fig:quali4} to demonstrate the better performance of our proposed normal estimator. 

\end{document}